\documentclass[10pt,twocolumn,letterpaper]{article}

\usepackage{cvpr}      % To produce the REVIEW version
\makeatletter
\@namedef{ver@everyshi.sty}{}
\makeatother
\usepackage{tikz}

\usepackage{graphicx}
\usepackage{amsmath}
\usepackage{amssymb}
\usepackage{booktabs}
\usepackage{mdframed}
\usepackage{xcolor,soul}
\usepackage{enumitem}
\usepackage{pifont}% 
\usepackage{makecell}
\usepackage{varwidth}
\usepackage{multirow}
\usepackage[super]{nth}
\usepackage[pagebackref,breaklinks,colorlinks,hypertexnames=false]{hyperref}

\usepackage[capitalize]{cleveref}
\crefname{section}{Sec.}{Secs.}
\Crefname{section}{Section}{Sections}
\Crefname{table}{Table}{Tables}
\crefname{table}{Tab.}{Tabs.}

\def\cvprPaperID{7457} % *** Enter the CVPR Paper ID here
\def\confName{CVPR}
\def\confYear{2022}

\begin{document}
%\definecolor{myOrange}{rgb}{1.0, 0.49, 0.0}
\definecolor{myOrange}{rgb}{1.00, 0.60, 0}
%{1.0, 0.55, 0.0}
\definecolor{myblue}{rgb}{0.0,0.40,0.80}
\newcommand{\model}{\textit{CCN}}
\pdfoutput=1
%{0.216, 0.494, 0.722}
%{0.19, 0.55, 0.91}

\newcommand{\cmark}{\ding{51}}%
\newcommand{\xmark}{\ding{55}}%

\newcommand{\vcenteredinclude}[1]{\begingroup
\setbox0=\hbox{\includegraphics[scale=0.17]{#1}}%
\parbox{\wd0}{\box0}\endgroup}

\newcommand{\mycircle}[2][black,fill=black]{\tikz[baseline=-0.5ex]\draw[#1,radius=#2] (0,0) circle ;}%

\definecolor{applegreen}{rgb}{0.55, 0.71, 0.0}
\definecolor{harlequin}{rgb}{0.25, 1.0, 0.0}
\def\boxit#1{%
  \smash{\color{harlequin}\fboxrule=1pt\relax\fboxsep=4pt\relax%
  \llap{\rlap{\fbox{\vphantom{0}\makebox[#1]{}}}~}}\ignorespaces
}

\title{Open-Domain, Content-based, Multi-modal Fact-checking of\\Out-of-Context Images via Online Resources}

\author{Sahar Abdelnabi, Rakibul Hasan, and Mario Fritz\\
CISPA Helmholtz Center for Information Security\\
{\tt\small \{sahar.abdelnabi,rakibul.hasan,fritz\}@cispa.de}}
%\and
%\maketitle
\twocolumn[{%
\renewcommand\twocolumn[1][]{#1}%
\maketitle
\begin{center}
    \centering
    \captionsetup{type=figure}
    \includegraphics[width=0.90\textwidth]{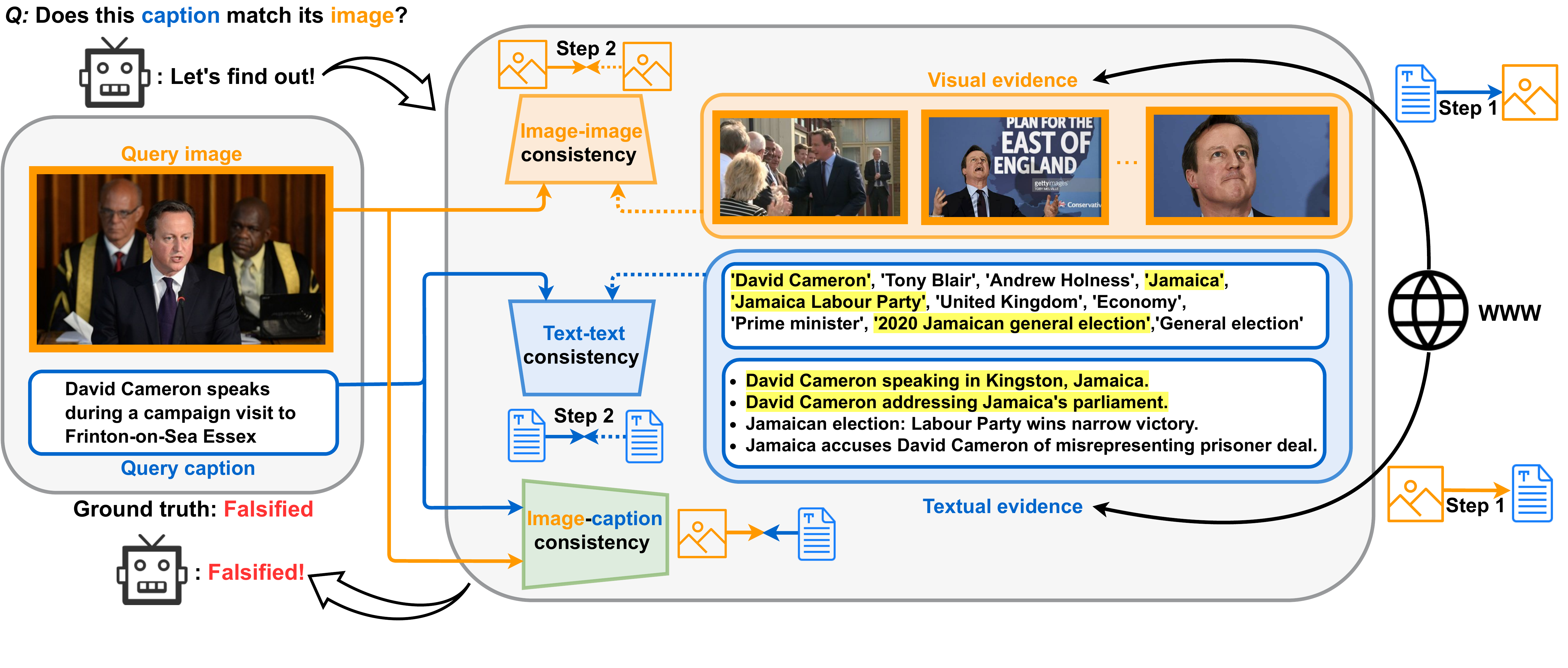}
    \captionof{figure}{To evaluate the veracity of \textbf{\textcolor{myOrange}{image}}-\textbf{\textcolor{myblue}{caption}} pairings, we leverage \textbf{\textcolor{myOrange}{visual}} and \textbf{\textcolor{myblue}{textual}} evidence gathered by querying the Web. We propose a novel framework to detect the consistency of the claim-evidence (\textbf{\textcolor{myblue}{text}}-\textbf{\textcolor{myblue}{text}} and \textbf{\textcolor{myOrange}{image}}-\textbf{\textcolor{myOrange}{image}}), in addition to the \textbf{\textcolor{myOrange}{image}}-\textbf{\textcolor{myblue}{caption}} pairing. Highlighted evidence represents the model's highest attention, showing a difference in location compared to the query \textbf{\textcolor{myblue}{caption}}.} \label{fig:teaser}
\end{center}%
}]

%%%%%%%%% ABSTRACT

\begin{abstract}
Misinformation is now a major problem due to its potential high risks to our core democratic and societal values and orders. \textbf{Out-of-context} misinformation is one of the easiest and effective ways used by adversaries to spread viral false stories. In this threat, a real image is \textbf{re-purposed} to support other narratives by misrepresenting its context and/or elements. 
The internet is being used as the go-to way to verify information using different sources and modalities. Our goal is an inspectable method that automates this time-consuming and reasoning-intensive process by fact-checking the \textbf{\textcolor{myOrange}{image}}-\textbf{\textcolor{myblue}{caption}} pairing using Web evidence. 
To integrate evidence and cues from both modalities, we introduce the concept of \textbf{`multi-modal cycle-consistency check'} \vcenteredinclude{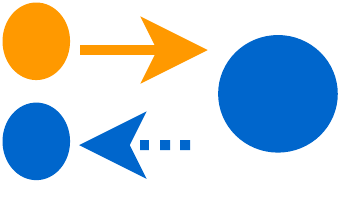} / \vcenteredinclude{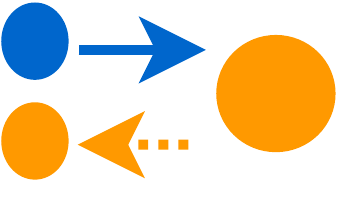}; starting from the \textbf{\textcolor{myOrange}{image}}/\textbf{\textcolor{myblue}{caption}}, we gather \textbf{\textcolor{myblue}{textual}}/\textbf{\textcolor{myOrange}{visual}} evidence, which will be compared against the other paired \textbf{\textcolor{myblue}{caption}}/\textbf{\textcolor{myOrange}{image}}, respectively. 
Moreover, we propose a novel architecture, \textbf{Consistency-Checking Network (CCN)}, that mimics the layered human reasoning across the same and different modalities: the \textbf{\textcolor{myblue}{caption}} vs. \textbf{\textcolor{myblue}{textual evidence}}, the \textbf{\textcolor{myOrange}{image}} vs. \textbf{\textcolor{myOrange}{visual evidence}}, and the \textbf{\textcolor{myOrange}{image}} vs. \textbf{\textcolor{myblue}{caption}}. Our work offers the first step and benchmark for \textbf{open-domain, content-based, multi-modal fact-checking}, and significantly outperforms previous baselines that did not leverage external evidence\footnote{For code, checkpoints, and dataset, check: \url{https://s-abdelnabi.github.io/OoC-multi-modal-fc/}}.

\end{abstract}

%%%%%%%%% BODY TEXT
\vspace{-5mm}
\section{Introduction}

Recently, there has been a growing and widespread concern about `fake news' and its harmful societal, personal, and political consequences~\cite{agarwal2019protecting,huh2018fighting,lazer2018science}, including people's own health during the pandemic\cite{nightingale2020examining,press_bbc_covid,press_bbc_covid2}. Misusing generative AI technologies  
to create deepfakes~\cite{goodfellow2014generative,karras2020analyzing,radford2019language} further fuelled these concerns~\cite{press1,press2}. 
However, \textbf{\textit{image-repurposing}}--- where a real image is misrepresented and used \textit{out-of-context} with another false or unrelated narrative to create more credible stories and mislead the audience---is still one of the easiest and most effective ways to create realistically-looking misinformation. Image-repurposing does not require profound technical knowledge or experience~\cite{luo2021newsclippings,aneja2021catching}, which potentially amplifies its risks. Images usually accompany real news~\cite{tan2020detecting}; thus, adversaries may augment their stories with images as `supporting evidence' to capture readers' attention~\cite{luo2021newsclippings,hameleers2020picture,wang2021understanding}.

\textbf{Image re-purposing datasets and threats.} Gathering large-scale labelled out-of-context datasets is hard due to the scarcity and substantial manual efforts. Thus, previous work attempted to construct synthetic out-of-context datasets~\cite{jaiswal2017multimedia,sabir2018deep}.  
A recent work~\cite{luo2021newsclippings} proposed to automatically, yet non-trivially, match images accompanying real news with other real news captions. The authors used trained language and vision models to retrieve a close and convincing image given a caption. While this work contributes to misinformation detection research by automatically creating datasets, it also highlights the threat that \textbf{\textit{machine-assisted}} procedures may ease creating misinformation at scale. Furthermore, the authors reported that both defense models and humans struggled to detect the out-of-context images. 
In this paper, we use this dataset as a challenging benchmark; we leverage external evidence to push forward the automatic detection. 

\textbf{Fact-checking.} To fight misinformation, huge fact-checking efforts are done by different organizations~\cite{fc1,fc2}. However, they require substantial manual efforts~\cite{press_vice}. Researchers have proposed several automated methods and benchmarks to automate fact-checking and verification~\cite{thorne2018fever,popat2018declare}. However, most of these works focus on textual claims. Fact-checking multi-modal claims has been under-explored.
%\vspace{-3mm}

\textbf{Our approach.} People frequently use the Internet to verify information. We aggregate evidence from images, articles, different sources, and we measure their consensus and consistency. Our goal is to design an inspectable framework that automates this multi-modal fact-checking process and assists users, fact-checkers, and content moderators. 

More specifically, we propose to gather and reason over evidence to judge the veracity of the \textbf{\textcolor{myOrange}{image}}-\textbf{\textcolor{myblue}{caption}} pair. First \vcenteredinclude{figs/icon3.pdf}, we use the \textbf{\textcolor{myOrange}{image}} to find its other occurrences on the internet, from which, we crawl \textbf{\textcolor{myblue}{textual evidence}} (e.g., captions), which we compare against the paired \textbf{\textcolor{myblue}{caption}}. Similarly \vcenteredinclude{figs/icon4.pdf}, we use the \textbf{\textcolor{myblue}{caption}} to find other images as \textbf{\textcolor{myOrange}{visual evidence}} to compare against the paired \textbf{\textcolor{myOrange}{image}}. We call this process: \textbf{`multi-modal cycle-consistency check'}. Importantly, we retrieve evidence in a fully automated and flexible \textbf{open-domain} manner~\cite{chen2017reading}; no `golden evidence' is pre-identified or curated 
and given to the model. 

To evaluate the claim's veracity, we propose a novel architecture, the \textbf{Consistency-Checking Network (\model{})}, that consists of 1)~memory networks components to evaluate the consistency of the claim against the evidence (described above), 2)~a CLIP~\cite{radford2021learning} component to evaluate the consistency of the \textbf{\textcolor{myOrange}{image}} and \textbf{\textcolor{myblue}{caption}} pair themselves. As the task requires machine comprehension and visual understanding, we perform different evaluations to design the memory components and the evidence representations. Moreover, we conduct two user studies to 1)~measure the human performance on the detection task and, 2)~understand if the collected evidence and the model's attention over the evidence help people distinguish true from falsified pairs.
~\autoref{fig:teaser} depicts our framework, showing a falsified example from the dataset along with the retrieved evidence.

\textbf{Contributions.} We summarize our contributions as follows: 1) we formalize a new task of multi-modal fact-checking. 2) We propose the \textbf{`multi-modal cycle-consistency check'} to gather evidence about the multi-modal claim from both modalities. 3) We propose a new inspectable framework, \textbf{\model{}}, to mimic the aggregation of observations from the claims and world knowledge. 4) We perform numerous evaluations, ablations, and user studies and show that our evidence-augmented method significantly improves the detection over baselines. 

\section{Related Work}
\textbf{Multi-modal Misinformation.}
Previous work has studied multi-modal misinformation~\cite{wang2018eann,khattar2019mvae,nakamura2020fakeddit}. For instance, Khattar et al.~\cite{khattar2019mvae} studied multi-modal fake news on Twitter by learning representations of images and captions which were used in classification. The images in the dataset could be edited. 
In contrast, we focus on out-of-context real news images and verifying them using evidence. 

Moreover, Zlatkova et al.~\cite{zlatkova2019fact} studied the factuality of the image-claim pairs using information from the Web. They collected features about the claim image, such as its URL. The actual content of the claim image is not considered against evidence. Our work is different in how we collect both \textbf{\textcolor{myOrange}{visual}} and \textbf{\textcolor{myblue}{textual}} evidence to perform the \textbf{cycle-consistency} check. In addition, they only calculate features from the claim text such as TF-IDF, while we use memory networks with learned representations. 

Related to the out-of-context threat, Aneja et al.~\cite{aneja2021catching} constructed a large, yet unlabelled, dataset of different contexts of the same image. They propose a self-supervised approach to detect whether two captions (given an image) are having the same context. However, unlike our work, they do not judge the veracity of a single image-caption claim. Also, the unlabelled dataset collected in this work does not allow the veracity detection training and evaluation. 

In order to produce labelled out-of-context images, previous work created synthetic datasets by changing the captions, either by naive swapping or named entities manipulations~\cite{jaiswal2017multimedia,sabir2018deep}, however, the falsified examples were either too naive 
or contained linguistic biases that are easy to detect even by language-only models~\cite{luo2021newsclippings}. 

Therefore, Luo et al.~\cite{luo2021newsclippings} proposed to create falsified examples by matching real images with real captions~\cite{liu2020visualnews}. They created the large-scale NewsCLIPpings dataset that contains both \textit{pristine} and convincing \textit{falsified} examples. The matching was done automatically using trained language and vision models (such as SBERT-WK~\cite{wang2020sbert}, CLIP~\cite{radford2021learning}, or scene embeddings~\cite{zhou2017places}). The falsified examples could misrepresent the context, the place, or people in the image, with inconsistent entities or semantic context. The authors show that both machine and human detection are limited, indicating that the task is indeed challenging. 
Thus, to improve the detection, we propose to use external Web evidence to verify the \textbf{\textcolor{myOrange}{image}}-\textbf{\textcolor{myblue}{caption}} claim.

\textbf{Open-domain QA and Fact-verification.}
Our work is similar to textual work in open-domain QA~\cite{chen2017reading} and fact-verification~\cite{thorne2018fever} (from Wikipedia) in having a large-scale and open-domain task that involves automatic retrieval and comprehension. We do not assume that the input to the model is already labelled and identified as relevant, simulating real-life fact-checking. Moreover, we do not restrict the evidence to be from a specific curated source only, such as fact-checking websites, in contrast to~\cite{vo2020facts}. 

Similar to our work, Popat et al.~\cite{popat2018declare} built a credibility assessment end-to-end method of textual claims using external evidence. However, to the best of our knowledge, no previous work attempted to verify multi-modal claims using both modalities. Also, their model is designed to predict the per-source credibility of claims, while we learn the aggregated consistency from multiple sources.

\section{Dataset and Evidence Collection} \label{sec:dataset}
\textbf{Dataset.} We use the NewsCLIPpings~\cite{luo2021newsclippings} that contains both pristine and falsified (`out-of-context') images. It is built on the VisualNews~\cite{liu2020visualnews} corpus that contains news pieces from 4 news outlets: The Guardian, BBC, USA Today, and The Washington Post. The NewsCLIPpings dataset contains different subsets depending on the method used to match the images with captions (e.g., text-text similarity, image-image similarity, etc.). We use the `balanced' subset that has representatives of all matching methods and consists of 71,072 train, 7,024 validation, and 7,264 test examples. 
To kick-start our \textit{evidence-assisted detection}, we use the \textbf{\textcolor{myOrange}{image}}-\textbf{\textcolor{myblue}{caption}} pairs as queries to perform Web search, as depicted in~\autoref{fig:teaser}. 
%\vspace{-3mm}

\textbf{\textcolor{myblue}{Textual evidence.}} We use the query \textbf{\textcolor{myOrange}{image}} in an inverse search mode using Google Vision APIs~\cite{google3} to retrieve \textbf{\textcolor{myblue}{textual evidence}} \vcenteredinclude{figs/icon3.pdf}. The API returns a list of \textbf{entities} that are associated with that image, which we collect as part of the textual evidence. They might describe the content of the image and, further, the contexts of where these images appeared, such as the entities' list in~\autoref{fig:teaser}.

In addition, the API returns the images' URLs and the containing pages' URLs. In contrast to previous work~\cite{zlatkova2019fact} that only considered the containing pages' titles, we also collect the images' captions. We designed a Web crawler that visits the page, searches for the image's tag using its URL or by image content matching (using perceptual hashing), then retrieves the \textbf{captions} if found. We scrape the \textless$\textit{figcaption}$\textgreater\ tag, as well as the \textless\textit{img}\textgreater\ tag's textual attributes such as \textit{alt}, \textit{image-alt}, \textit{caption}, \textit{data-caption}, and \textit{title}. In addition, we observed the returned pages for a few hundreds of the API calls and implemented other strategies to scrape the captions based on them. We also save the \textbf{titles} of the pages. From each page, we collect all the non-redundant text snippets that we found. The API returns up to 20 search results. We discard a page if the detected language of the title is not English, using the fastText library~\cite{fasttext} for language identification. We collect the \textbf{domains} of each evidence item as metadata.

\textbf{\textcolor{myOrange}{Visual evidence.}} Second, we use the \textbf{\textcolor{myblue}{caption}} as textual queries to search for \textbf{\textcolor{myOrange}{images}} \vcenteredinclude{figs/icon4.pdf}. We use the Google custom search API~\cite{google} to perform the image search. We retrieve up to 10 results, while also saving their \textbf{domains}. It is important to note that, unlike the inverse image search, the search results here are not always corresponding to the exact match of the textual query. Therefore, the \textbf{\textcolor{myOrange}{visual evidence}} might be more loosely related to the query \textbf{\textcolor{myOrange}{image}}. However, even if it is not exactly related to the event, it works as a useful baseline of the type of images that could be associated with that topic. 

%\vspace{-5mm}

\textbf{Dataset decomposition.}
We summarize the dataset components and task as follows:
\definecolor{cosmiclatte}{rgb}{1.0, 0.97, 0.91}
\definecolor{darkchampagne}{rgb}{0.76, 0.7, 0.5}
\definecolor{gray(x11gray)}{rgb}{0.75, 0.75, 0.75}
\definecolor{aliceblue}{rgb}{0.94, 0.97, 1.0}
\definecolor{lightgray}{rgb}{0.93, 0.93, 0.93}
\begin{mdframed}[linecolor=gray(x11gray),backgroundcolor=lightgray,roundcorner=20pt,linewidth=1pt]
\textbf{Dataset.} Unless no search results were found, a single example in the dataset consists of the following:
\begin{itemize}[noitemsep,topsep=0pt]
\item A query \textbf{\textcolor{myOrange}{image}} $I^q$.
\item A query \textbf{\textcolor{myblue}{caption}} $C^q$.
\item \textbf{\textcolor{myOrange}{Visual evidence}}: 
    \begin{itemize}[noitemsep,topsep=0pt] 
    \item A list of \textbf{images}: $I^e = [I^e_1, ..., I^e_K]$.
    \end{itemize}
\item \textbf{\textcolor{myblue}{Textual evidence}}:
    \begin{itemize}[noitemsep,topsep=0pt]
        \item A list of \textbf{entities}: $\textit{ENT} = [\textit{E}_1, ..., \textit{E}_M]$. 
        \item A list of \textbf{captions/sentences}:
        
        $S = [S_1, ..., S_N]$.
    \end{itemize}
\end{itemize}
\textbf{Task.} Classify $\{I^q,C^q\}$ to: $\textit{Pristine}$ or $\textit{Falsified}$.
\end{mdframed}

\section{The Consistency-Checking Network}

We introduce the task of evidence-assisted veracity assessment of \textbf{\textcolor{myOrange}{image}}-\textbf{\textcolor{myblue}{caption}} pairing. As shown in~\autoref{fig:teaser}, we perform the \textbf{`multi-modal cycle-consistency check'} by comparing the \textbf{\textcolor{myblue}{textual evidence}} against the query \textbf{\textcolor{myblue}{caption}}, and the \textbf{\textcolor{myOrange}{visual evidence}} against the query \textbf{\textcolor{myOrange}{image}}.

\textbf{Challenges.} The task is significantly more complex than the merely one-to-one matching of the query against the evidence. First, many search results may be  unrelated to the query (neither falsify nor support) and act as noise. Second, comparing the query against the evidence requires further comprehension and reasoning. For pristine examples, the \textbf{\textcolor{myblue}{textual evidence}} might range from being paraphrases of the query \textbf{\textcolor{myblue}{caption}} to distantly related but supporting. For falsified examples, they might range from having different named entities to having the same ones but in a different context, such as the example in~\autoref{fig:teaser}. Similarly, comparing the \textbf{\textcolor{myOrange}{visual evidence}} against the query \textbf{\textcolor{myOrange}{image}} requires visual and scene understanding or regions comparison.
\begin{figure}[!t]
\centering
\includegraphics[width=0.8\linewidth]{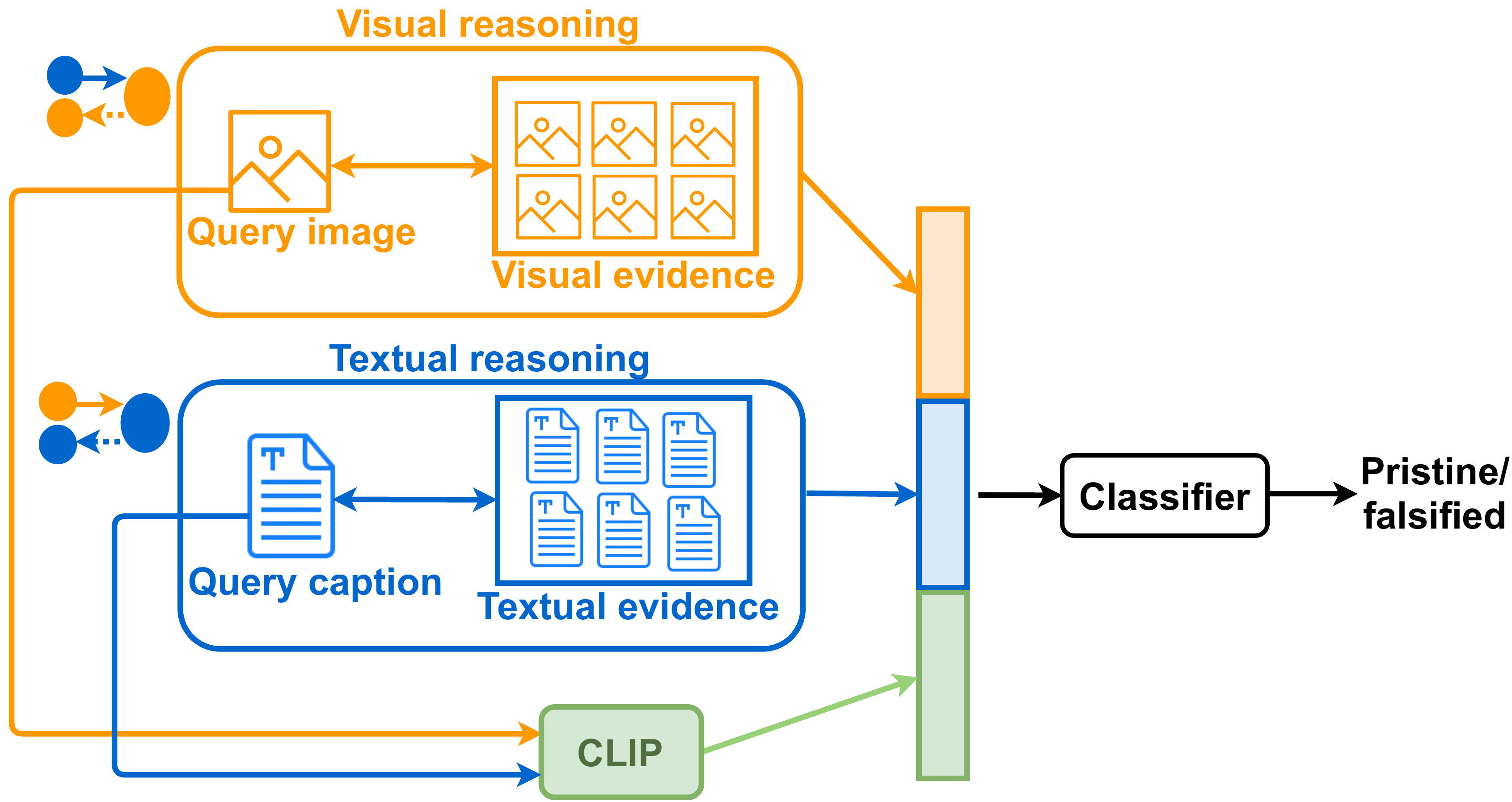}
\caption{Overview of our Consistency-Checking Network, \model{}.} 
\label{fig:method_small}
\vspace{-3mm}
\end{figure}

We propose a novel architecture, the Consistency-Checking Network (\model{}), to meet these challenges. We show an overview of the method in~\autoref{fig:method_small}. At the core of our approach is the memory networks architecture~\cite{sukhbaatar2015end,kumar2016ask,mohtarami2018automatic,chunseong2017attend}, which selectively compares the claim to the relevant items of the possibly large list of evidence. In addition, the attention mechanism allows inspecting which evidence items were most relevant to the decision. The model consists of a \textbf{\textcolor{myOrange}{visual reasoning}} component, a \textbf{\textcolor{myblue}{textual reasoning}} component, and a `CLIP' component. 
\begin{figure}[!b]
\vspace{-2mm}
\centering
\includegraphics[width=0.85\linewidth]{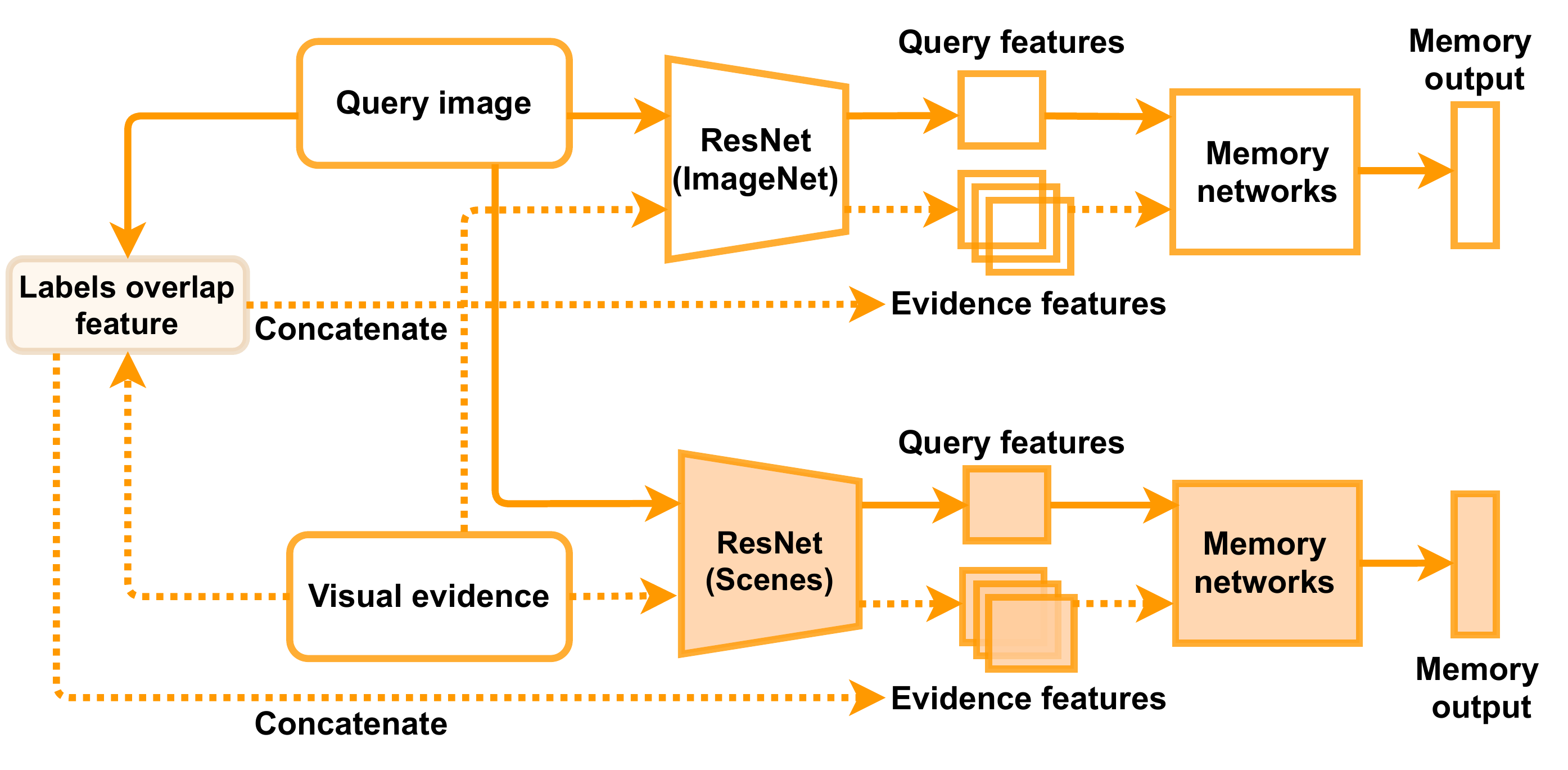}
\caption{Visual evidence reasoning component.} 
\label{fig:vis_mem}
\end{figure}

\subsection{Visual Reasoning}
\autoref{fig:vis_mem} outlines the visual reasoning component that inspects the consistency between the query \textbf{\textcolor{myOrange}{image}} and the \textbf{\textcolor{myOrange}{visual evidence}}. First, we represent the images (query and evidence) using ResNet152~\cite{he2016deep}, pretrained on the ImageNet dataset. Each image is represented as: $I^q/I^e\in\mathbb{R}^{2048}$, where $q$ denotes the query representation and $e$ denotes evidence. 
Moreover, to reason over the overlap of regions and objects in the query image vs. evidence images, we used the label detection Google API~\cite{google2} to get a list of labels for each image. Then, for each evidence image, we compute the number of \textit{overlapping labels} between it and the query. We use this number as an additional feature, and we concatenate it with the evidence images' representations. 

The memory holds the evidence images. Each input to the memory is embedded into input and output memory representations~\cite{sukhbaatar2015end}, denoted by $a$ and $c$, respectively. The image memory vectors $m_i\in\mathbb{R}^{1024}$ are represented by:
\begin{equation}
m_i^a = \text{ReLU}(W_i^aI^e+b_i^a),
\end{equation}
\begin{equation}
m_i^c = \text{ReLU}(W_i^cI^e+b_i^c)
\end{equation}
The learned parameters are $W_i^a$ and $W_i^c$ $\in\mathbb{R}^{2048\times1024}$, and $b_i^a$ and $b_i^c$ $\in\mathbb{R}^{1024}$. The query image $I^q$ is also projected into a $1024$-dimension vector ($\hat{I^q}$) by another linear layer for modelling convenience. The matching between $\hat{I^q}$ and the memory vectors $m_i^a$ are then computed by: 
\begin{equation} \label{eqn:softmax}
{p_i}_j = \text{Softmax}(\hat{I^q}^T{m_i^a}_j),
\end{equation}
where $i$ denotes the image memory, $j$ is a counter for the memory items, and $p_i$ is a probability vector over the items. The output of the memory is the sum of the query and the average of the output representations $m_i^c$, weighted by $p_i$: 
\begin{equation} \label{eqn:output}
o_i = \sum_{j}{p_i}_j{m_i^c}_j + \hat{I^q}
\end{equation}
In addition, for some mismatched examples, there could be context discrepancies based on the place. To make the model aware of scenes and places similarity, we also represent the images using a ResNet50 trained on the Places365 dataset~\cite{zhou2017places}. We form a separate memory for the scene representations to allow more flexibility. Similar to the previous formulation, each image is represented as: $P^q/P^e\in\mathbb{R}^{2048}$, and the scenes memory vectors $m_p\in\mathbb{R}^{1024}$ are represented by:
\begin{equation}
m_p^{a/c} = \text{ReLU}(W_p^{a/c}P^e+b_p^{a/c})
\end{equation}
Similar to Eqn.~\ref{eqn:softmax} and Eqn.~\ref{eqn:output}, we get the output of the scenes (places) memory $o_p$.
\subsection{Textual Reasoning}
The second component of our model evaluates the consistency between the query \textbf{\textcolor{myblue}{caption}} and the \textbf{\textcolor{myblue}{textual evidence}}. As shown in~\autoref{fig:teaser}, we have two types of textual evidence: sentences (captions or pages' titles), and entities. As they have different granularities and might differ in importance, we form a separate memory for each. 

\begin{figure}[!t]
\centering
\includegraphics[width=0.9\linewidth]{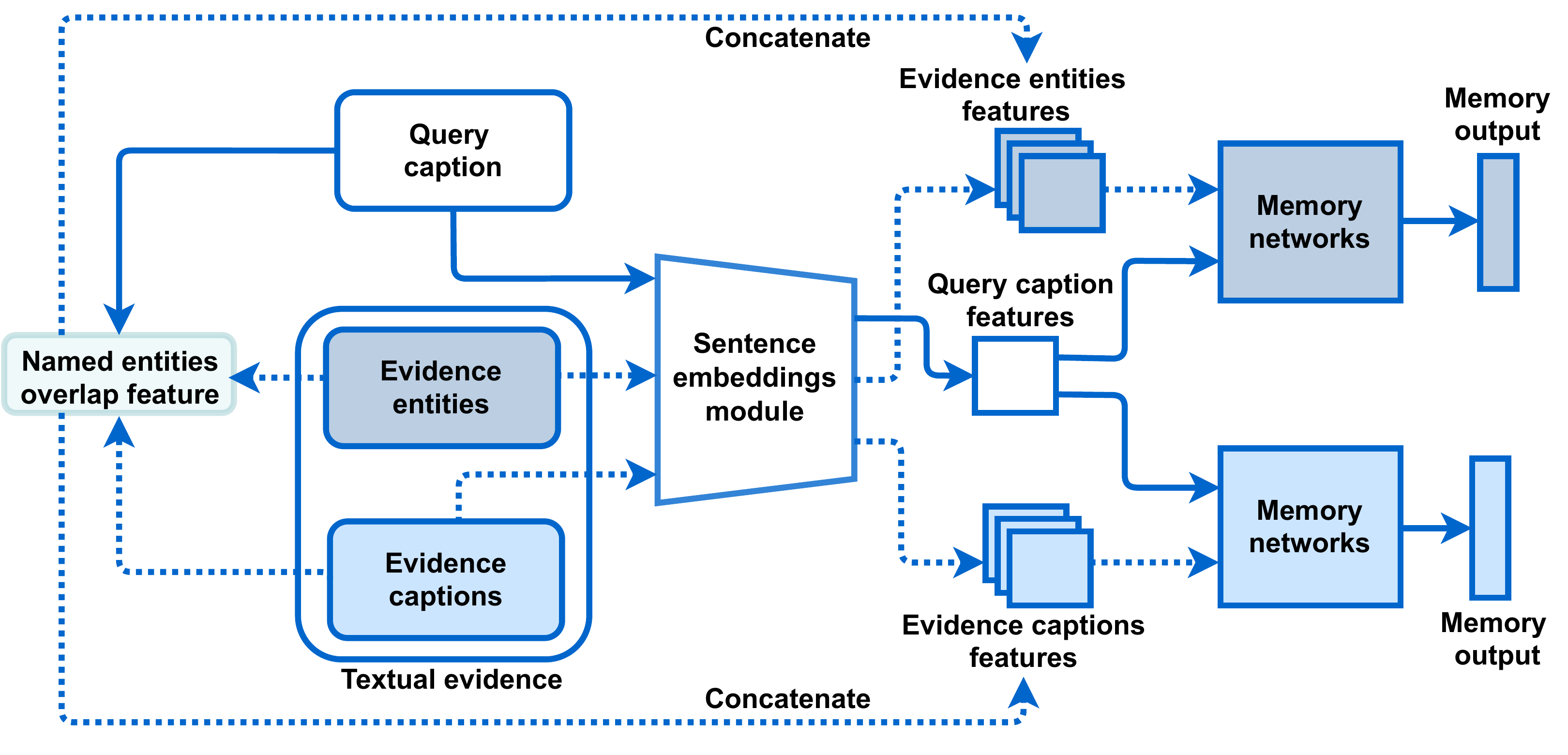}
\caption{Textual evidence reasoning component.} 
\label{fig:text_mem}
\vspace{-2mm}
\end{figure}

As shown in~\autoref{fig:text_mem}, we represent the query caption and each evidence item using a sentence embedding model. 
We experiment with state-of-the-art inference models that were trained on large corpuses such as Wikipedia and were shown to implicitly store world knowledge~\cite{petroni2019language,lee2020language,roberts2020much}, making them suitable for our task. We evaluate two methods in our experiments: 1) a pre-trained sentence transformer model~\cite{reimers-2019-sentence-bert} that is trained for sentence similarity, 2) using BERT~\cite{devlin2019bert} to get strong contextualized embeddings, in addition to an LSTM to encode the sequence. In the second method, we use the second-to-last BERT layer~\cite{zellers2019recognition} as the tokens' embeddings. We concatenate the last time-step LSTM output and the average of all time-steps' outputs.

In addition, to help the model be entity-aware, we utilize a binary indicator feature to denote if there is a named entity overlap between the query caption and the evidence item. We used the spaCy NER~\cite{honnibal2017spacy} to extract the entities and concatenated the binary feature with the evidence (both captions and entities) representations. 

Using either of these previously mentioned methods, we get embeddings for the query caption $C^q$, the evidence entities $E$, and the evidence captions/sentences $S$. The entities input and output memory representations are given by: 
\begin{equation}
m_e^{a/c} = \text{ReLU}(W_e^{a/c}E+b_e^{a/c}),
\end{equation}
similarly, the captions/sentences input and output memory representations are given by: 
\begin{equation}
m_s^{a/c} = \text{ReLU}(W_s^{a/c}S+b_s^{a/c}),
\end{equation}
where $W_e^{a/c},W_s^{a/c}\in\mathbb{R}^{d\times d}$ and $b_e^{a/c},b_s^{a/c}\in\mathbb{R}^{d}$ are trainable weights, and $d$ is the dimension of the sentence embedding model (768 in the case of the pre-trained model, and 512 in the case of using BERT+LSTM).

As per Eqn.~\ref{eqn:softmax} and Eqn.~\ref{eqn:output}, we compute the output of the entities and sentences memories as $o_e$ and $o_s$, respectively. 
%\vspace{-3mm}

\textbf{Encoding the evidence's domain.} 
Features of websites, e.g., how frequently they appear and the types of news they feature, could help to prioritize evidence items. 
Thus, we learn an embedding of the evidence's domain names. We represent the domains as one-hot vectors and project them into a 20-dimensional space. We consider the domains that appeared at least three times, resulting in 17148 unique domains, the rest are set to $\textit{UNK}$. The domain embeddings are then concatenated with the evidence representations (both visual and textual, excluding entities).

\subsection{CLIP}
In addition to reasoning over evidence, we leverage CLIP~\cite{radford2021learning}, used in~\cite{luo2021newsclippings}, to integrate the query \textbf{\textcolor{myOrange}{image}}-\textbf{\textcolor{myblue}{text}} consistency into the decision. We first fine-tune CLIP ViT/B-32 on the task of classifying image-caption pairs into pristine or falsified, without considering the evidence. 

During fine-tuning, we pass the image and text through the CLIP encoders and normalize their embeddings. We produce a joint embedding that is a dot product of the image and text ones, and we add a linear classifier on top. The model is trained to classify the pair into pristine or falsified. Then, we freeze the fined-tuned CLIP and integrate the joint CLIP embeddings ($J_\text{clip}$) into the final classifier of \model{}.  

\subsection{Classifier}
Now that we individually evaluated the \textbf{\textcolor{myblue}{text}}-\textbf{\textcolor{myblue}{text}}, \textbf{\textcolor{myOrange}{image}}-\textbf{\textcolor{myOrange}{image}}, and \textbf{\textcolor{myOrange}{image}}-\textbf{\textcolor{myblue}{text}} consistency, we aggregate these observations in order to reach a unified decision. 

We found it helpful during training to apply a batch normalization layer~\cite{ioffe2015batch} to the output of each component. We then concatenate all previous components in one feature vector $o_t$ as follows:
\begin{equation} \label{eqn:bn}
    o_t = \text{BN}(o_i)\oplus\text{BN}(o_p)\oplus\text{BN}(o_e)\oplus\text{BN}(o_s)\oplus\text{BN}(J_\text{clip}), 
\end{equation}
where $\text{BN}$ denotes the batch normalization. $o_t$ is then fed to a simple classifier that has two fully connected layers with ReLU and batch normalization after the first one (dimension: 1024), and Sigmoid after the second one that outputs a final falsified probability ($p_f$). The model is trained, with freezing the backbone embedding networks, to binary classify the examples using the binary cross-entropy loss: 
\begin{equation}
L = -y_\text{true}\log(p_f) - (1-y_\text{true})\log(1-p_f)
\end{equation}
More implementation details can be found in Supp.~\ref{sec:implem}.
\section{Experimental Results}
In this section, we show the quantitative analysis of different variants of the model and baselines. We then present our user studies, qualitative analysis, and discussion. 

\newcommand{\hlc}[2][yellow]{{\sethlcolor{#1}\hl{#2}}}
\definecolor{light_red}{rgb}{0.96, 0.76, 0.76}
\definecolor{light_green}{rgb}{0.56, 0.93, 0.56}
\definecolor{light_yellow}{rgb}{0.99, 0.97, 0.37}

\newcommand*\rot{\rotatebox{90}}
	
\begin{table} [!b]
%\vspace{-1mm}
\centering
\resizebox{0.88\linewidth}{!}{%
\begin{tabular}{c c c c c c | c c c |c c c ||c}
\toprule
\# & \rot{Evidence type} & \rot{Separate mem.} & \rot{BN} & \rot{Dataset filter} & \rot{CLIP} & \rot{ResNet (ImageNet)} & \rot{ResNet (Scenes)} & \rot{Labels} & \rot{Sent. transformer} & \rot{BERT+LSTM}  &  \rot{NER} & \textbf{\rot{Accuracy}} \\ \hline
1 & all & \cmark & \xmark & \xmark & \xmark & \cmark & \xmark & \xmark &  \cmark & \xmark & \xmark & 73.5\% \\  [0.1cm]

2 & \textbf{\underline{\hlc[light_yellow]{all w/o Images}}} & \cmark  & \xmark & \xmark & \xmark  & - & - & - &  \cmark & \xmark & \xmark & 62.5\% \\    [0.1cm]

3 & \textbf{\underline{\hlc[light_yellow]{all w/o Captions}}} & \cmark & \xmark & \xmark & \xmark  & \cmark  & \xmark & \xmark &  \cmark & \xmark & \xmark & 57.4\% \\   [0.1cm]
4 & \textbf{\underline{\hlc[light_yellow]{all w/o Entities}}} & \cmark & \xmark & \xmark & \xmark  & \cmark  & \xmark & \xmark &  \cmark & \xmark & \xmark & 71.8\% \\  [0.1cm]

5 & all & \cmark & \textbf{\underline{\hlc[light_yellow]{\cmark}}} & \xmark & \xmark & \cmark  & \xmark & \xmark &  \cmark & \xmark & \xmark & 84.2\% \\  [0.1cm]
6 & all & \textbf{\underline{\hlc[light_yellow]{\xmark}}} & \cmark & \xmark & \xmark  & \cmark  & \xmark & \xmark &  \cmark & \xmark & \xmark & 81.7\% \\ [0.1cm]

7 & all & \cmark & \cmark & \textbf{\underline{\hlc[light_yellow]{\cmark}}} & \xmark  & \cmark  & \xmark & \xmark &  \cmark & \xmark & \xmark & 80.3\% \\  [0.1cm]

8 & all & \cmark &  \cmark & \cmark & \xmark  & \cmark  & \xmark & \xmark &  \cmark & \xmark & \textbf{\underline{\hlc[light_yellow]{\cmark}}} & 81.2\% \\  [0.1cm]

9 & all & \cmark &  \cmark & \cmark & \textbf{\underline{\hlc[light_yellow]{\cmark}}}  & \cmark  & \xmark & \xmark &  \cmark & \xmark & \cmark & 82.6\% \\  [0.1cm]

10 & all & \cmark &  \cmark & \cmark & \cmark  & \cmark  & \textbf{\underline{\hlc[light_yellow]{\cmark}}} & \xmark &  \cmark & \xmark & \cmark & 83.4\% \\  [0.1cm]

11 & all & \cmark &  \cmark & \cmark & \cmark   & \cmark & \cmark & \textbf{\underline{\hlc[light_yellow]{\cmark}}} &  \cmark & \xmark & \cmark & 83.9\% \\  [0.1cm]

12 & \boxit{3.8in} all & \cmark &  \cmark & \cmark & \cmark  &  \cmark & \cmark & \cmark &  \xmark & \textbf{\underline{\hlc[light_yellow]{\cmark}}} & \cmark & \textbf{84.7\%} \\  [0.1cm]

13 & \textbf{\underline{\hlc[light_yellow]{all w/o domains}}} &\cmark &  \cmark & \cmark & \cmark  &  \cmark & \cmark & \cmark &  \xmark & \cmark & \cmark & 83.9\% \\  [0.1cm]
\bottomrule
\end{tabular}}
\caption{Classification performance on the test set for different variants of the model. Highlighted cells represent the changed factor in that experiment. The green box represents the best model.} \label{table:ablation}
\end{table}
\subsection{Quantitive Analysis}
We evaluated our model and other variants of it in order to understand the effect of each component.~\autoref{table:ablation} shows our experiments. We summarize different aspects and highlight the most interesting observations in what follows.

\textbf{Evidence types.} We first show the effect of each evidence type in the first four rows. Removing the evidence \textbf{\textcolor{myOrange}{images}} or the evidence \textbf{\textcolor{myblue}{captions}} dropped the performance significantly; these results indicate the importance of integrating both modalities for verification. Removing the \textbf{\textcolor{myblue}{Entities}} had less effect. This might be due to having some redundant information with the evidence captions already, or because of sometimes having generic named entities that are not helpful to verify the caption claim. 

\textbf{Memory design.} Adding a batch normalization layer after each component, as in Eqn.~\ref{eqn:bn}, improved the training and increased the accuracy by nearly 11 percentage points. Another variant we studied had a unified memory containing images, captions, and entities. The query here was a concatenation of the image and caption pairs. As shown in row 6, this was less successful than the separate memory setup, suggesting that the explicit \textbf{\textcolor{myblue}{text}}-\textbf{\textcolor{myblue}{text}} and \textbf{\textcolor{myOrange}{image}}-\textbf{\textcolor{myOrange}{image}} consistency comparison aids the learning. 

\textbf{Evidence filtering.} As the dataset is constructed from real news articles, the Google search may return the exact news as the query search (i.e., exact news with the exact webpage). While this is needed in a real fact-checking setup, it might bias the training; the model might use it/or its absence as a shortcut to predict pristine/falsified pairs, respectively, without stronger reasoning. Therefore, we filtered the evidence as follows: for pristine examples, we discard an evidence item if it \textit{matches} the query and comes from the \textit{same website} as the query. To detect matching, we use perceptual hashing for \textbf{\textcolor{myOrange}{images}}. For \textbf{\textcolor{myblue}{captions}}, we remove punctuations and lower-case all the sentences and then check if they are an exact match. We then trained and evaluated with this filtered dataset. As shown in row 7, this did not significantly reduce the accuracy, suggesting that the model reasons about consistency beyond exact matches.

\textbf{Other improvements.} We show that our other enhancements, including adding CLIP and improving visual and textual representations, recovered the performance drop due to the evidence filtering. CLIP had relatively the largest effect, with around a 1.5 percentage points increase. Training the LSTM with BERT embeddings performed better than using a pre-trained sentence transformer model. This might be because it allowed the model to learn on the token level and focus on the consistency in, e.g., named entities, location, etc., which are more specific cues in our use-case than general sentence entailment tasks. Finally, the last row shows that including the evidence's domain 
helps to some extent, as it might help the model to attend to and prioritize evidence items. Additional experiments are in Supp.~\ref{sec:add_exp}.

\textbf{Baselines.}
We compare our evidence-assisted detection against the CLIP-only baseline used in~\cite{luo2021newsclippings} in~\autoref{table:baseline}. We fine-tuned CLIP~\cite{radford2021learning}, reaching a higher accuracy than originally reported in~\cite{luo2021newsclippings} on this dataset subset. As the dataset pairing is not trivial, this baseline achieved a relatively low performance. 
In contrast, we achieve a significant improvement of a nearly 19 percentage points increase, indicating that leveraging evidence is important to solve the task. 
\begin{table} [!t]
\centering
\resizebox{0.75\linewidth}{!}{%
\begin{tabular}{c| c c| c c c} 
\toprule
\textbf{Method} & \textbf{Evidence} & \textbf{Pair} & \textbf{All} & \textbf{Falsified} & \textbf{Pristine}  \\ \midrule
CLIP & \xmark & \cmark & 66.1\% & 68.1\% & 64.2\% \\
Averaged & \cmark & \xmark & 70.6\% & 72.4\% & 68.9\% \\ 
\textbf{\model{}} & \cmark & \cmark & \textbf{84.7\%} & \textbf{84.8\%} & \textbf{84.5\%} \\ 
\bottomrule
\end{tabular}}
\caption{Classification performance on the test set for our model in comparison with baselines.
} \label{table:baseline}
\vspace{-2mm}
\end{table}

As there are no previous baselines for evidence-assisted out-of-context detection, we design a baseline that uses evidence. We use the pretrained image and text representations of ResNet-152 and sentence transformer in the same setup of \textbf{\textcolor{myblue}{text}}-\textbf{\textcolor{myblue}{text}} and \textbf{\textcolor{myOrange}{image}}-\textbf{\textcolor{myOrange}{image}} similarity. We compute the matching between the query and the evidence via dot product. Then, we use an average pooling layer across all evidence items, which will be used for classification. As shown in~\autoref{table:baseline}, this baseline outperforms the CLIP-only. However, our proposed model with the other improvements achieves a $\sim$14 percentage points increase.
\subsection{User Studies}
%\textbf{Motivation.} 
We conducted user studies to estimate the human performance on the dataset and evaluate the usefulness of the evidence in detection, as well as the relevancy of the evidence items that the model highly attends to.
\vspace{-2mm}
\subsubsection{Study 1: Human Performance Baseline}
We aim to establish a human baseline as an upper bound estimate of the out-of-context images detection accuracy. Due to the automatic open-world evidence retrieval, we do not have a labelled dataset to indicate if an evidence item is relevant to the claim. Furthermore, some examples might not have any relevant evidence retrieved. Also, the falsified examples could be very close to the original context, making them hard to verify even with the presence of evidence.

\textbf{Setup.} We randomly selected 100 examples (48 pristine, 52 falsified) from the test dataset. Along with the \textbf{\textcolor{myOrange}{image}}-\textbf{\textcolor{myblue}{caption}} pairs, we presented the gathered evidence (\textbf{\textcolor{myOrange}{images}}, \textbf{\textcolor{myblue}{captions}}, and \textbf{\textcolor{myblue}{entities}}). For each pair, first, we asked users if the \textbf{\textcolor{myblue}{caption}} matches the \textbf{\textcolor{myOrange}{image}}, considering any of: inconsistency cues between them, the evidence presented, or their prior knowledge about the subject. Then, they answered which source(s) of information helped them label the pair, or indicated `None' if it was hard to verify. 
We instructed them \textit{not to} search for other evidence, so that both our model and humans have access to the same evidence, and to evaluate the usefulness of the evidence gathered by our framework. We recruited 8 experienced native English-speaking crowd workers through Amazon Mechanical Turk.  

\textbf{Results.}
\autoref{table:user_study} shows the average performance across all workers and the results of the best worker.   
Compared to the findings reported in~\cite{luo2021newsclippings}, human performance significantly increased when presented with evidence (average detection was 65.6\%, with only 35\% falsified detection rate). Additionally, \model{} achieved 80\% accuracy on these 100 examples, which is lower than the best worker but on a par with the average worker. 

\begin{table} [!t]

\centering
\resizebox{0.8\linewidth}{!}{%
\begin{tabular}{c|cccc} \toprule
& \textbf{Study} & \textbf{All} & \textbf{Falsified} & \textbf{Pristine}   \\ \midrule
\multirow{3}{*}{\textbf{Average}} & \nth{1} & 81.0\%$\pm$4.71 & 79.5\%$\pm$8.31 & 82.3\%$\pm$9.31 \\ \cline{2-5}
 & \nth{2}, Highest & 86.2\%$\pm$4.9 & 84.5\%$\pm$9.3 & 88.0\%$\pm$7.2 \\ 
 & \nth{2}, Lowest  & 77.7\%$\pm$6.0 &  76.0\%$\pm$9.0 &  79.5\%$\pm$7.5 \\
 \midrule 
\multirow{3}{*}{\textbf{Best worker}} & \nth{1} & 89.0\%  & 92.0\% & 93.7\% \\ \cline{2-5}
&\nth{2}, Highest & 94.0\%  & 98.0\% & 98.0\% \\ 
& \nth{2}, Lowest & 88.0\%  & 90.0\% & 86.0\% \\ \bottomrule
\end{tabular}}
\caption{The results of our two user studies. The first is to label 100 examples, selected randomly. 
The second is to label another 100 examples (that have enough evidence retrieved) 
using 1) the highest-attention, and 2) the lowest-attention evidence.
} \label{table:user_study}
\vspace{-2mm}
\end{table}
\begin{figure}[!b]
    \centering
    \includegraphics[width=0.71\linewidth]{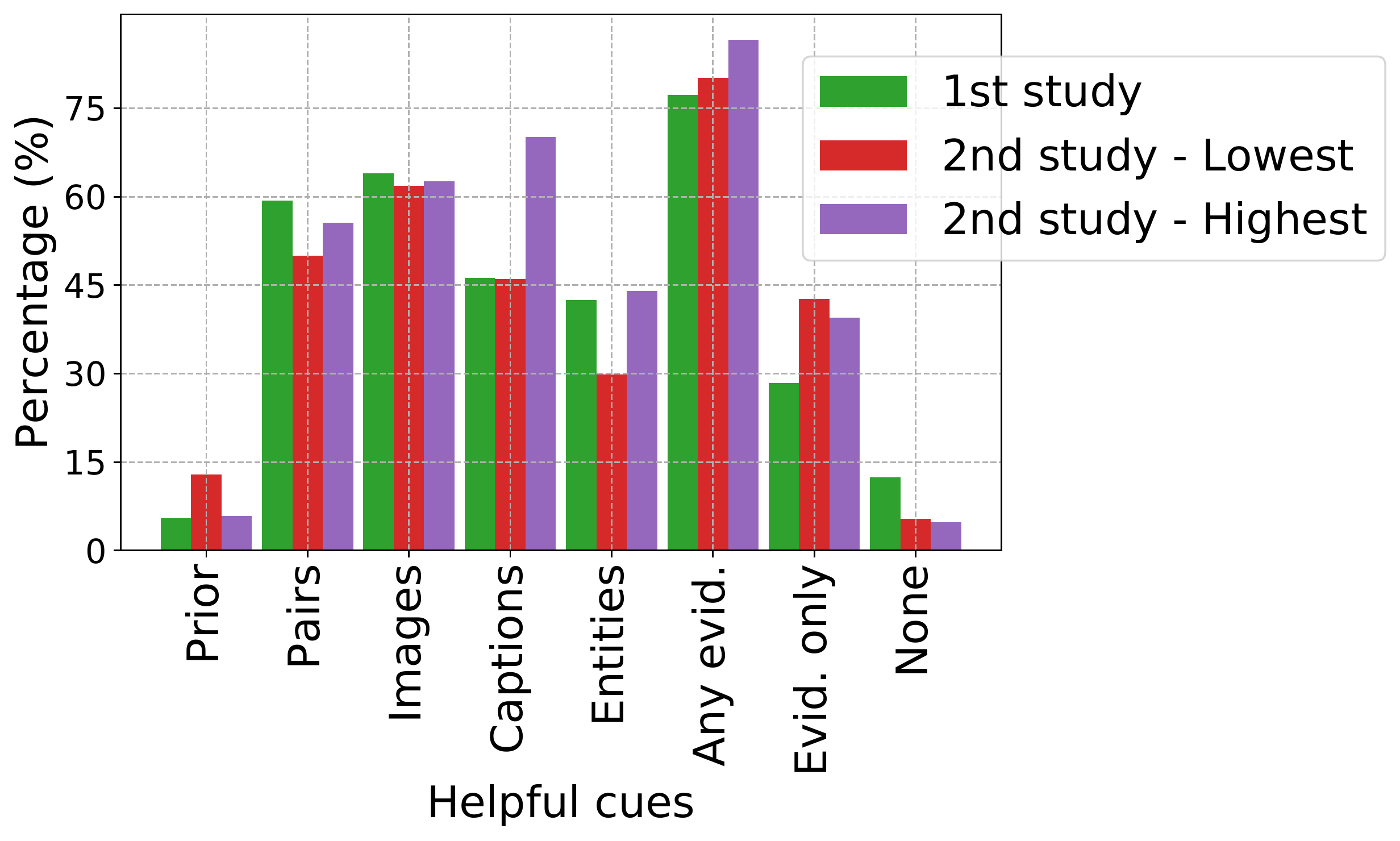}
    \caption{ 
    Workers indicated the factors that helped their decision. %(prior knowledge, the image-caption pair, different types of evidence, or none). 
    `Any evid.' means that any evidence type was helpful. `Evid. only' means that only the evidence was helpful.}
    \label{fig:study}

\end{figure}
~\autoref{fig:study} shows which information helped workers to label the \textbf{\textcolor{myOrange}{image}}-\textbf{\textcolor{myblue}{caption}} pairs during the study. We highlight the following observations: 1) In 77.2\% of the examples, on average, the evidence contributed to the workers' decision, in comparison with 59.3\% only for the \textbf{\textcolor{myOrange}{image}}-\textbf{\textcolor{myblue}{caption}} pair. In 28.3\%, the evidence was the only helpful cue. 2) Among the evidence types, the \textbf{\textcolor{myOrange}{images}} were the most helpful (64\%), possibly because it is easier to grasp different images at a glance. 3) 12.3\% of the examples were hard to verify. When checking some of them (Supp.~\ref{sec:user_study_ex}), we observed that they do not have obvious cues (e.g., generic scenes with event-specific captions, an image for the same person with a similar context). Also, they sometimes had poor retrieval (the inverse search did not find the \textbf{\textcolor{myOrange}{image}}, so there are no evidence \textbf{\textcolor{myblue}{captions}}, and the evidence \textbf{\textcolor{myOrange}{images}} are unrelated or not conclusive). Our model struggled in detecting these examples as well. Augmenting with looser retrieval (e.g., searching with keywords of the caption, finding captions of other similar images) might help in these cases.
\vspace{-2mm}
\subsubsection{Study 2: Evaluating the Attention}
One of our main goals is to have an automated fact-checking tool while also allowing humans to be in the loop, if needed. We hypothesize that the attention weights given by the model can be used to retrieve the most relevant and useful evidence, which enables a quick inspection.

We design a second study to evaluate this hypothesis. We randomly selected 100 examples (50 each) that at least have 8 evidence items in each type\footnote{In this first study, some examples might not have enough evidence. However, we keep them to have a representative set of the dataset.}. We designed two variants using the same 100 pairs; in the first, we display the highest-attention 4 items from each evidence type, in the second, we display the lowest-attention 4 ones. The two variants are labelled by non-overlapping groups (8 workers each). We follow the rest of the first study's setup and instructions.

\textbf{Results.}~\autoref{table:user_study} and~\autoref{fig:study} show that the highest-attention evidence had higher performance and generally better ratings as `helpful' compared to the lowest-attention evidence. These findings suggest that the model learned to prioritize the most relevant items, as intended, and can potentially be beneficial for 1) inspectability and, 2) assistive fact-checking; as workers had a higher performance with only a subset of evidence. 

\fboxsep=1mm%padding thickness
\fboxrule=2pt%border thickness
\definecolor{amaranth}{rgb}{0.9, 0.17, 0.31}
\definecolor{ao(english)}{rgb}{0.0, 0.5, 0.0}
\definecolor{lightgreen}{rgb}{0.83,0.9, 0.83}
\definecolor{lightred}{rgb}{0.97,0.80,0.80}
\definecolor{darkred}{rgb}{1.0,0.20,0.20}

\newcommand{\largericon}[1]{\begingroup
\setbox0=\hbox{\includegraphics[scale=0.27]{#1}}%
\parbox{\wd0}{\box0}\endgroup}
\begin{table*}[!t]
\centering
\resizebox{0.90\linewidth}{!}{%
\begin{tabular}{c|c c}
\toprule
\textbf{\textcolor{myOrange}{\large{Image}}}-\textbf{\textcolor{myblue}{\large{caption}}} \large{pair} & \large{\textbf{\textcolor{myblue}{Textual evidence}}} \largericon{figs/icon3.pdf} & \large{\textbf{\textcolor{myOrange}{Visual evidence}}} \largericon{figs/icon4.pdf} \\ \midrule
\makecell{\fcolorbox{ao(english)}{lightgreen}{
\begin{varwidth}{\textwidth} \begin{center}\fcolorbox{myOrange}{white}{\includegraphics[width=3cm,keepaspectratio]{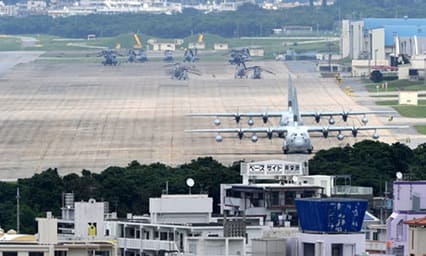}}\end{center} 
\fcolorbox{myblue}{white}{\begin{varwidth}{\textwidth}\normalsize{The Futenma marine corps\\airbase on the southern\\Japanese island of Okinawa}\end{varwidth} }\end{varwidth}}} & 

\makecell{\fcolorbox{myblue}{white}{\begin{varwidth}{\textwidth} \normalsize{`United States', `Ginowan',\\`Governor', \hlc[light_yellow]{`Military base'},\\`Politics', `Japan', `Takeshi Onaga',\\ \hlc[light_yellow]{`Governor of Okinawa Prefecture'},\\ ,`Hirokazu Nakaima',`Shinzo Abe',\\\hlc[light_yellow]{`Okinawa'}, \hlc[light_yellow]{`airport'}} \end{varwidth} }   
\fcolorbox{myblue}{white}{\begin{varwidth}{\textwidth} \normalsize{\hlc[light_yellow]{1- Hercules aircraft parked on the tarmac}\\\hlc[light_yellow]{at Marine Corps Air Station Futenma}\\\hlc[light_yellow]{in Ginowan on Okinawa.}\\2- Japan Decides to Stop Works on US\\Airbase Relocation in Okinawa.\\3- Japan Decides to Restart Relocation \\of US Base in Okinawa Despite Protests.} \end{varwidth} }}
& 
\makecell{ \fcolorbox{myOrange}{light_yellow}{\includegraphics[width=3.7cm,keepaspectratio]{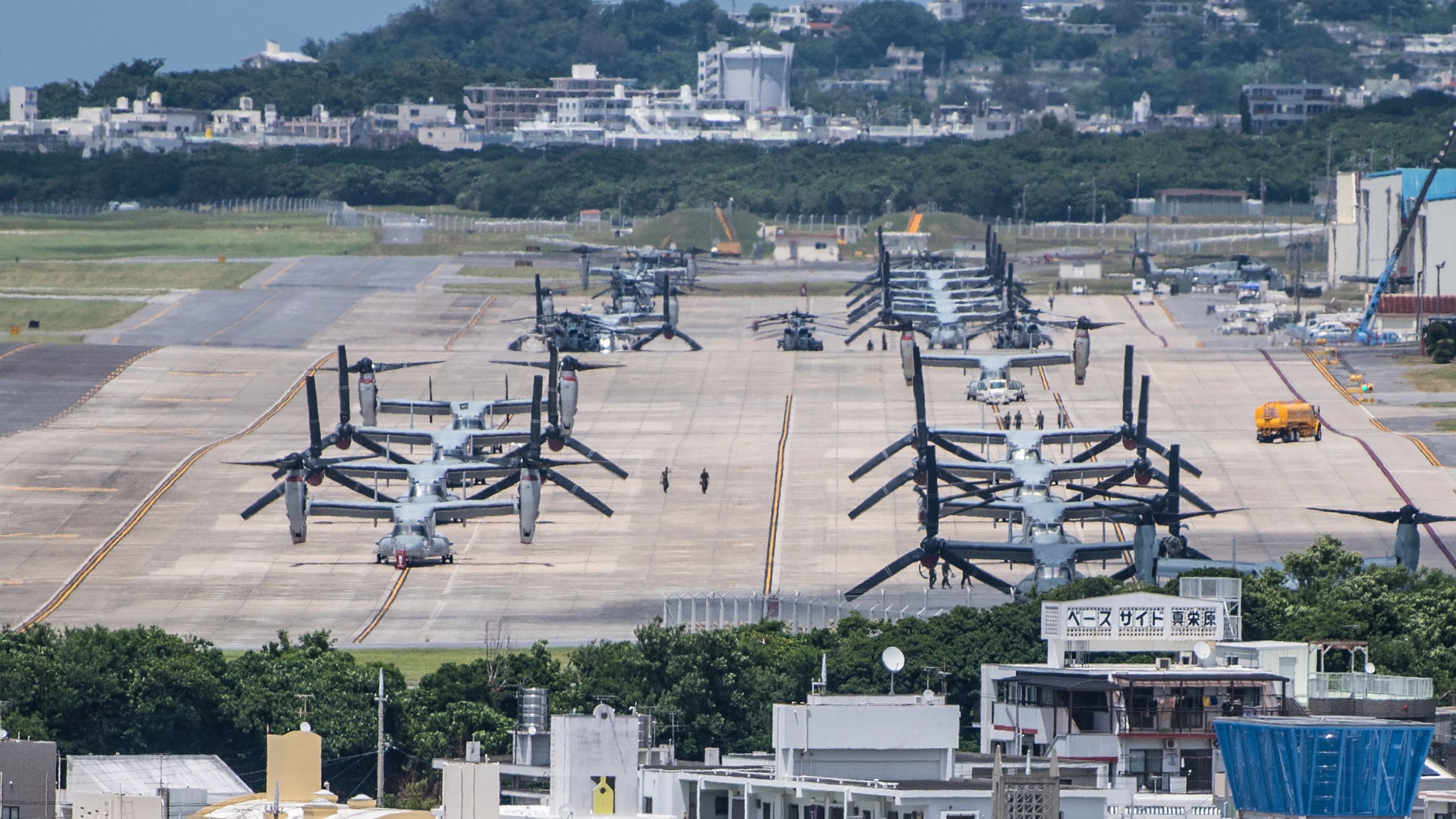}} \fcolorbox{myOrange}{white}{\includegraphics[width=3.3cm,keepaspectratio]{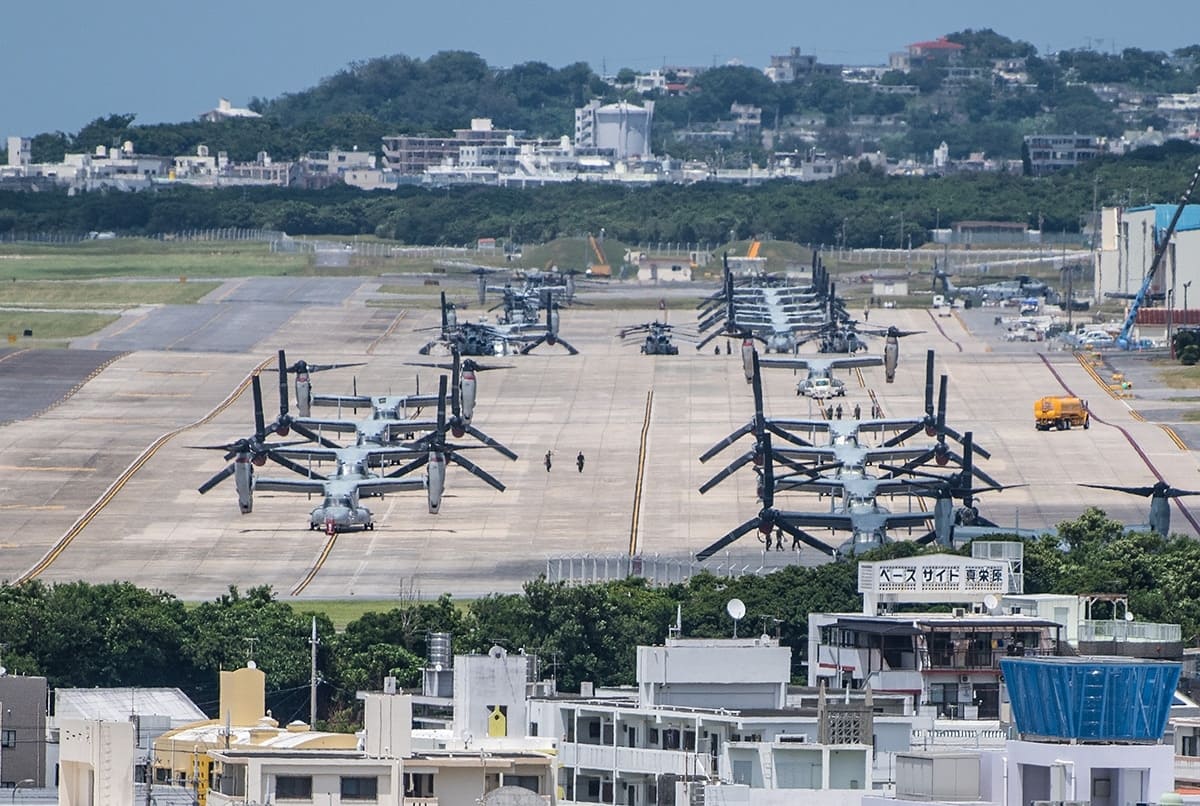}}
\fcolorbox{myOrange}{white}{\includegraphics[width=3.3cm,keepaspectratio]{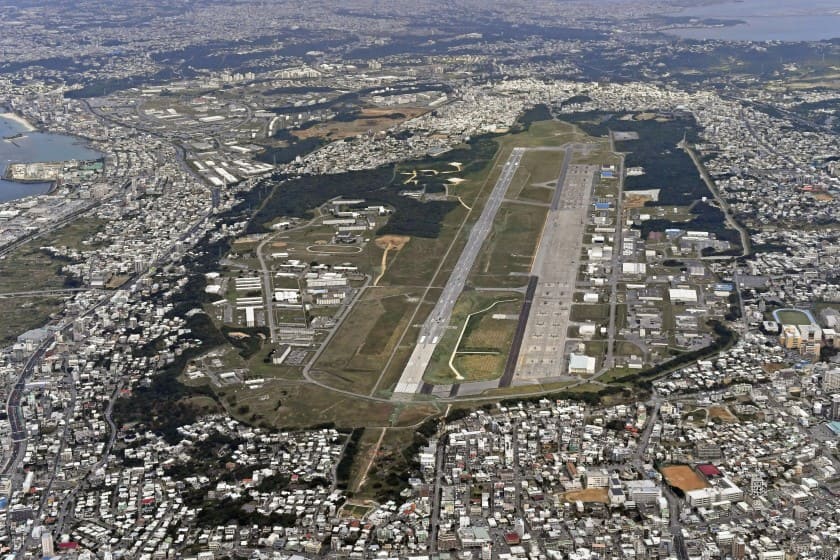}}} 
\\ & \multicolumn{2}{c}{\large{\textbf{Prediction: \textcolor{ao(english)}{Pristine}}}} \\

\makecell{\fcolorbox{ao(english)}{lightgreen}{\begin{varwidth}{\textwidth}   \begin{center} \fcolorbox{myOrange}{white}{\includegraphics[width=3cm,keepaspectratio]{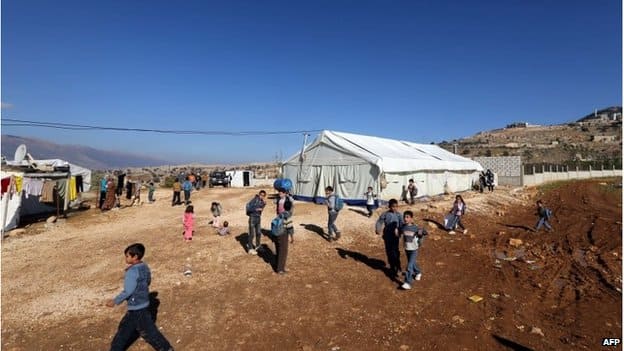}}\end{center}
\fcolorbox{myblue}{white}{\begin{varwidth}{\textwidth}\normalsize{The soaring number of\\Syrian refugees has\\sparked increasing\\resentment in Lebanon}\end{varwidth}}\end{varwidth}}} & 

\makecell{\fcolorbox{myblue}{white}{\begin{varwidth}{\textwidth} \normalsize{\hlc[light_yellow]{`Syria'}, \hlc[light_yellow]{`Lebanon'},\\`United Kingdom', `Tent',\\ \hlc[light_yellow]{`Syrians'}, `Language', \\`Refugee', `Recreation',\\`Tourism', `Camping',\\`Language barrier', \\\hlc[light_yellow]{`rural area'}} \end{varwidth} }    
\fcolorbox{myblue}{white}{\begin{varwidth}{\textwidth} \normalsize{\hlc[light_yellow]{1- Syrian refugees at a camp}\\\hlc[light_yellow]{in eastern Lebanon, December 2014.}\\2- Syrians entering Lebanon face\\new restrictions\\3- Among those displaced, 1.6\\million children have fled Syria.\\4- Syrian refugees in the UK: `We\\will be good people. We will build\\this country’} \end{varwidth} }}
& 
\makecell{ \fcolorbox{myOrange}{light_yellow}{\includegraphics[width=4cm,keepaspectratio]{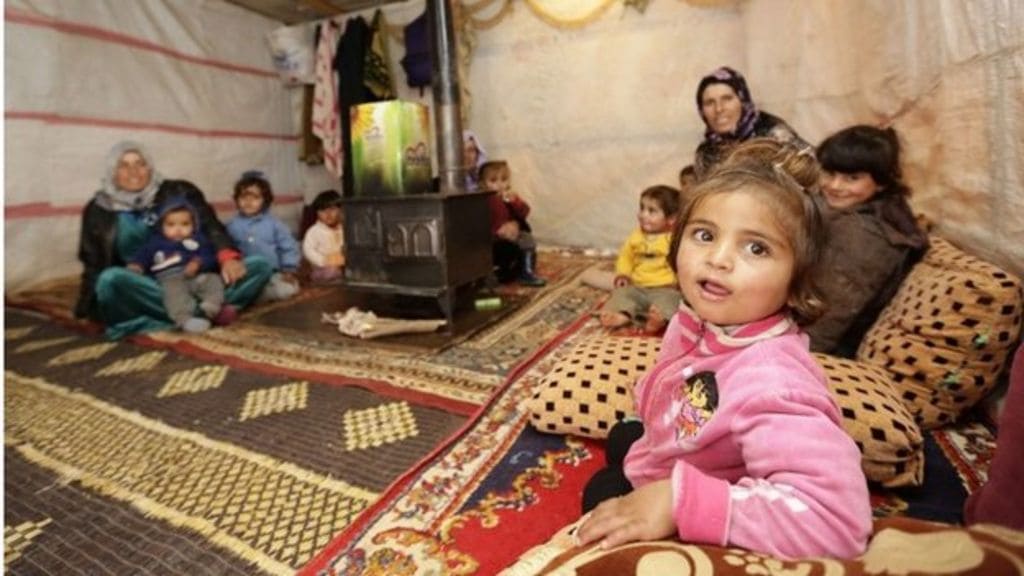}} \fcolorbox{myOrange}{white}{\includegraphics[width=4cm,keepaspectratio]{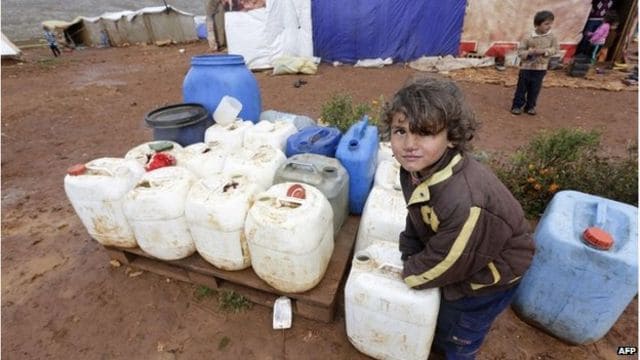}}
\fcolorbox{myOrange}{white}{\includegraphics[width=3.5cm,keepaspectratio]{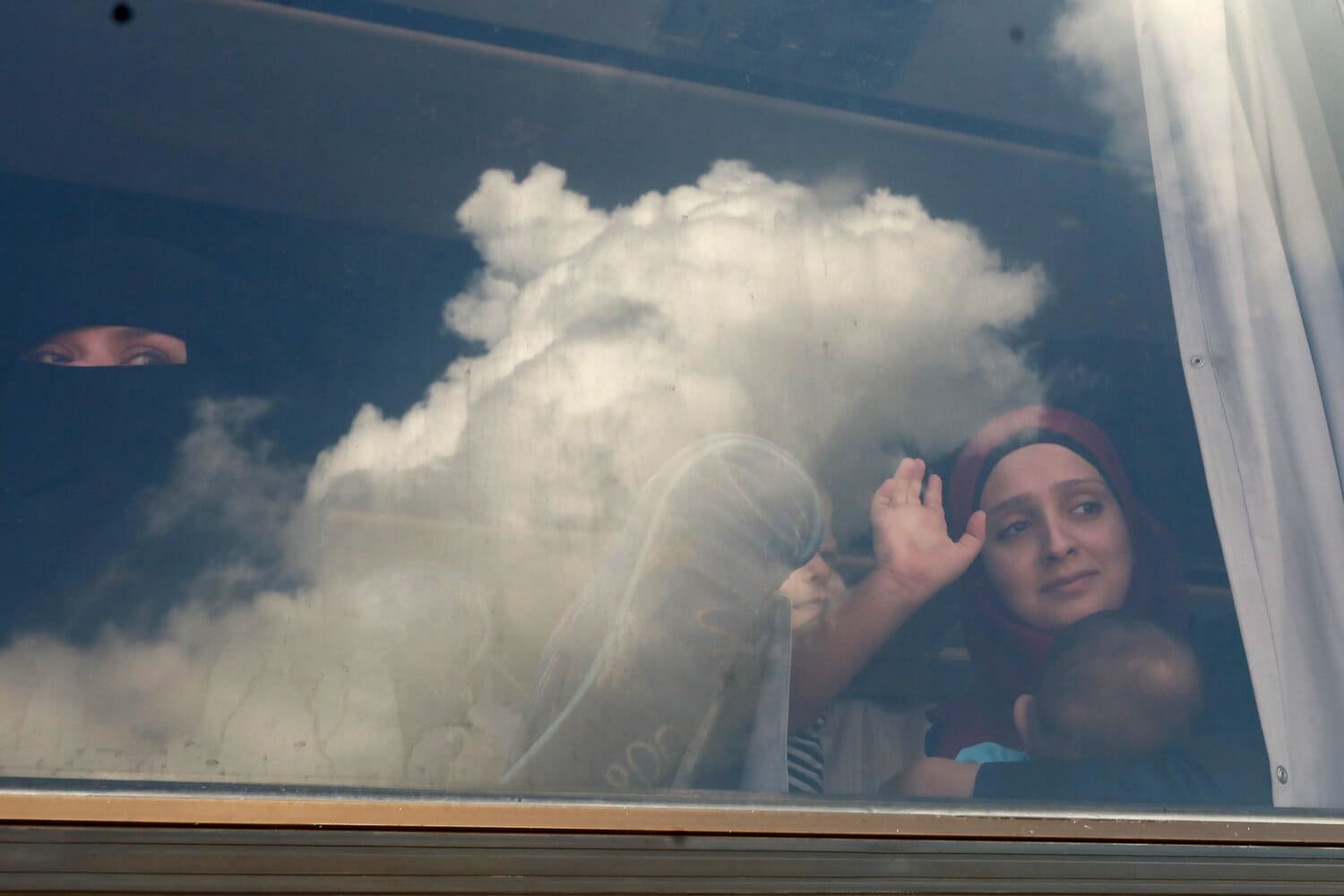}}} 
\\&\multicolumn{2}{c}{\large{\textbf{Prediction: \textcolor{ao(english)}{Pristine}}}}\\

\makecell{\fcolorbox{darkred}{lightred}{\begin{varwidth}{\textwidth}   \begin{center} \fcolorbox{myOrange}{white}{\includegraphics[width=3cm,keepaspectratio]{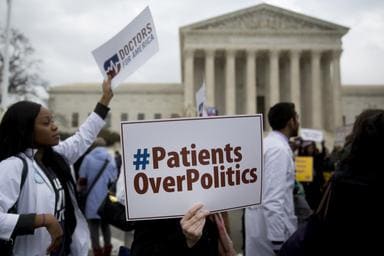}}\end{center}
\fcolorbox{myblue}{white}{\begin{varwidth}{\textwidth}\normalsize{Healthcare activists say the ruling\\against Novartis ensures poor\\people will be able to access\\cheap versions of cancer medicines}\end{varwidth}}\end{varwidth}}} & 

\makecell{\fcolorbox{myblue}{white}{\begin{varwidth}{\textwidth} \normalsize{\hlc[light_yellow]{`United States Capitol'},\\\hlc[light_yellow]{`Affordable Care Act'}\\
\hlc[light_yellow]{`Supreme Court of the United States'},\\\hlc[light_yellow]{`Presidency of Donald Trump'},\\`President of the United States',\\
`United States', `us capitol grounds'} \end{varwidth}}   
\fcolorbox{myblue}{white}{\begin{varwidth}{\textwidth} \normalsize{\hlc[light_yellow]{1- Demonstrators from Doctors for}\\\hlc[light_yellow]{America in support of Obamacare}\\\hlc[light_yellow]{march in front of the Supreme}\\\hlc[light_yellow]{Court on March 4, 2015.}\\2-The Affordable Care Act Is Back\\In Court, 5 Facts You Need To Know.\\3- As Court Hears Arguments in\\Lawsuit To Eliminate Obamacare,\\Conn. Senators Plead Their Case.} \end{varwidth} }}
& 
\makecell{ \fcolorbox{myOrange}{light_yellow}{\includegraphics[width=4cm,keepaspectratio]{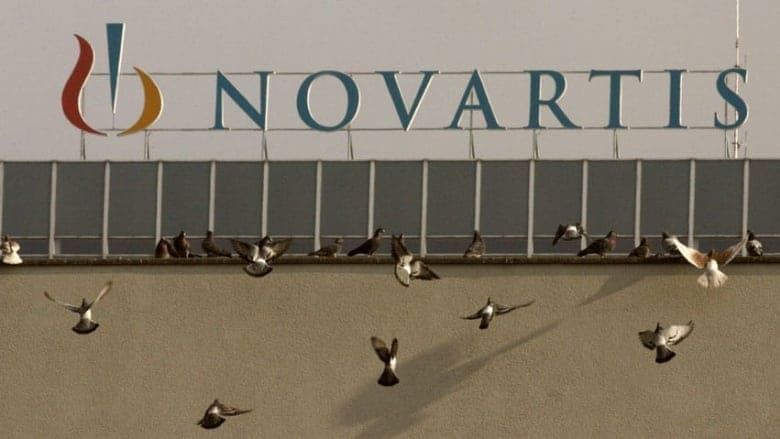}} \fcolorbox{myOrange}{white}{\includegraphics[width=3.5cm,keepaspectratio]{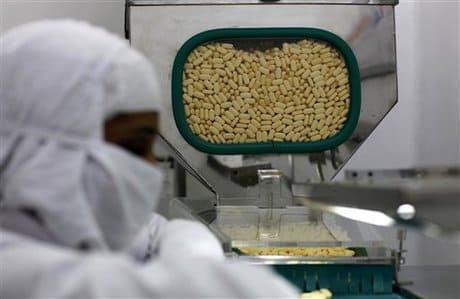}}
\fcolorbox{myOrange}{white}{\includegraphics[width=4cm,keepaspectratio]{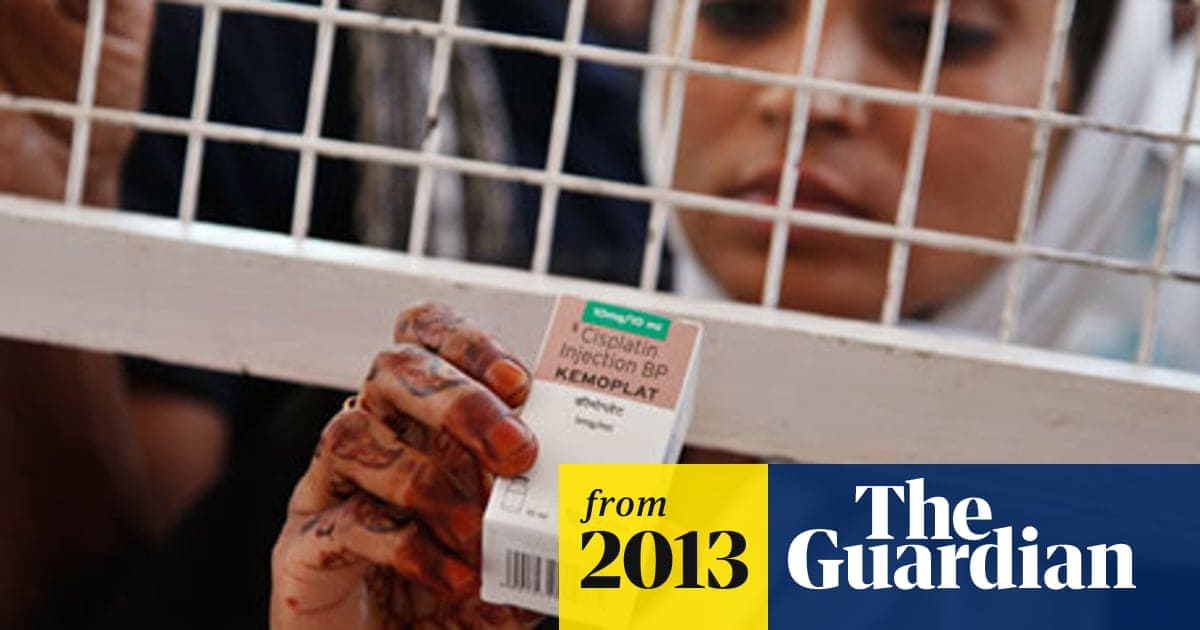}}} \\
&\multicolumn{2}{c}{\large{\textbf{Prediction: \textcolor{darkred}{Falsified}}}}\\ 

\makecell{\fcolorbox{darkred}{lightred}{\begin{varwidth}{\textwidth}   \begin{center} \fcolorbox{myOrange}{white}{\includegraphics[width=3cm,keepaspectratio]{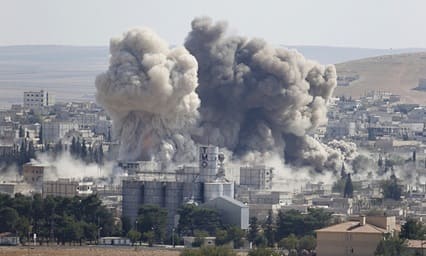}}\end{center}
\fcolorbox{myblue}{white}{\begin{varwidth}{\textwidth}\normalsize{Smoke rises following an\\Israeli air strike in Gaza City}\end{varwidth}}\end{varwidth}}} & 

\makecell{\fcolorbox{myblue}{white}{\begin{varwidth}{\textwidth} \normalsize{`Kobane', \hlc[light_yellow]{`Kurdistan Region'},\\`United States', `Peshmerga',\\`Turkey', \hlc[light_yellow]{`Kurds'},\hlc[light_yellow]{`Syria'},\\\hlc[light_yellow]{`Iraq'}, `kobani war'} \end{varwidth}}    
\fcolorbox{myblue}{white}{\begin{varwidth}{\textwidth} \normalsize{\hlc[light_yellow]{1- Smoke rises after a U.S.-led airstrike}\\\hlc[light_yellow]{in the Syrian town of Kobani}\\2- The border town of Kobani is under\\threat after the Islamists drove 180,000\\Kurds into Turkey.\\3-Former Kurdish Sniper Claims To Have\\Killed Around 250 ISIS Fighters.} \end{varwidth} }}
& 
\makecell{ \fcolorbox{myOrange}{light_yellow}{\includegraphics[width=4cm,keepaspectratio]{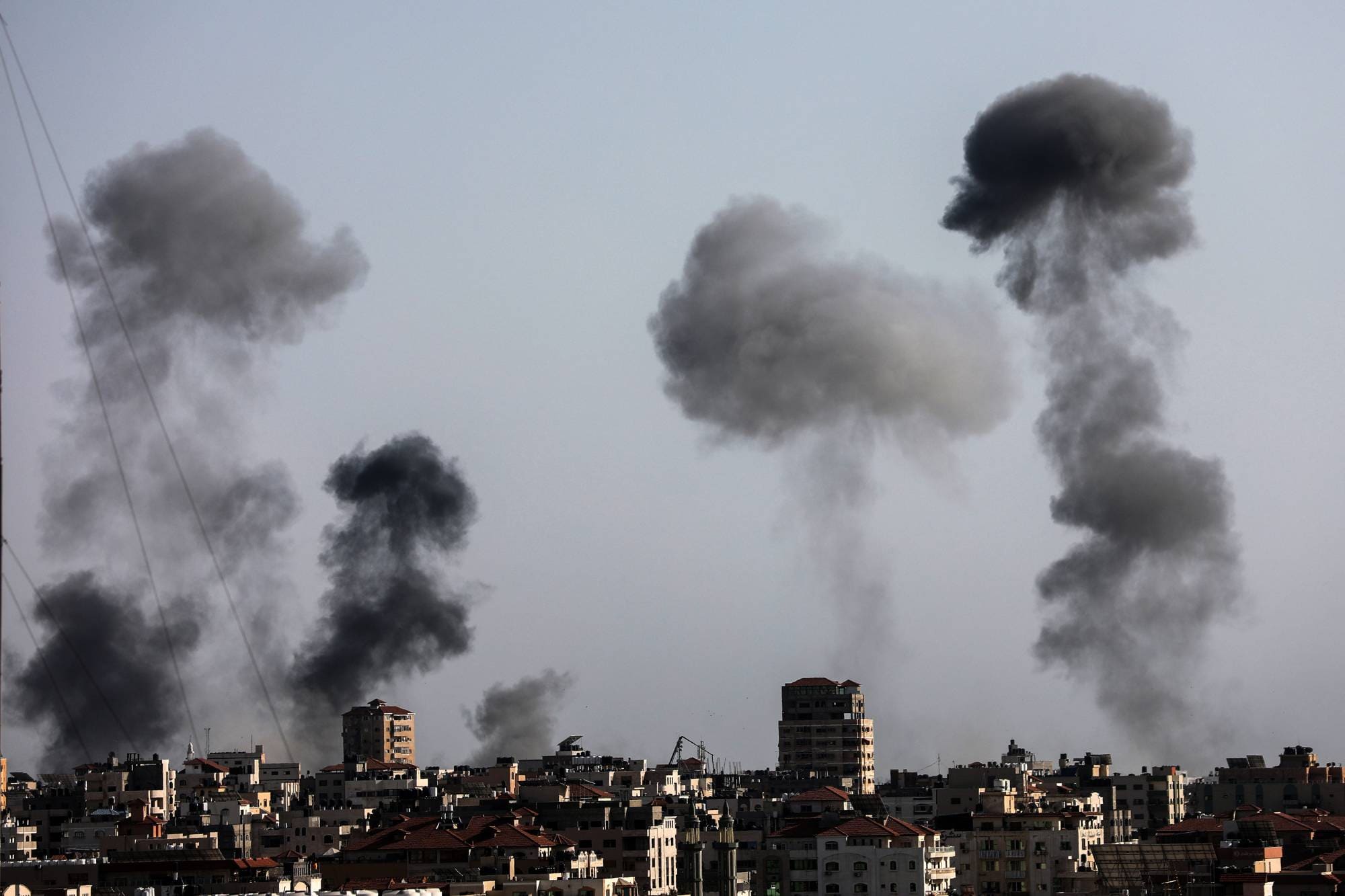}} \fcolorbox{myOrange}{white}{\includegraphics[width=4cm,keepaspectratio]{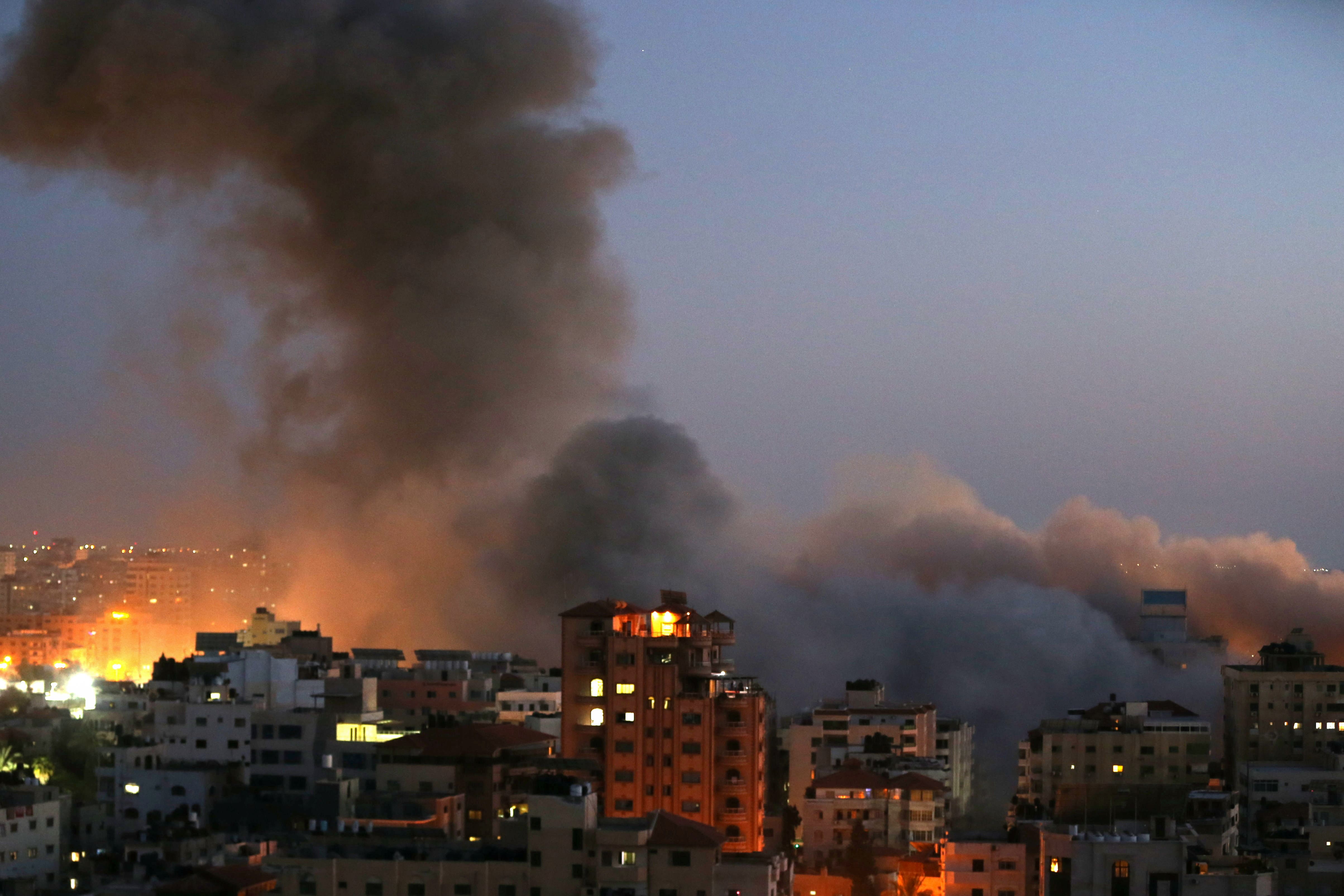}}
\fcolorbox{myOrange}{white}{\includegraphics[width=4cm,keepaspectratio]{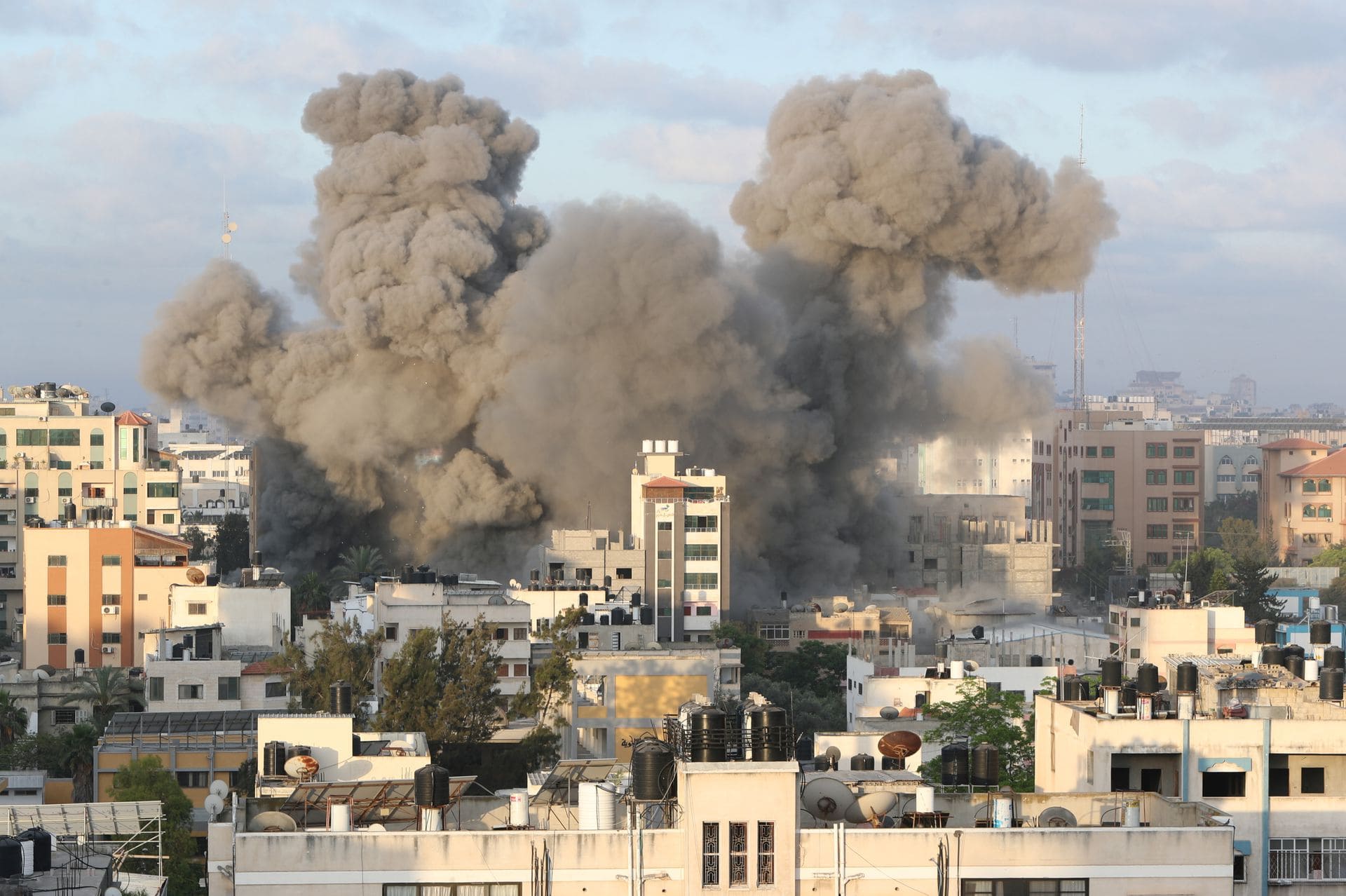}}} \\
&\multicolumn{2}{c}{\large{\textbf{Prediction: \textcolor{darkred}{Falsified}}}}\\ \midrule

\makecell{\fcolorbox{ao(english)}{lightgreen}{
\begin{varwidth}{\textwidth} \begin{center}\fcolorbox{myOrange}{white}{\includegraphics[width=3.5cm,keepaspectratio]{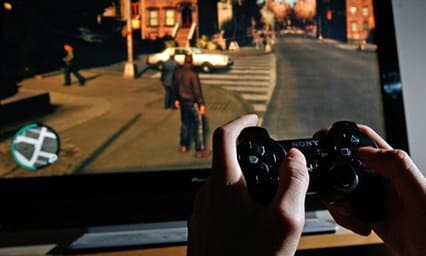}}\end{center} 
\fcolorbox{myblue}{white}{\begin{varwidth}{\textwidth}\normalsize{How can our young readers\\persuade their parents\\to get them a Playstation 3}\end{varwidth} }\end{varwidth}}} & 

\makecell{\fcolorbox{myblue}{white}{\begin{varwidth}{\textwidth} \normalsize{\hlc[light_yellow]{`Grand Theft Auto V'},`Gamer',\\ `Grand Theft Auto IV', `Wii', \\`Grand Theft Auto VI',\\ \hlc[light_yellow]{`PlayStation 3'},\\,`Rockstar Leeds',\\\hlc[light_yellow]{`terry seeborne marshall'},\\ \hlc[light_yellow]{`Gordon Hall'},\\`Rockstar Games'} \end{varwidth} }   
\fcolorbox{myblue}{white}{\begin{varwidth}{\textwidth} \normalsize{\hlc[light_yellow]{1- A court order banning Sony from}\\\hlc[light_yellow]{importing PS3s into the Netherlands}\\\hlc[light_yellow]{has been lifted.}\\2- Rockstar Games, creators of the\\Grand Theft Auto franchise, said\\it was "very saddened" to hear of\\Mr Hall's death\\3- Oakland Athletics to Begin\\Accepting Bitcoin for Private Suites} \end{varwidth} }} 
& 
\makecell{ \fcolorbox{myOrange}{light_yellow}{\includegraphics[width=4cm,keepaspectratio]{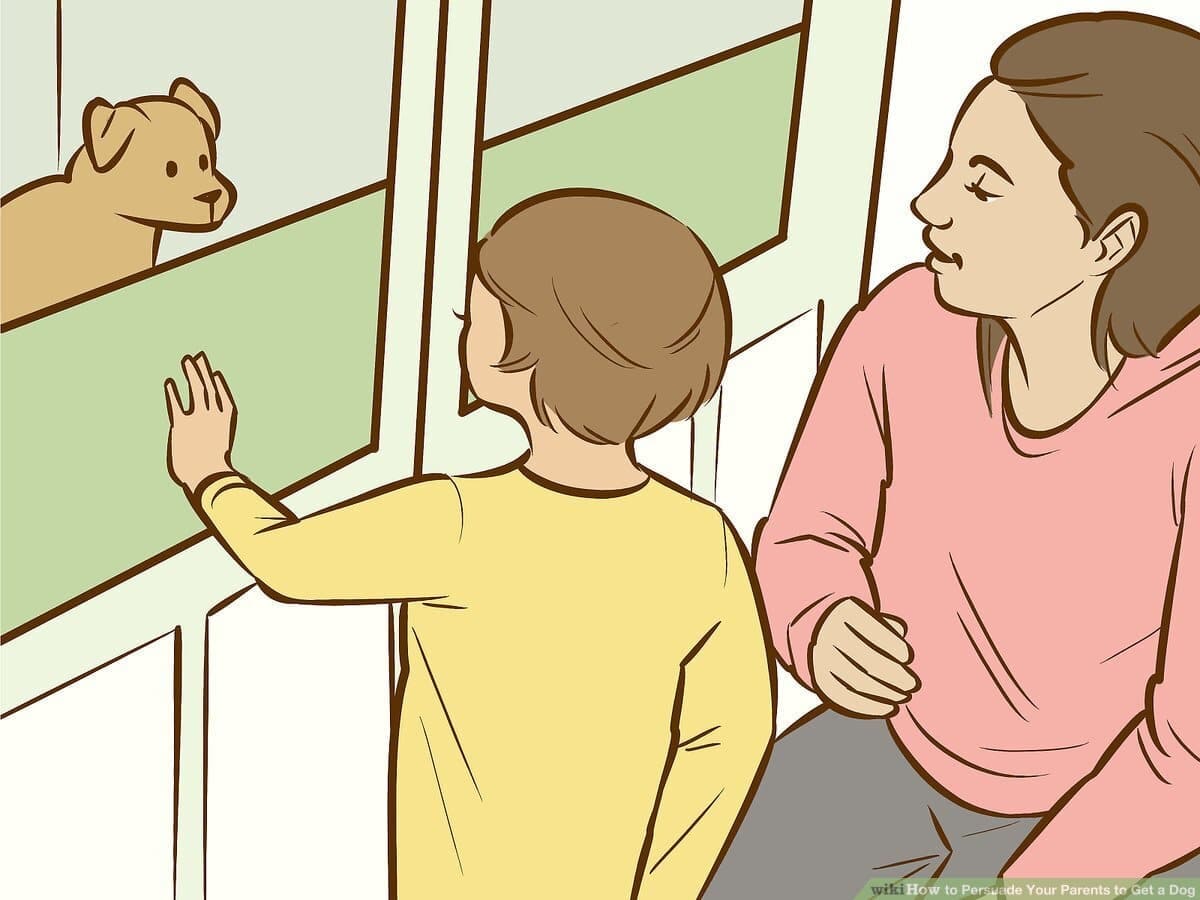}}
\fcolorbox{myOrange}{white}{\includegraphics[width=4cm,keepaspectratio]{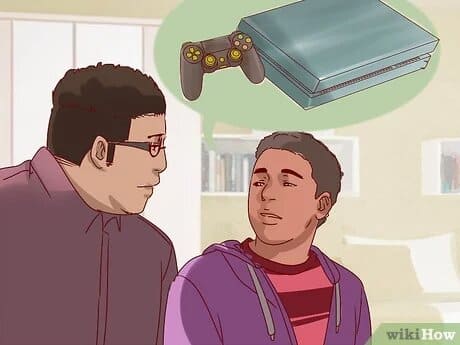}}
\fcolorbox{myOrange}{white}{\includegraphics[width=4cm,keepaspectratio]{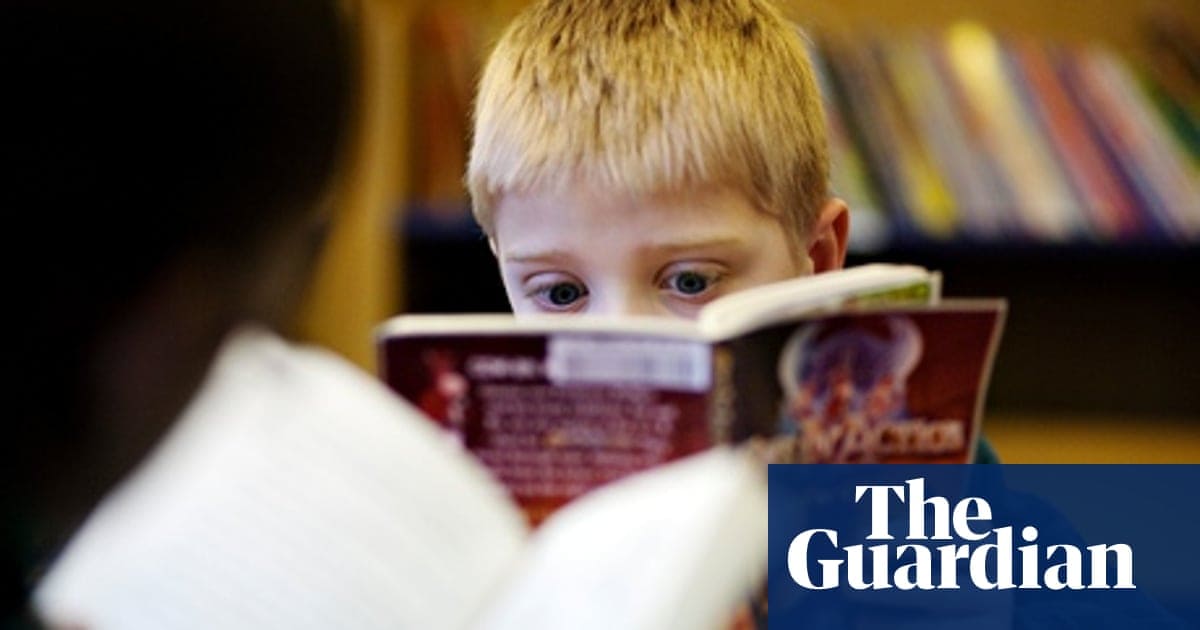}}}\\

& \multicolumn{2}{c}{\large{\textbf{Prediction: \textcolor{darkred}{Falsified}}}} \\

\bottomrule 
\end{tabular}}
\captionof{figure}{Qualitative examples of news pairs along with the collected evidence. Examples with \hlc[lightgreen]{green background} are pristine, \hlc[lightred]{red background} are falsified. \hlc[light_yellow]{Highlighted items} are the ones with the highest attention. Only a subset of the evidence is shown for display purposes.}
\label{tbl:qual}
%\vspace{-1mm}
\end{table*}
\subsection{Qualitative Analysis}
We show some successful predictions of our model in Figure~\ref{tbl:qual}. When inspecting the attention in the case of \textit{pristine} examples, we found that the highest attention is on items that are most relevant to the query (e.g., a similar \textbf{\textcolor{myOrange}{image}} in the first example, named \textbf{\textcolor{myblue}{entities}} that are present in or similar to the query \textbf{\textcolor{myblue}{caption}} such as cities' names, and semantically similar \textbf{\textcolor{myblue}{captions}}). The model also predicted the second example correctly, despite not having an \textbf{\textcolor{myOrange}{image}} of the same scene. For \textit{falsified} examples, we observe that the third one is predicted correctly despite having a similar falsified topic (\textit{`Affordable health care'} and \textit{`Lawsuits'}). Moreover, the fourth one shows the highest attention on contradicting locations in \textbf{\textcolor{myblue}{entities}}, and on the most syntactically similar \textbf{\textcolor{myblue}{caption}}. This was predicted correctly, despite having similar-style evidence to the query. Similarly, the falsified example in~\autoref{fig:teaser} was similar in the persons' names and images (\textit{`David Cameron'}), but different in context and scene details. Finally, the last example shows a \textit{pristine} example that was misclassified as \textit{falsified}. When inspecting the \textbf{\textcolor{myblue}{textual evidence}}, we observed that although it is revolving around the same topic, there is little connection to the context of the query \textbf{\textcolor{myblue}{caption}}, in addition to having a diverse set of \textbf{\textcolor{myOrange}{visual evidence}} that is not similar to the query \textbf{\textcolor{myOrange}{image}}. Other examples are in Supp.~\ref{sec:qual_analysis2}.

\subsection{Discussion and Limitations}
We propose a multi-modal fact-checking framework  
that significantly outperforms baselines and is comparable to human performance. However, the task has yet many challenges, and fully relying on automated tools might have dangerous consequences. 
Therefore, humans should still be in the loop. Thus, we offer an inspectable and assistive tool that helps to reduce the load of the otherwise fully manual process~\cite{press_vice}. We further discuss potential risks in Supp.~\ref{sec:risks}.

Moreover, our approach relies on the retrieval results of the search engine. However, as we show in our analysis (Tables~\ref{table:ablation} and~\ref{table:baseline}), naively considering the evidence is not adequate, and a careful design of the model is needed to meet the challenges of the task, including the noisy open-domain setup with no relevancy supervision, and the high resemblance of evidence across pristine and falsified examples.

Finally, in some situations, some evidence items might contradict others, e.g., due to the websites' opposing political orientations, or misinformation on the Web. We did not observe such scenarios with the used dataset; identifying and studying them might require poisoning the search results, or carefully curating claims that lead to contradicting results, which is beyond the scope of this work.
\section{Conclusion}
We mimic the complex fact-checking process in an automated framework, \model{}, that aggregates consistency signals and consensus from multi-modal evidence found on the Web, and the given \textbf{\textcolor{myOrange}{image}}-\textbf{\textcolor{myblue}{caption}} pairing. Our work significantly outperforms previous baselines and offers a new task and benchmark of multi-modal fact-checking, and an automated, inspectable tool to assist manual fact-checking.

\section*{Acknowledgment}
This work was supported by the Google Cloud Research Credits program. We also thank Rebecca Weil for helpful advice and feedback.

%-------------------------------------------------------------------------

%%%%%%%%% REFERENCES
{\small
\bibliographystyle{ieee_fullname}
\bibliography{egbib}
}
\clearpage
\setcounter{section}{0} 

\section*{Supplementary Material} In the supplementary material, we first discuss more implementation details in Section~\ref{sec:implem} and present additional experiments in Section~\ref{sec:add_exp}. Then we present some examples that were selected as `hard to verify' in the user study in Section~\ref{sec:user_study_ex}. We present other qualitative examples in Section~\ref{sec:qual_analysis2}. Finally, we discuss societal aspects and potential risks in Section~\ref{sec:risks}.

\section{Implementation Details} \label{sec:implem}
We elaborate on some implementation details of our framework. 
\begin{itemize}

\item \textbf{Sentence representation.}
We preprocessed the crawled captions to remove some artefacts (e.g., HTML tags). When using BERT+LSTM, we used the pre-trained `bert-base-uncased' model, whose dimension is 768. We set a maximum length of 150 tokens for the captions. Items (i.e., query captions, evidence captions, and entities) are padded to the maximum sequence length in this item's batch. When using the sentence transformer model, we used the `paraphrase-mpnet-base-v2' model\footnote{https://huggingface.co/sentence-transformers/paraphrase-mpnet-base-v2}. For both, we used the Hugging Face library\footnote{https://huggingface.co/}. We used the PyTorch framework\footnote{https://pytorch.org/} for all our experiments.

\item \textbf{Memory.}
The items in each memory (images, entities, and captions) are padded to the maximum number of evidence items in this memory's batch.

\item \textbf{CLIP.}
We used the pre-trained ViT-B/32 CLIP model\footnote{https://github.com/openai/CLIP}, where the text length is truncated at 77 tokens.

\item \textbf{Training details.}
When fine-tuning CLIP, we follow the implementation details in~\cite{luo2021newsclippings}, we used a learning rate of 5e-5 for
the linear classifier and 5e-7 for other layers of the CLIP model itself, in addition to using the Adam optimizer~\cite{kingma2014adam}. We used a batch size of 64 and trained the model for 100 epochs. For training \model{}, we used a batch size of 32, the Adam optimizer, and a cyclical learning rate~\cite{smith2017cyclical} with a maximum value of 6e-5. We trained the model for 30 epochs. We used a dropout~\cite{srivastava2014dropout} value of 0.05 to the input representations, 0.25 to domain embeddings, and 0.25 to the memory representations. Experiments were done on one NVIDIA A100 GPU. With precomputing the representations, the training takes roughly 5 hours. When training using BERT without precomputing, training takes roughly 30 hours.

\end{itemize}
\section{Additional Experiments} \label{sec:add_exp}

\paragraph{Evidence-only classification.} We examine whether claims (and consequently, the evidence) are having different characteristics (and thus, unwanted biases or naive give-aways) between pristine and falsified classes. The NewsCLIPpings dataset avoided linguistic biases in creating falsified examples by using real news \textbf{\textcolor{myblue}{captions}} mismatched with real news \textbf{\textcolor{myOrange}{images}}, instead of introducing manipulations in the captions. Also, to avoid text bias, each \textbf{\textcolor{myblue}{caption}} (and consequently, its \textbf{\textcolor{myOrange}{visual evidence}} in our dataset) appears twice (within the same split), once as pristine and once as falsified. Therefore, we hypothesize that the evidence websites for both classes are similar. To confirm, we ran an \textit{evidence-only} model, which achieved 53.4\% (\textit{basically chance level}), showing that \textit{reasoning against the query} is the distinguishing factor.

\begin{table}[!b]
\begin{center}
\vspace{-1mm}
\resizebox{0.65\linewidth}{!}{
\begin{tabular}{cccc}
\toprule
Conc. & Avg-pool & Max-pool & Multiply \\ \midrule
\textbf{83.9} &  82.46 & 82.48 & 77.1 \\ \bottomrule
\end{tabular}}
\end{center}
\vspace{-6mm}
\caption{Accuracy (\%) vs. aggregation strategies.}
\vspace{-3mm}
\label{rebuttal_tab:ablation1}
\end{table}

\paragraph{Additional ablation studies.} We include further experiments related to the fusion of the different components in our model (visual reasoning, textual reasoning, and CLIP). We tried a late fusion by having a separate classifier on top of each branch and aggregating the decision, however, this performed worse than the current intermediate fusion we employ. We also tried other strategies (\autoref{rebuttal_tab:ablation1}) to combine visual and textual memories before concatenating with CLIP, where we found that concatenation had the highest performance. 

Finally, we found that changing the dimension of the penultimate layer had a relatively small effect; e.g., increasing the dimension to 2048 increased the accuracy by 0.3 percentage points.

\section{User Study: `Hard to Verify' Examples} \label{sec:user_study_ex}
In Figure~\ref{tbl:qual_study_appendix}, we show some examples that were selected as `hard to verify' in the user study. This is possibly due to: 1) the \textbf{\textcolor{myblue}{captions}} could contain specific context information (e.g., locations such as \textit{`Denver'} or \textit{`Massachusetts'}) that is hard to verify with the \textbf{\textcolor{myOrange}{image}} alone, 2) the lack of \textbf{\textcolor{myblue}{textual}} evidence returned by the search \vcenteredinclude{figs/icon3.pdf}; the \textbf{\textcolor{myOrange}{images}} were not found by the inverse image search, so there are no \textbf{\textcolor{myblue}{captions/titles}} found. Moreover, the \textbf{\textcolor{myblue}{entities}} are generic descriptions of the \textbf{\textcolor{myOrange}{image}}, or not at all related (the first example). The performance of the model on these examples is possibly dependent on how similar the \textbf{\textcolor{myOrange}{visual}} evidence is to the query \textbf{\textcolor{myOrange}{image}} \vcenteredinclude{figs/icon4.pdf}. 

Another possible reason is having falsified pairs that are highly similar in context to the original ones (and, therefore, to the evidence as well). For instance, the last example shows a `hard to verify' falsified example (that was also misclassified by our model); the \textbf{\textcolor{myOrange}{image}} shows the same people mentioned in the \textbf{\textcolor{myblue}{caption}}, and thus, they also appeared in the \textbf{\textcolor{myOrange}{visual}} evidence. Additionally, the \textbf{\textcolor{myblue}{caption}} mentions the band name \textit{`One Direction'} that is also mentioned in the \textbf{\textcolor{myblue}{textual}} evidence, without strong contradictions. Meanwhile, the actual \textbf{\textcolor{myOrange}{image}} of this \textbf{\textcolor{myblue}{caption}} showed the band performing on a stage, however, this was not clearly emphasized by the \textbf{\textcolor{myblue}{caption}}; that is possibly why the \textbf{\textcolor{myOrange}{visual}} evidence is generic.

\section{Qualitative Examples} \label{sec:qual_analysis2}
In Figure~\ref{tbl:qual_appendix}, we show more qualitative examples. \model{} predicted many examples correctly despite not having a one-to-one matching with the evidence in the case of pristine examples and having close similarity to the evidence in the case of falsified examples. 

For instance, in the first three examples (pristine), we observed that the model highly attended to supporting evidence such as persons' and countries' names, topics, and events. Additionally, in the third example, we observed that the model prioritized the \textbf{\textcolor{myOrange}{image}} that is from the same scene and the evidence \textbf{\textcolor{myblue}{caption}} that contains a subset from the query \textbf{\textcolor{myblue}{caption}} (\textit{`soon to be a Trump International Hotel'}). 

The fourth and fifth examples (falsified) suggest that the model does not simply rely on having any similarity or overlap between the query and evidence in order to identify pristine examples. Despite having the same persons in the evidence, they were correctly predicted as falsified, possibly as they have contradicting location information and different scene details (e.g., lighting, stage setup, or colours), indicating a different context or event. The last falsified example also indicates that both \textbf{\textcolor{myblue}{textual}} and \textbf{\textcolor{myOrange}{visual}} evidence is helpful, as the evidence \textbf{\textcolor{myOrange}{images}} are clearly different from the falsified one (showing a different building and place). 

%As for the last example, it highlights one of the `hard to detect' examples in the dataset, even with the presence of evidence, as the falsified image is also showing a similar object \textit{`iPhone'} without a strongly different context. 
%\clearpage

\section{Limitations and Societal Aspects}  \label{sec:risks}
%Automating fact-checking can be beneficial to fight the spread of misinformation. 
Nowadays, with the spread and reliance on social media to digest and get updated with news, misinformation (e.g., on Twitter) can reach hundreds of millions of users~\cite{vo2020facts}. This crucially motivates the need to fact-check and verify the credibility of online content, especially during critical times such as a pandemic or political instabilities. On the other hand, manual fact-checking is usually time-consuming, needing from less than one hour to many days to verify a claim~\cite{thorne2018automated}. Therefore, automating fact-checking can be extremely beneficial to alleviate the burden upon fact-checkers and journalists. 

However, completely or overly relying on automated tools might give an unwanted sense of security and could have many dangerous consequences. These include the dangers of flagging many true examples as falsified due to the real-life class imbalance, and missing out challenging falsified examples that require more fine-grained and complex reasoning. In addition, a currently active and much-needed research direction in the textual domain shows that fact-verification models might be partially relying on dataset biases without in-depth understanding and reasoning~\cite{schuster2019towards}. They might also be brittle to complex claims that require multi-hop reasoning~\cite{hidey2020deseption}. Additionally, as facts are continuously evolving, we face the danger of relying on old retrieved evidence~\cite{schuster2021get} or even possibly outdated world knowledge that is implicitly stored in pre-trained language models during training~\cite{schuster2019towards}.

In addition to their inherent limitations in reasoning and interpretation, several works have shown that textual verifications models are also vulnerable to adversarial attacks~\cite{thorne2019evaluating}, such as inserting trigger words~\cite{atanasova2020generating}, introducing lexical variations~\cite{hidey2020deseption}, or paraphrasing\cite{thorne2019evaluating}. As we have a multi-modal task, our model might also be vulnerable to image-based adversarial attacks~\cite{goodfellow2014explaining}. %Beyond manipulating claims via adversarial attacks, adversaries can also poison the evidence and introduce items that lead to the required entailment. 
Another potential misuse scenario is using the fact-checking model as an adversarial filter in order to curate hard examples that might be misclassified by fact-checking models in general. 

As a conclusion, we believe that automating fact-checking is strongly beneficial and that there have been many encouraging advancements to improve and harden it in the textual domain and the multi-modal domain, as we propose. However, due to their limitations and vulnerabilities to active attacks and manipulation, they should be used to assist humans and speed up the process, while still keeping them in the loop to avoid such dangers and consequences. In this regard, in our framework, we show that the model can filter and select the most important evidence, which would enable quicker inspection of the evidence items.

\begin{table*}[!t]
\centering
\resizebox{\linewidth}{!}{%
\begin{tabular}{c|c c}
\toprule
\textbf{\textcolor{myOrange}{\large{Image}}}-\textbf{\textcolor{myblue}{\large{caption}}} \large{pair} & \large{\textbf{\textcolor{myblue}{Textual evidence}}} \largericon{figs/icon3.pdf} & \large{\textbf{\textcolor{myOrange}{Visual evidence}}} \largericon{figs/icon4.pdf} \\ \midrule
\makecell{\fcolorbox{ao(english)}{lightgreen}{
\begin{varwidth}{\textwidth} \begin{center}\fcolorbox{myOrange}{white}{\includegraphics[width=4.5cm,keepaspectratio]{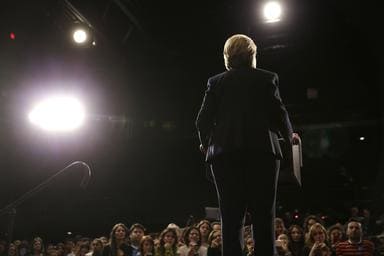}}\end{center} 
\fcolorbox{myblue}{white}{\begin{varwidth}{\textwidth}\normalsize{Clinton speaks at a rally in\\Purchase NY on March 31 2016}\end{varwidth} }\end{varwidth}}} & 

\makecell{\fcolorbox{myblue}{white}{\begin{varwidth}{\textwidth} \normalsize{`Rock concert', `Concert', `stage',\\`Performance art', `Musician',\\`Night', `Rock', `Art',\\`Performance', `Artist'} \end{varwidth}}
\fcolorbox{myblue}{white}{\begin{varwidth}{\textwidth} \normalsize{No pages found.} \end{varwidth}}}
& 
\makecell{ \fcolorbox{myOrange}{white}{\includegraphics[width=4.5cm,keepaspectratio]{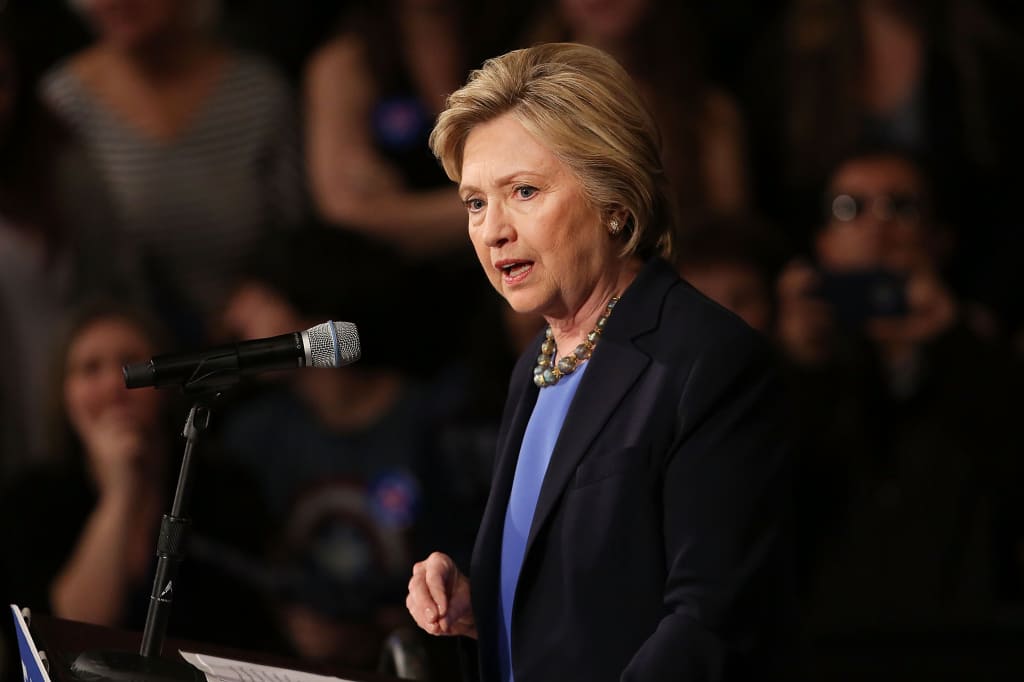}} \fcolorbox{myOrange}{white}{\includegraphics[width=4.5cm,keepaspectratio]{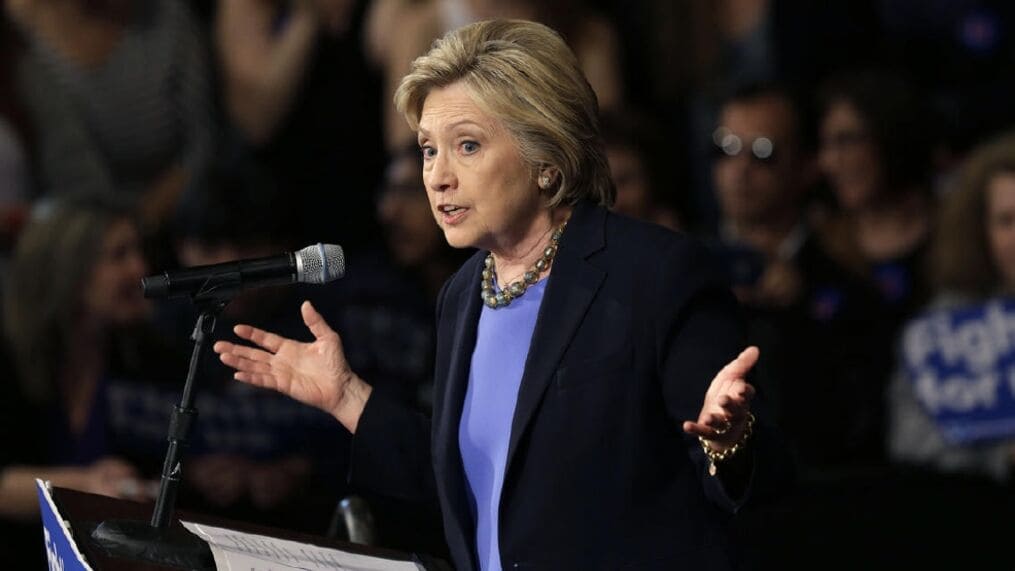}}
\fcolorbox{myOrange}{white}{\includegraphics[width=4.5cm,keepaspectratio]{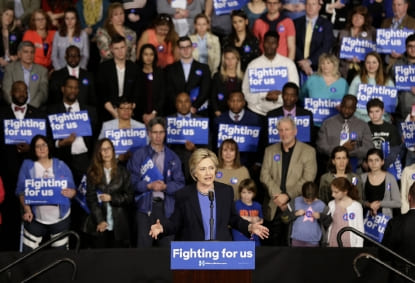}}
\fcolorbox{myOrange}{white}{\includegraphics[width=4.5cm,keepaspectratio]{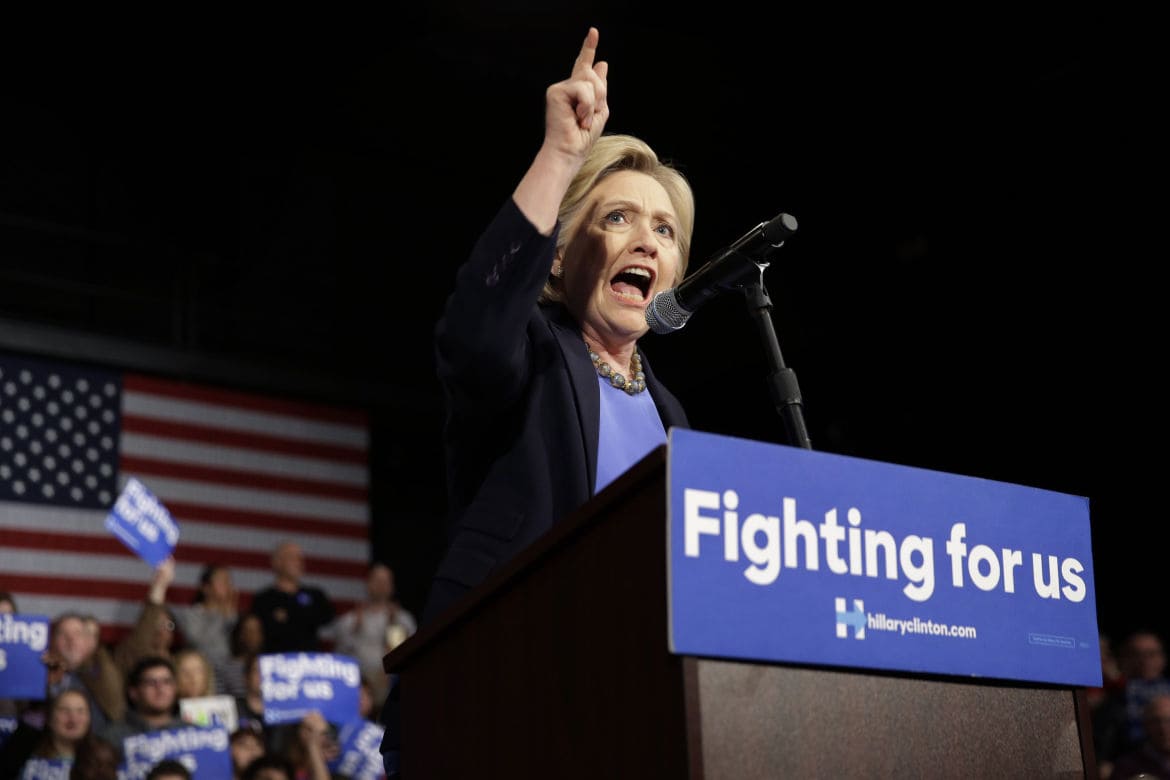}}}  
\\ & \multicolumn{2}{c}{\hspace{-8cm}\large{\textbf{Prediction: \textcolor{ao(english)}{Pristine}}}} \\

\makecell{\fcolorbox{ao(english)}{lightgreen}{
\begin{varwidth}{\textwidth} \begin{center}\fcolorbox{myOrange}{white}{\includegraphics[width=4.5cm,keepaspectratio]{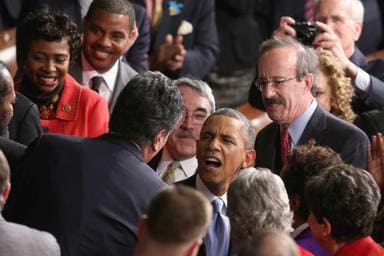}}\end{center} 
\fcolorbox{myblue}{white}{\begin{varwidth}{\textwidth}\normalsize{President Obama makes his\\way out of chamber after\\delivering the State of the\\Union address to a joint\\session of Congress
}\end{varwidth} }\end{varwidth}}} & 

\makecell{\fcolorbox{myblue}{white}{\begin{varwidth}{\textwidth} \normalsize{`Event', `Statistics', `crowd'} \end{varwidth}}
\fcolorbox{myblue}{white}{\begin{varwidth}{\textwidth} \normalsize{No pages found.} \end{varwidth}}}
& 
\makecell{ \fcolorbox{myOrange}{white}{\includegraphics[width=4.5cm,keepaspectratio]{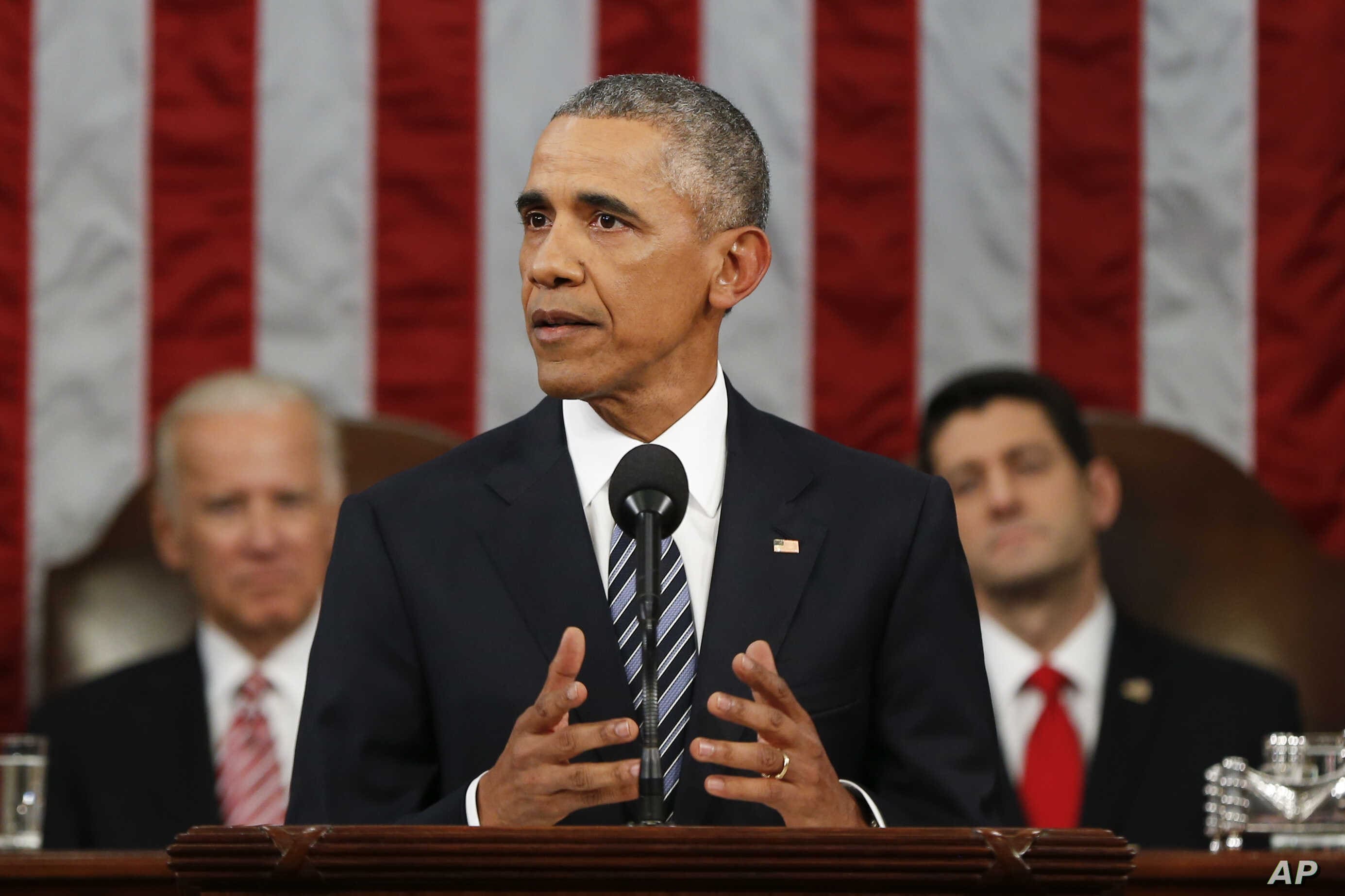}} \fcolorbox{myOrange}{white}{\includegraphics[width=4.5cm,keepaspectratio]{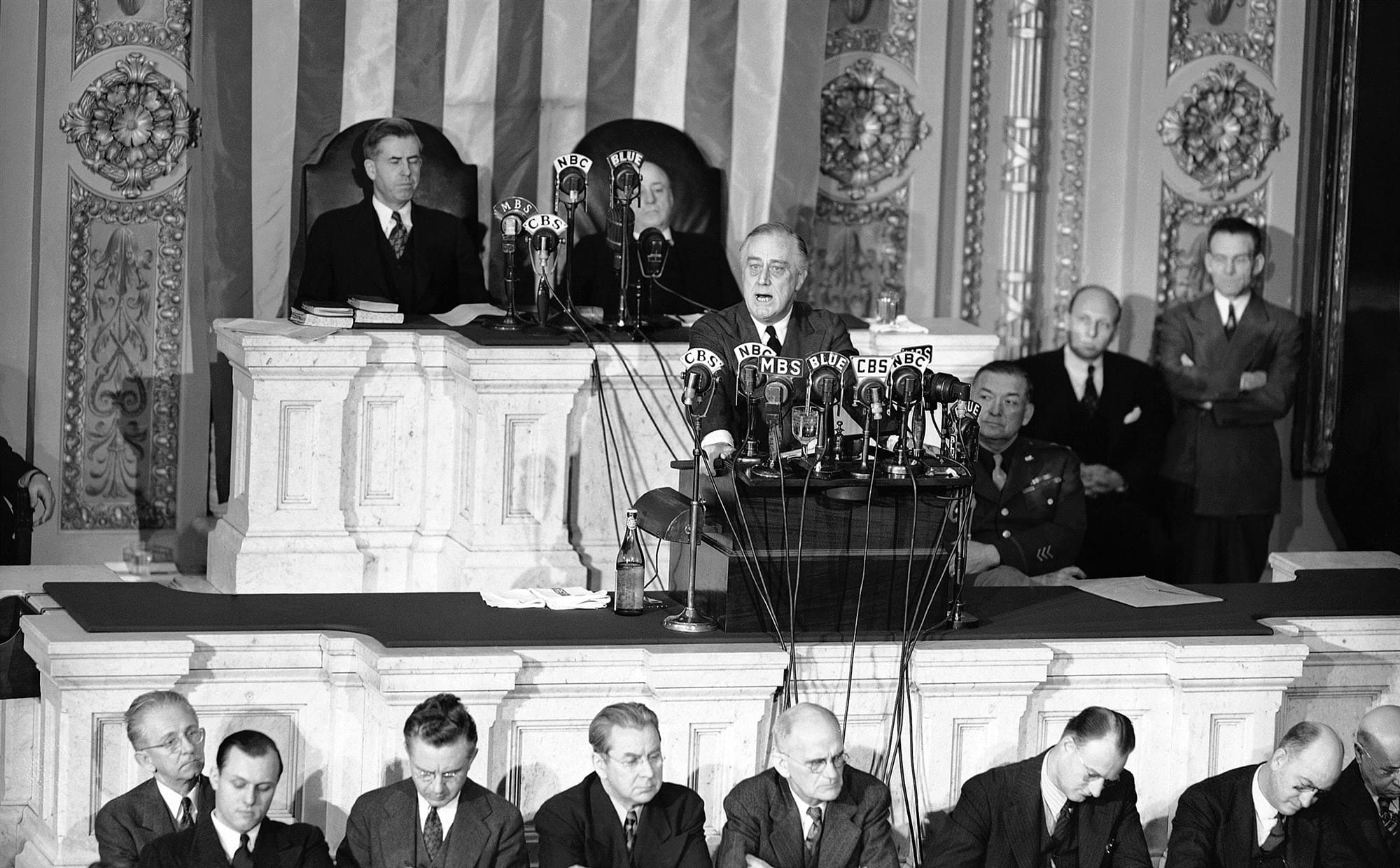}}
\fcolorbox{myOrange}{white}{\includegraphics[width=4.5cm,keepaspectratio]{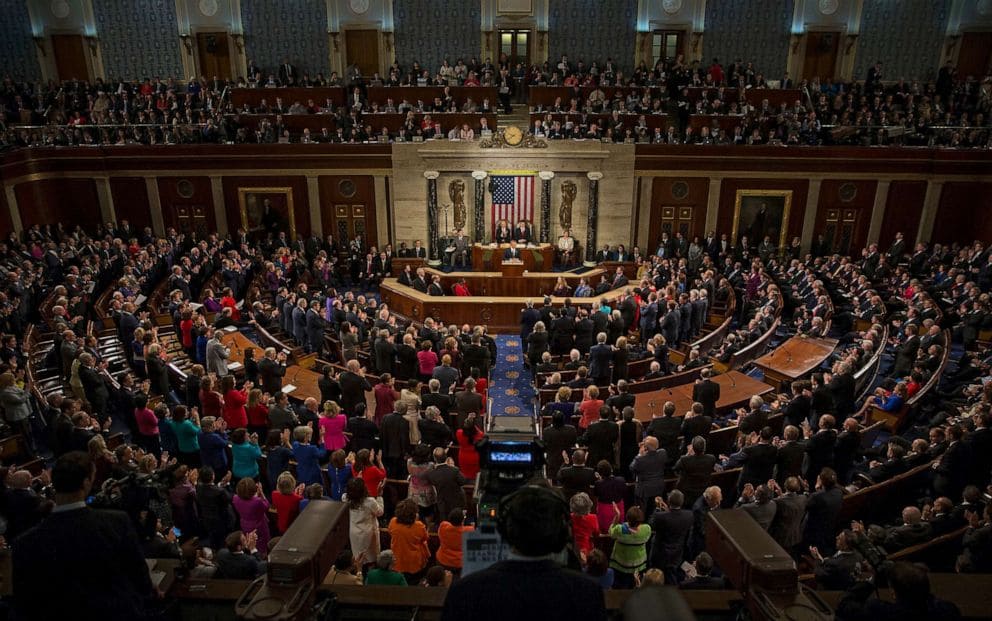}}
\fcolorbox{myOrange}{white}{\includegraphics[width=4.5cm,keepaspectratio]{figs/appendix/3122/4.jpg}}}  
\\ & \multicolumn{2}{c}{\hspace{-8cm}\large{\textbf{Prediction: \textcolor{darkred}{Falsified}}}} \\

\makecell{\fcolorbox{ao(english)}{lightgreen}{
\begin{varwidth}{\textwidth} \begin{center}\fcolorbox{myOrange}{white}{\includegraphics[width=4.5cm,keepaspectratio]{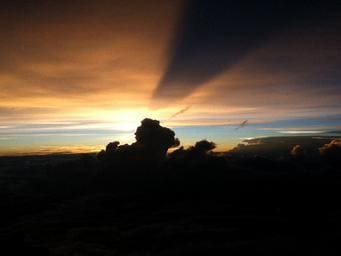}}\end{center} 
\fcolorbox{myblue}{white}{\begin{varwidth}{\textwidth}\normalsize{Dark shadows Lowlevel clouds cast\\a shadow across a higher cloud\\level at sunset last week over Denver}\end{varwidth} }\end{varwidth}}} & 

\makecell{\fcolorbox{myblue}{white}{\begin{varwidth}{\textwidth} \normalsize{`Red sky at morning',\\`Cumulus',`Sky',`sky'\\`Sunlight', `Wallpaper',\\`Atmosphere', `Ecoregion',\\`Computer', `Phenomenon'} \end{varwidth}}
\fcolorbox{myblue}{white}{\begin{varwidth}{\textwidth} \normalsize{No pages found.} \end{varwidth}}}
& 
\makecell{ \fcolorbox{myOrange}{white}{\includegraphics[width=4.5cm,keepaspectratio]{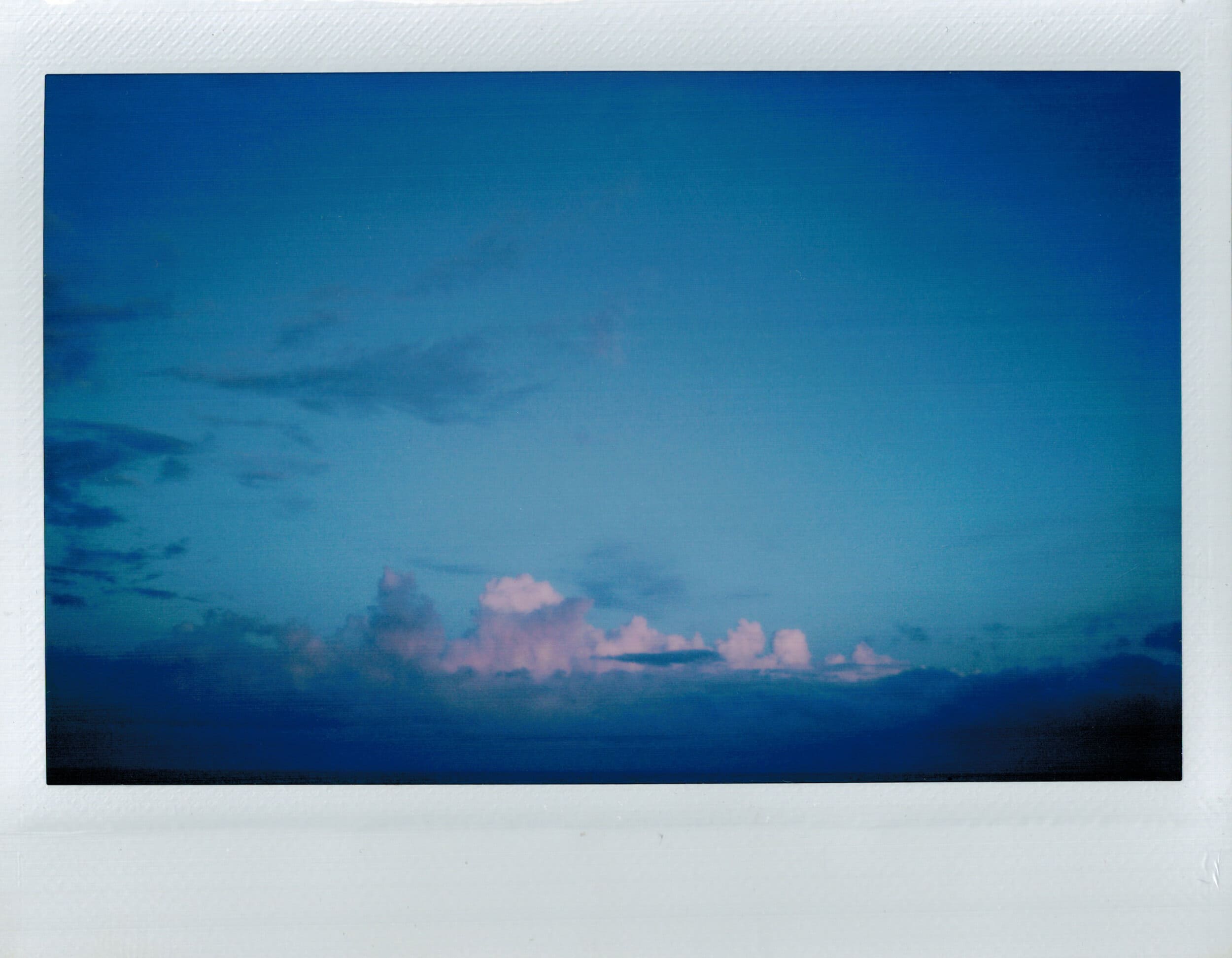}} \fcolorbox{myOrange}{white}{\includegraphics[width=4.5cm,keepaspectratio]{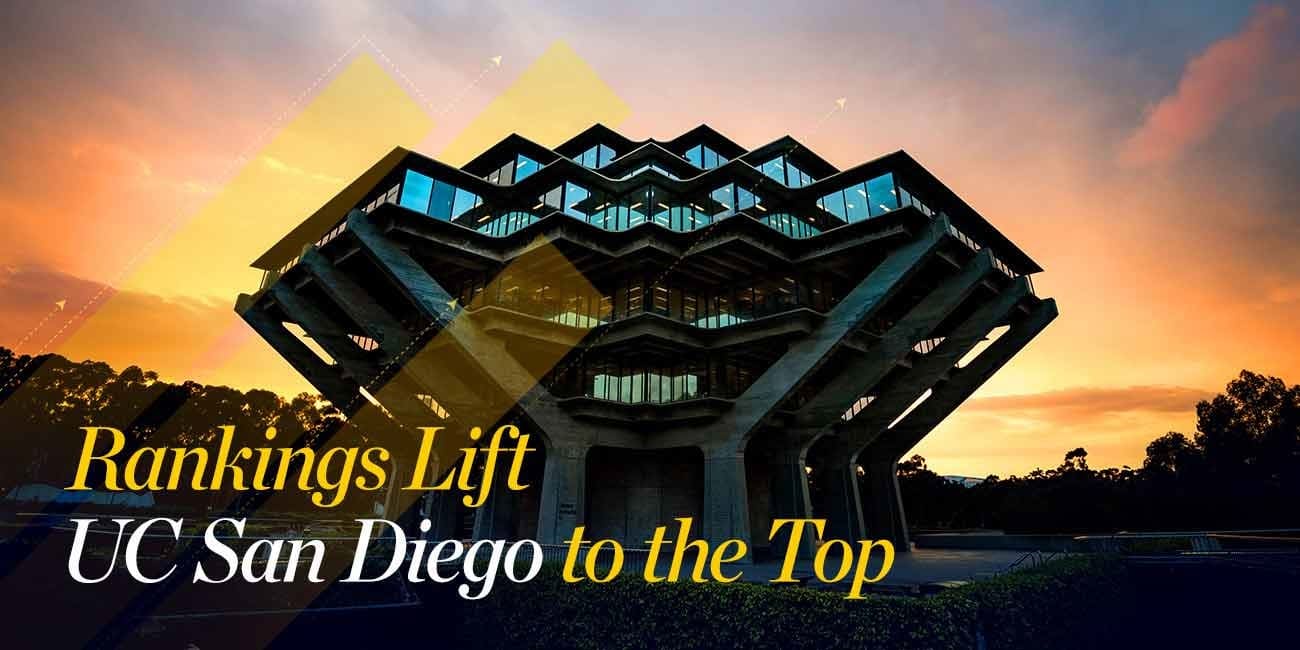}}
\fcolorbox{myOrange}{white}{\includegraphics[width=4.5cm,keepaspectratio]{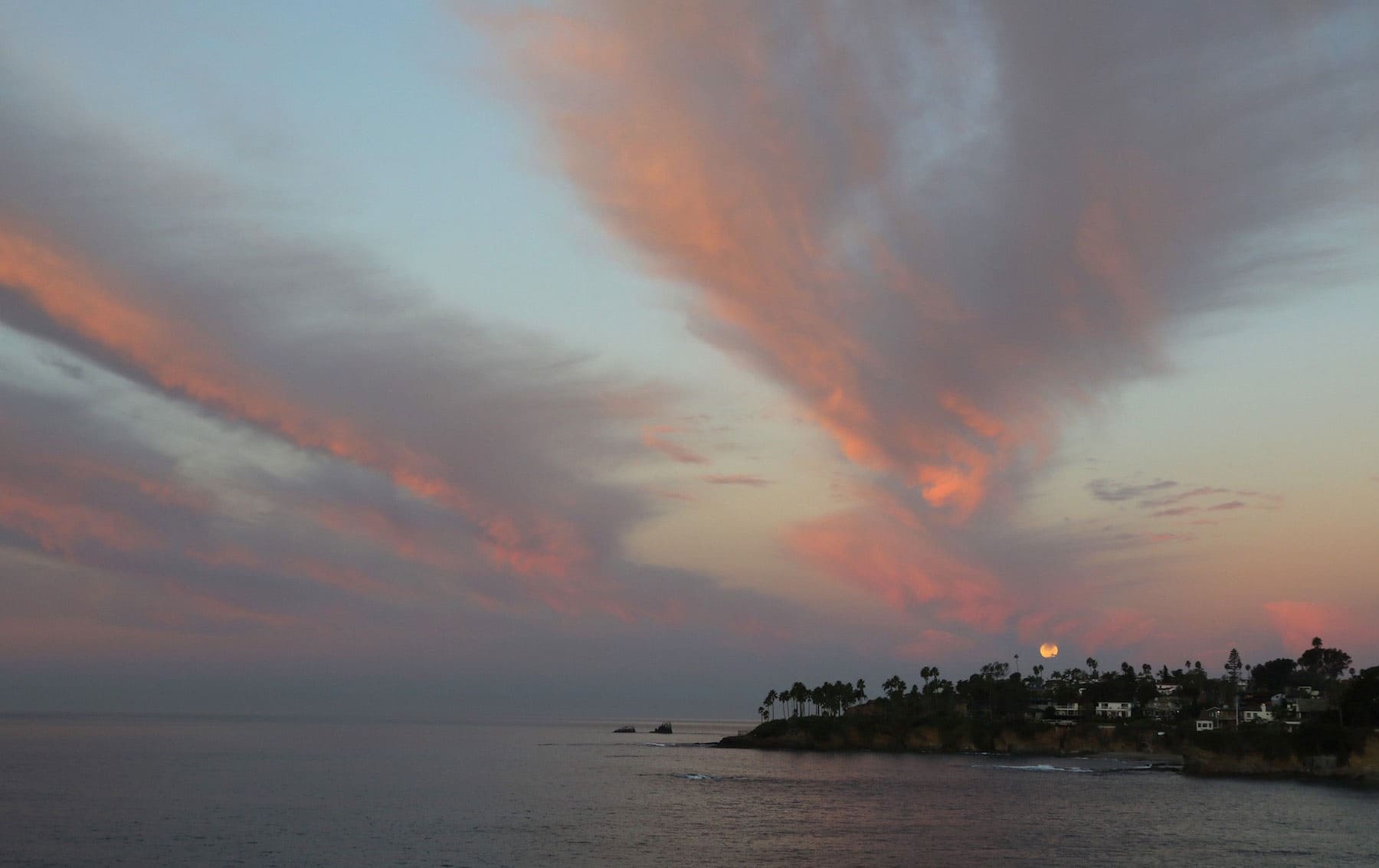}}
\fcolorbox{myOrange}{white}{\includegraphics[width=4.5cm,keepaspectratio]{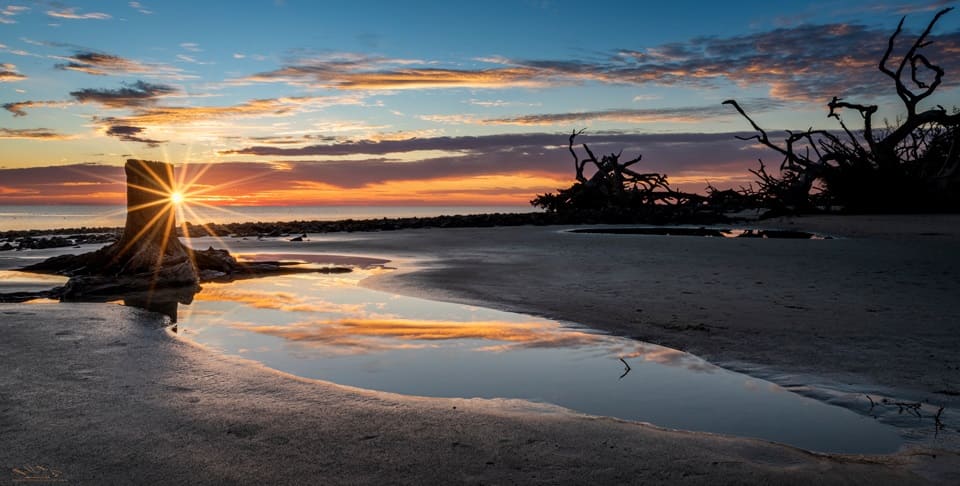}}}  
\\ & \multicolumn{2}{c}{\hspace{-8cm}\large{\textbf{Prediction: \textcolor{ao(english)}{Pristine}}}} \\

\makecell{\fcolorbox{darkred}{lightred}{\begin{varwidth}{\textwidth}   \begin{center} \fcolorbox{myOrange}{white}{\includegraphics[width=4.5cm,keepaspectratio]{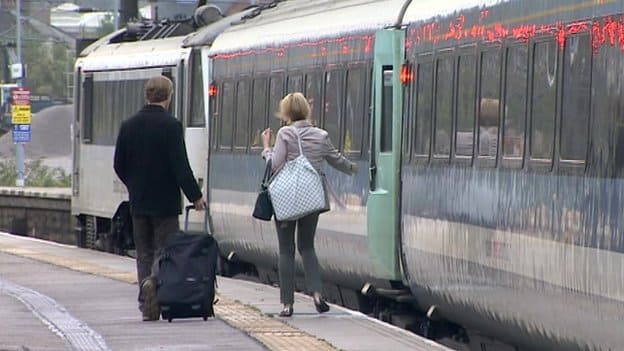}}\end{center}
\fcolorbox{myblue}{white}{\begin{varwidth}{\textwidth}\normalsize{Massachusetts Bay Transportation\\Authority trains sit idle\\early Saturday in Boston}\end{varwidth}}\end{varwidth}}} & 

\makecell{\fcolorbox{myblue}{white}{\begin{varwidth}{\textwidth} \normalsize{`Rail transport', `Rapid transit', \\`Train', `Railroad car', `track',\\`Transport',`Passenger M',\\`M / 06d', `Track',\\`M Line', `Passenger'} \end{varwidth}}   
\fcolorbox{myblue}{white}{\begin{varwidth}{\textwidth} \normalsize{No pages found.} \end{varwidth} }}
& 
\makecell{ \fcolorbox{myOrange}{white}{\includegraphics[width=4.5cm,keepaspectratio]{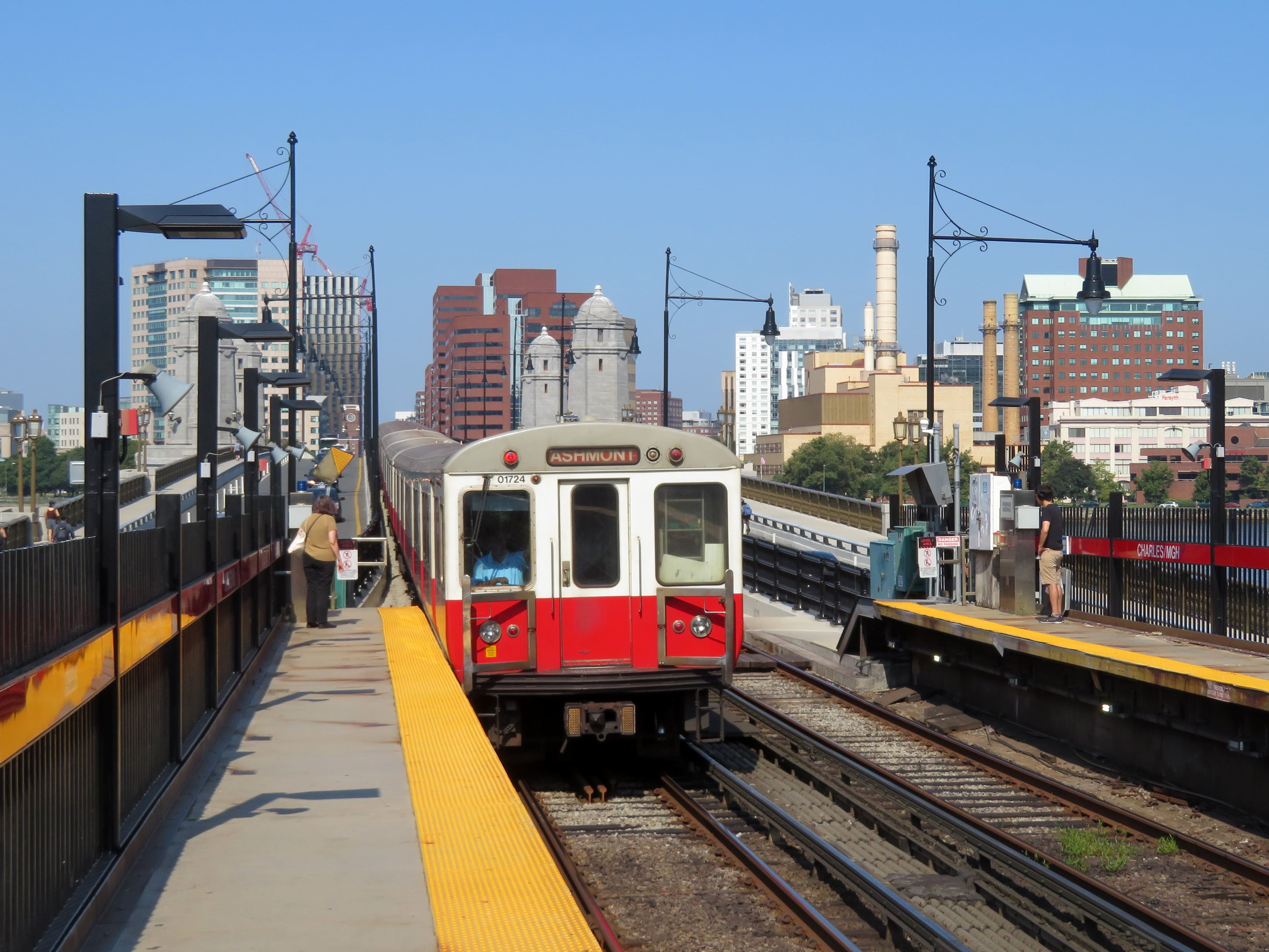}} \fcolorbox{myOrange}{white}{\includegraphics[width=4.5cm,keepaspectratio]{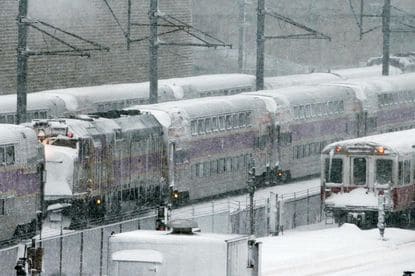}}
\fcolorbox{myOrange}{white}{\includegraphics[width=4.5cm,keepaspectratio]{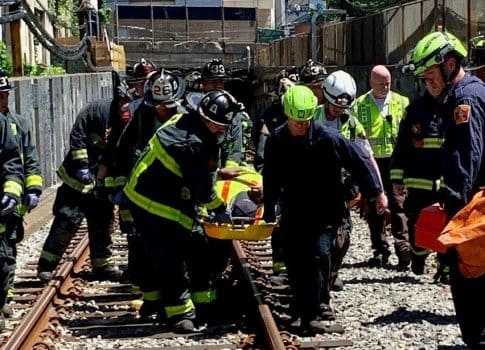}}
\fcolorbox{myOrange}{white}{\includegraphics[width=4.5cm,keepaspectratio]{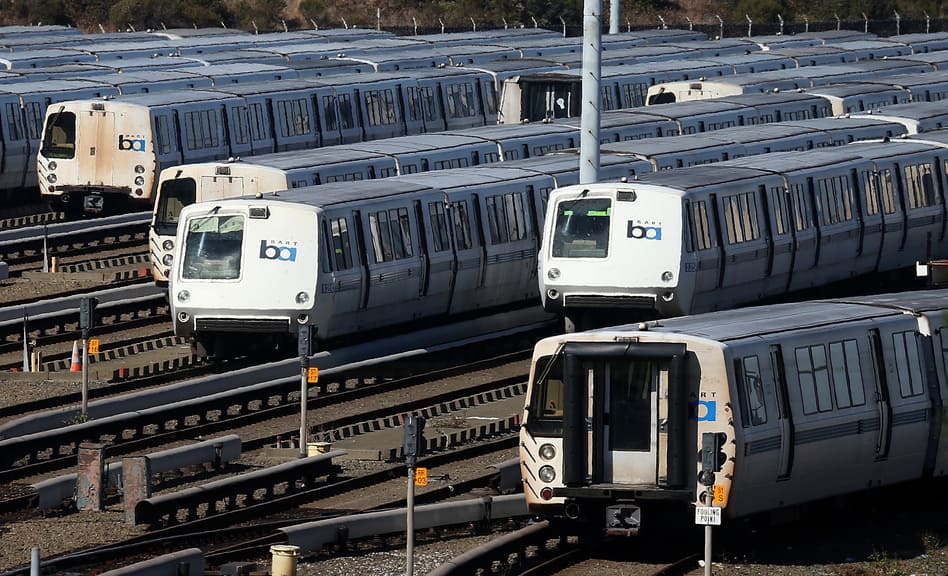}}} \\
&\multicolumn{2}{c}{\hspace{-8cm}\large{\textbf{Prediction: \textcolor{ao(english)}{Pristine}}}}\\ 

\makecell{\fcolorbox{darkred}{lightred}{\begin{varwidth}{\textwidth}   \begin{center} \fcolorbox{myOrange}{white}{\includegraphics[width=4.5cm,keepaspectratio]{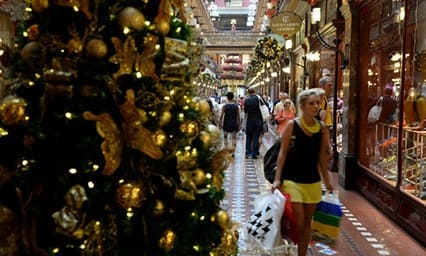}}\end{center}
\fcolorbox{myblue}{white}{\begin{varwidth}{\textwidth}\normalsize{A world which has left\\the Soviet era\\behind shoppers in Moscow}\end{varwidth}}\end{varwidth}}} & 

\makecell{\fcolorbox{myblue}{white}{\begin{varwidth}{\textwidth} \normalsize{`Christmas Tree', `Christmas Day',\\`Bazaar', 'Christmas lights',\\`Marketplace', `Shopping',\\`Tree', `Public space', `Meter',\\`Tradition', `Public',`Lighting',\\`Space',`Market m*'} \end{varwidth}}   
\fcolorbox{myblue}{white}{\begin{varwidth}{\textwidth} \normalsize{No pages found.} \end{varwidth} }}
& 
\makecell{ \fcolorbox{myOrange}{white}{\includegraphics[width=4.5cm,keepaspectratio]{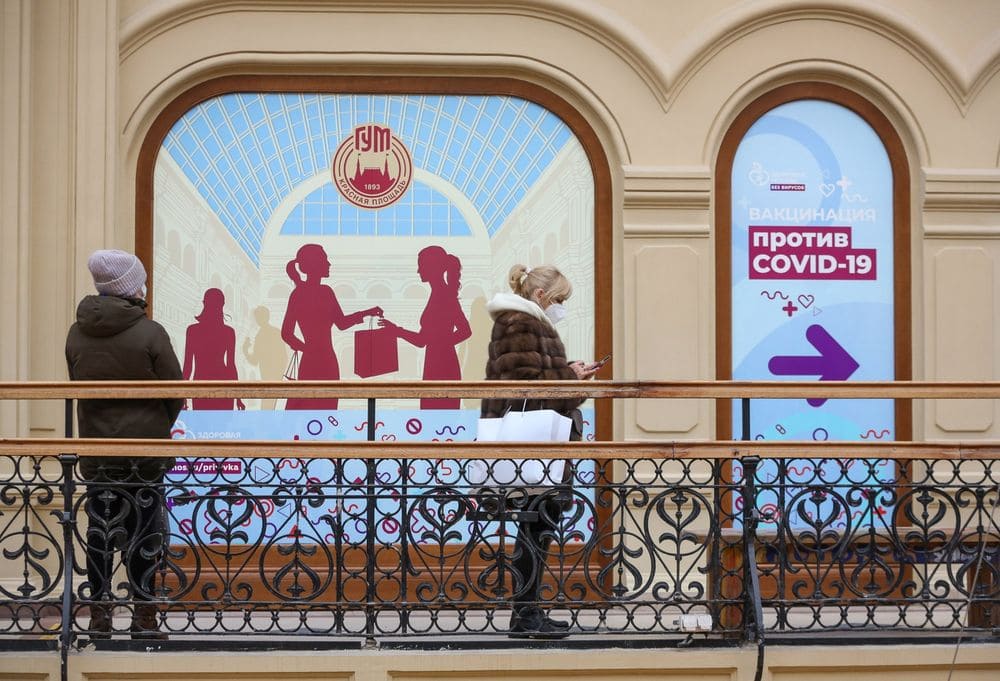}} \fcolorbox{myOrange}{white}{\includegraphics[width=4.5cm,keepaspectratio]{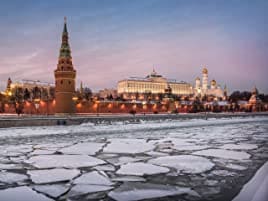}}
\fcolorbox{myOrange}{white}{\includegraphics[width=4.5cm,keepaspectratio]{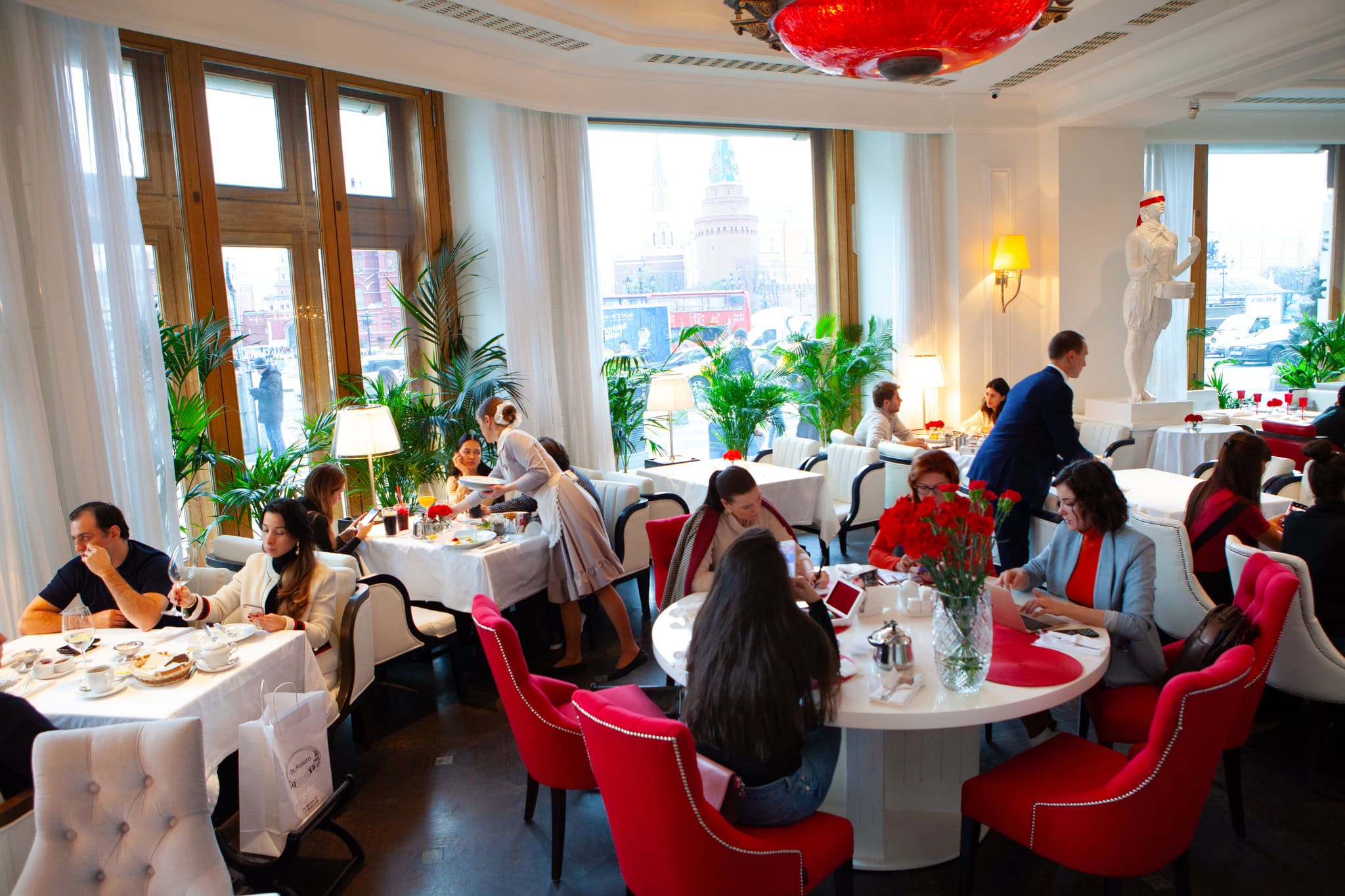}}
\fcolorbox{myOrange}{white}{\includegraphics[width=4.5cm,keepaspectratio]{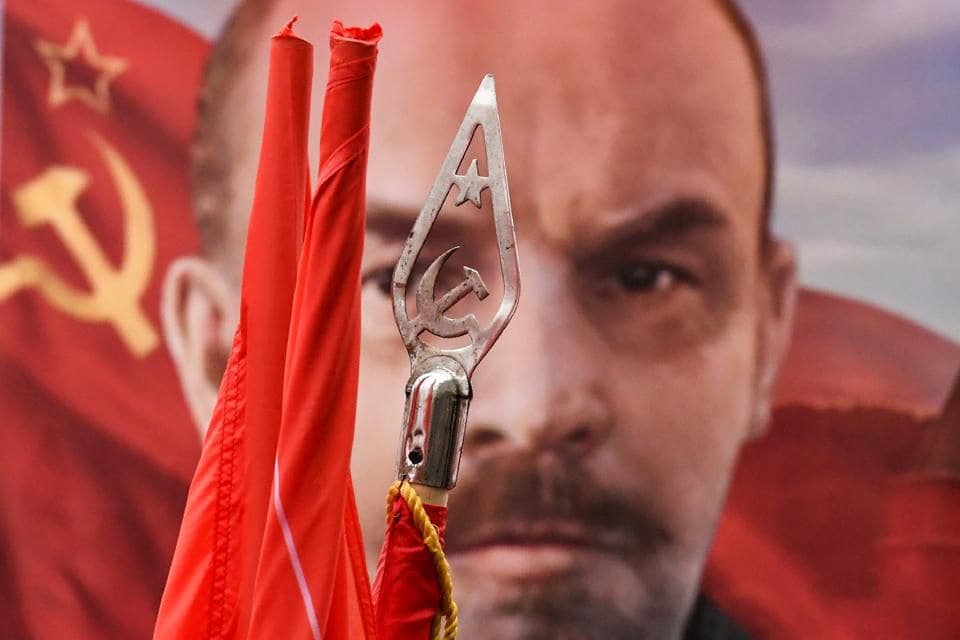}}} \\
&\multicolumn{2}{c}{\hspace{-8cm}\large{\textbf{Prediction: \textcolor{darkred}{Falsified}}}}\\ \midrule

\makecell{\fcolorbox{darkred}{lightred}{\begin{varwidth}{\textwidth}   \begin{center} \fcolorbox{myOrange}{white}{\includegraphics[width=4.5cm,keepaspectratio]{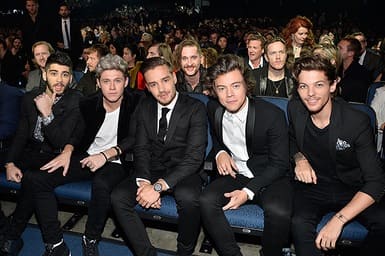}}\end{center}
\fcolorbox{myblue}{white}{\begin{varwidth}{\textwidth}\normalsize{Niall Horan from left Harry Styles\\Liam Payne Zayn Malikand\\Louis Tomlinson of the musical\\group One Direction perform}\end{varwidth}}\end{varwidth}}} & 

\makecell{\fcolorbox{myblue}{white}{\begin{varwidth}{\textwidth} \normalsize{`Harry Styles','2013', `Pop rock'\\`American Music Awards of 2013',\\`Live From the Red Carpet:\\The 2013 American Music Awards', \\`One Direction', `Award', `Image',\\`American Music Awards',\\`Zayn', `Louis Tomlinson',\\`Liam Payne',\\`american music awards\\2013 one direction'} \end{varwidth}}   
\fcolorbox{myblue}{white}{\begin{varwidth}{\textwidth} \normalsize{1- AMA Awards: One Direction seated.\\2- One Direction American\\Music Awards 2013\\3- Red Carpet Pictures from the 2013\\American Music Awards(AMA)\\4-One Direction Wins Pop/Rock\\Band/Duo/Group - AMA 2013} \end{varwidth} }}
& 
\makecell{ \fcolorbox{myOrange}{white}{\includegraphics[width=4.5cm,keepaspectratio]{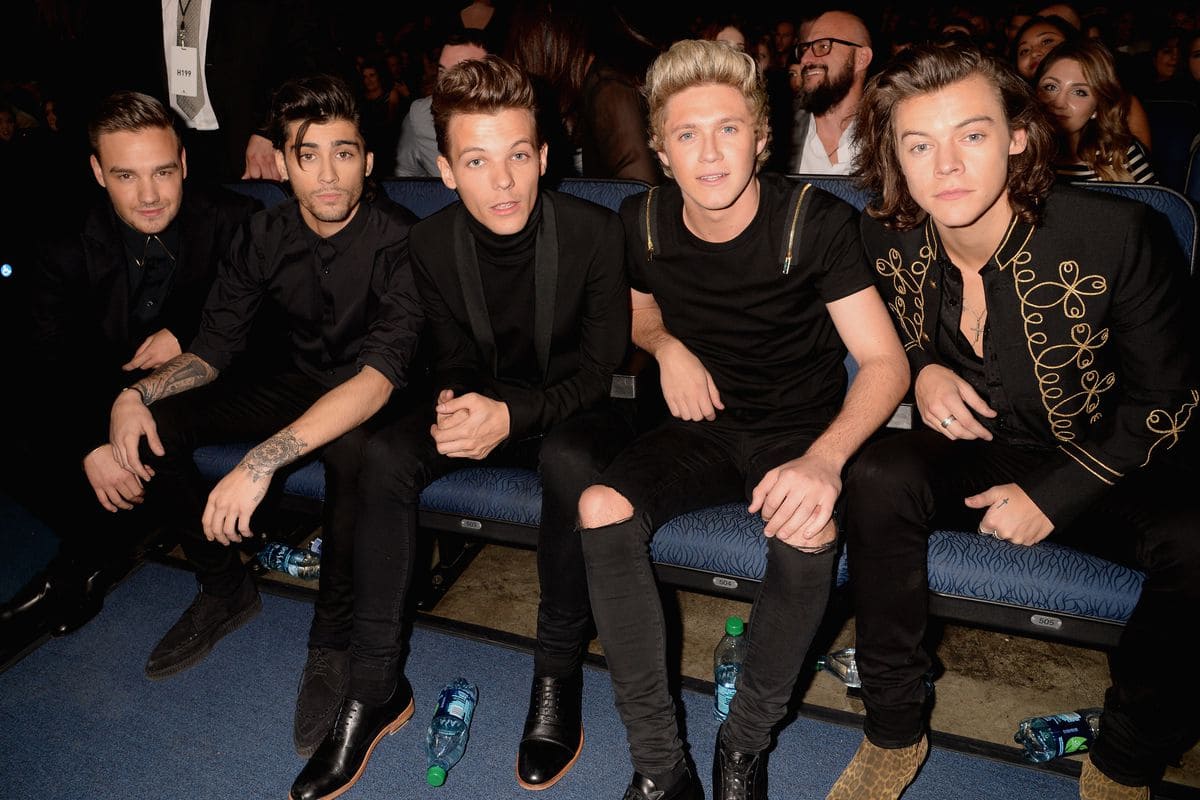}} \fcolorbox{myOrange}{white}{\includegraphics[width=4.5cm,keepaspectratio]{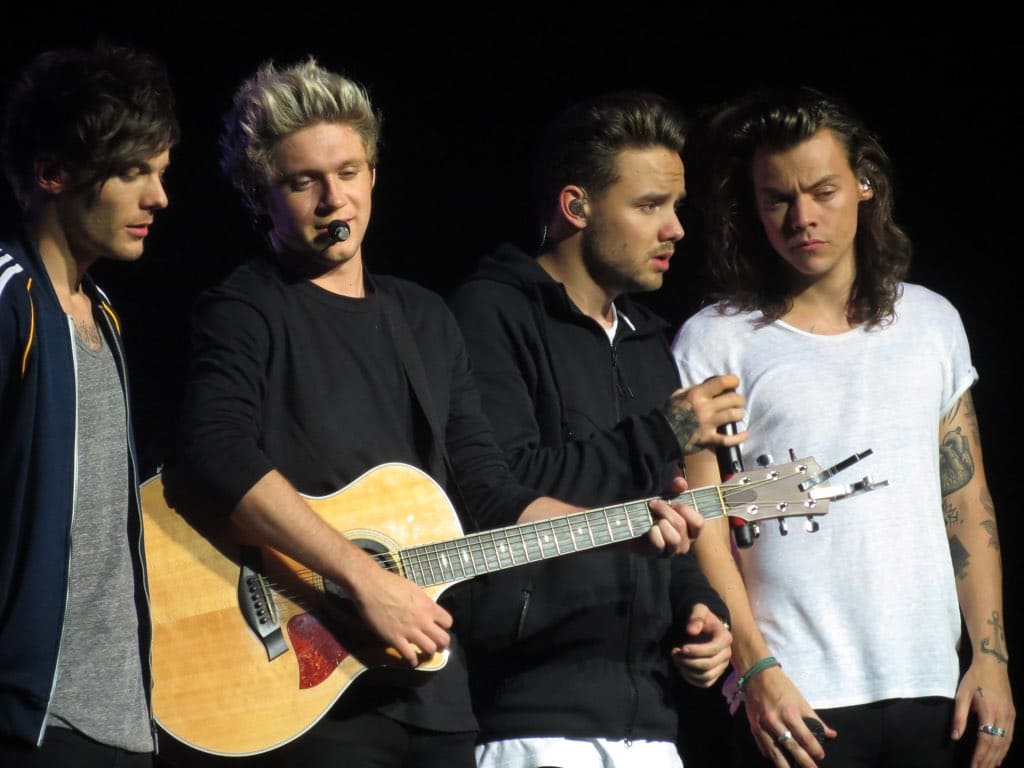}}
\fcolorbox{myOrange}{white}{\includegraphics[width=4.5cm,keepaspectratio]{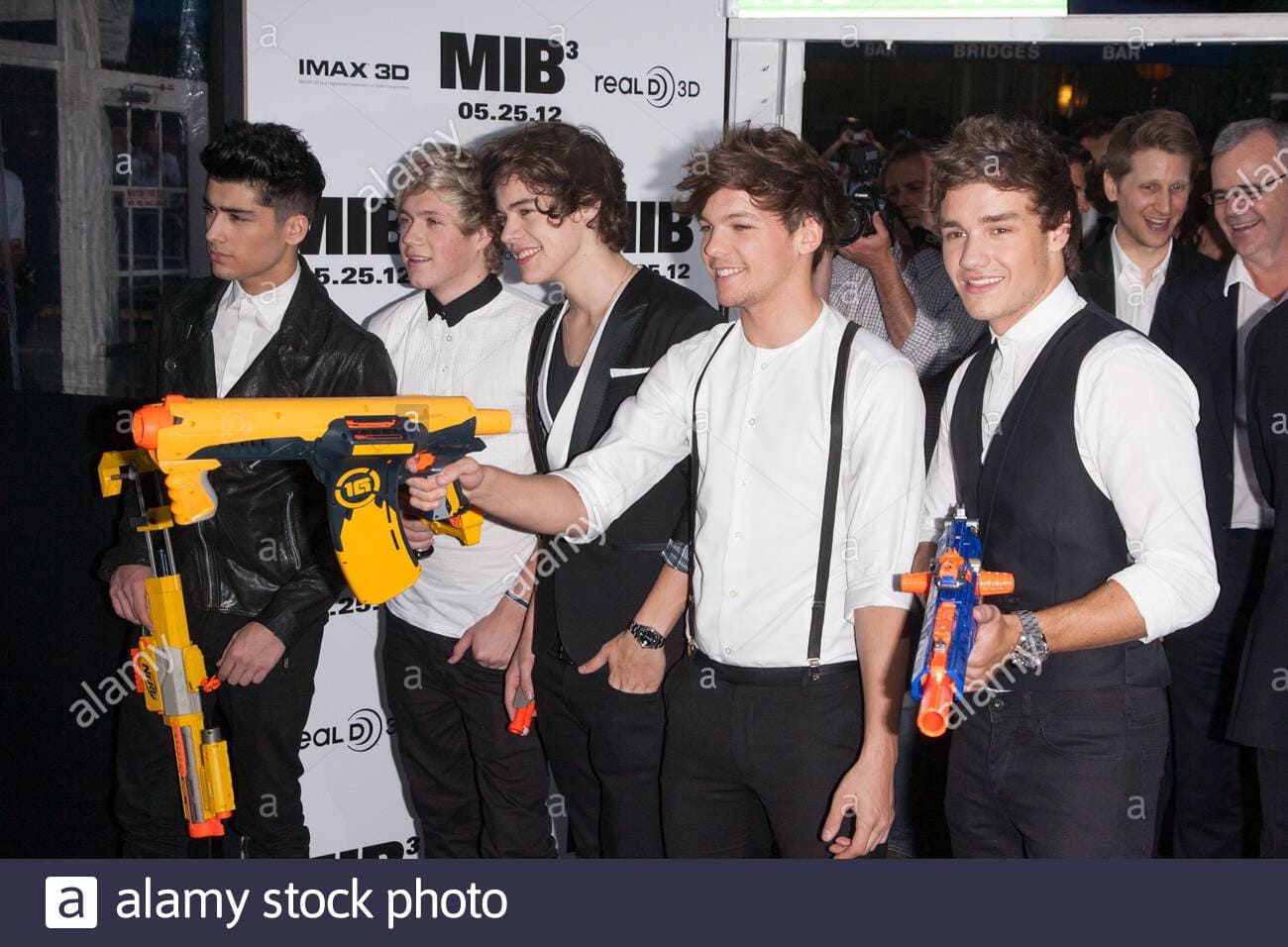}}
\fcolorbox{myOrange}{white}{\includegraphics[width=4.5cm,keepaspectratio]{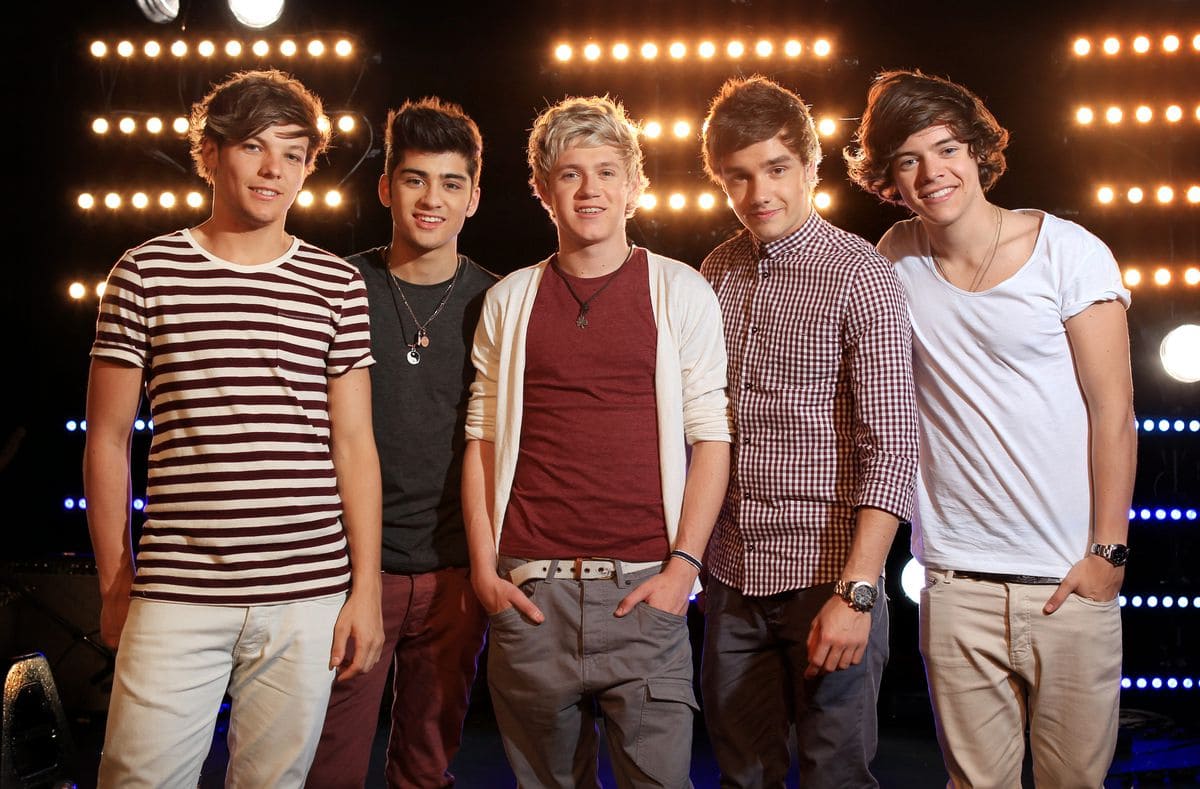}}} \\ 
&\multicolumn{2}{c}{\hspace{-8cm}\large{\textbf{Prediction: \textcolor{ao(english)}{Pristine}}}}\\ \bottomrule 
\end{tabular}}
\captionof{figure}{Some of the examples that were selected as {\bf `hard to verify'} in the user study. The ground truth is indicated by the pairs' background colour; examples with \hlc[lightgreen]{green background} are pristine, \hlc[lightred]{red background} are falsified. The model's prediction is indicated below each example's set; \textbf{\textcolor{ao(english)}{green}} for predicting pristine and \textbf{\textcolor{darkred}{red}} for predicting falsified. The first group of examples does not have textual evidence retrieved, making it harder to verify the context. The last example shows a falsified example with a similar context to the evidence, therefore, the evidence is highly similar to the query.}
\label{tbl:qual_study_appendix}
\end{table*}

\begin{table*}[!t]
\centering
\resizebox{\linewidth}{!}{%
\begin{tabular}{c|c c}
\toprule
\textbf{\textcolor{myOrange}{\large{Image}}}-\textbf{\textcolor{myblue}{\large{caption}}} \large{pair} & \large{\textbf{\textcolor{myblue}{Textual evidence}}} \largericon{figs/icon3.pdf} & \large{\textbf{\textcolor{myOrange}{Visual evidence}}} \largericon{figs/icon4.pdf} \\ \midrule
\makecell{\fcolorbox{ao(english)}{lightgreen}{
\begin{varwidth}{\textwidth} \begin{center}\fcolorbox{myOrange}{white}{\includegraphics[width=5cm,keepaspectratio]{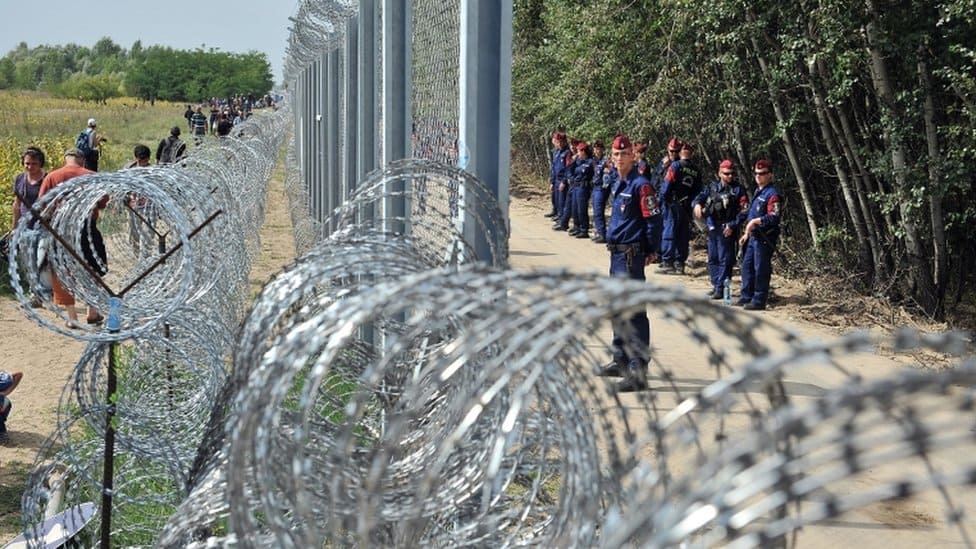}}\end{center} 
\fcolorbox{myblue}{white}{\begin{varwidth}{\textwidth}\normalsize{Hungary has erected a fence on\\its border with Serbia}\end{varwidth} }\end{varwidth}}} & 

\makecell{\fcolorbox{myblue}{white}{\begin{varwidth}{\textwidth} \normalsize{\hlc[light_yellow]{`Hungary'}, \hlc[light_yellow]{`European migrant crisis'},\\\hlc[light_yellow]{`Refugee'}, `Human migration',\\\hlc[light_yellow]{`Immigration'}, `Border', \\`Fence',`Hungarians', `Asylum seeker'\\`Hungary–Serbia border',\\`Hungarian border barrier',\\`International law',`Refugee law',\\`hungary fences refugees'} \end{varwidth}}
\fcolorbox{myblue}{white}{\begin{varwidth}{\textwidth} \normalsize{\hlc[light_yellow]{1- Hungary police recruit}\\\hlc[light_yellow]{border-hunters.}\\2-Migrants and refugees walk\\near razor-wire along a 3-meter-high\\fence secured by Hungarian police\\at the official border crossing\\between Serbia and Hungary.} \end{varwidth}}}
& 
\makecell{ \fcolorbox{myOrange}{light_yellow}{\includegraphics[width=5cm,keepaspectratio]{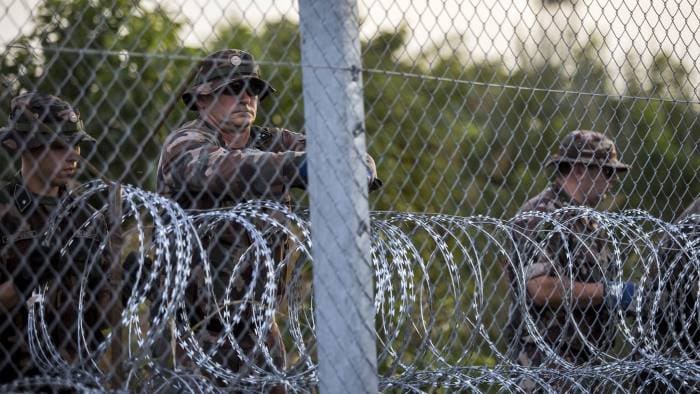}} \fcolorbox{myOrange}{white}{\includegraphics[width=5cm,keepaspectratio]{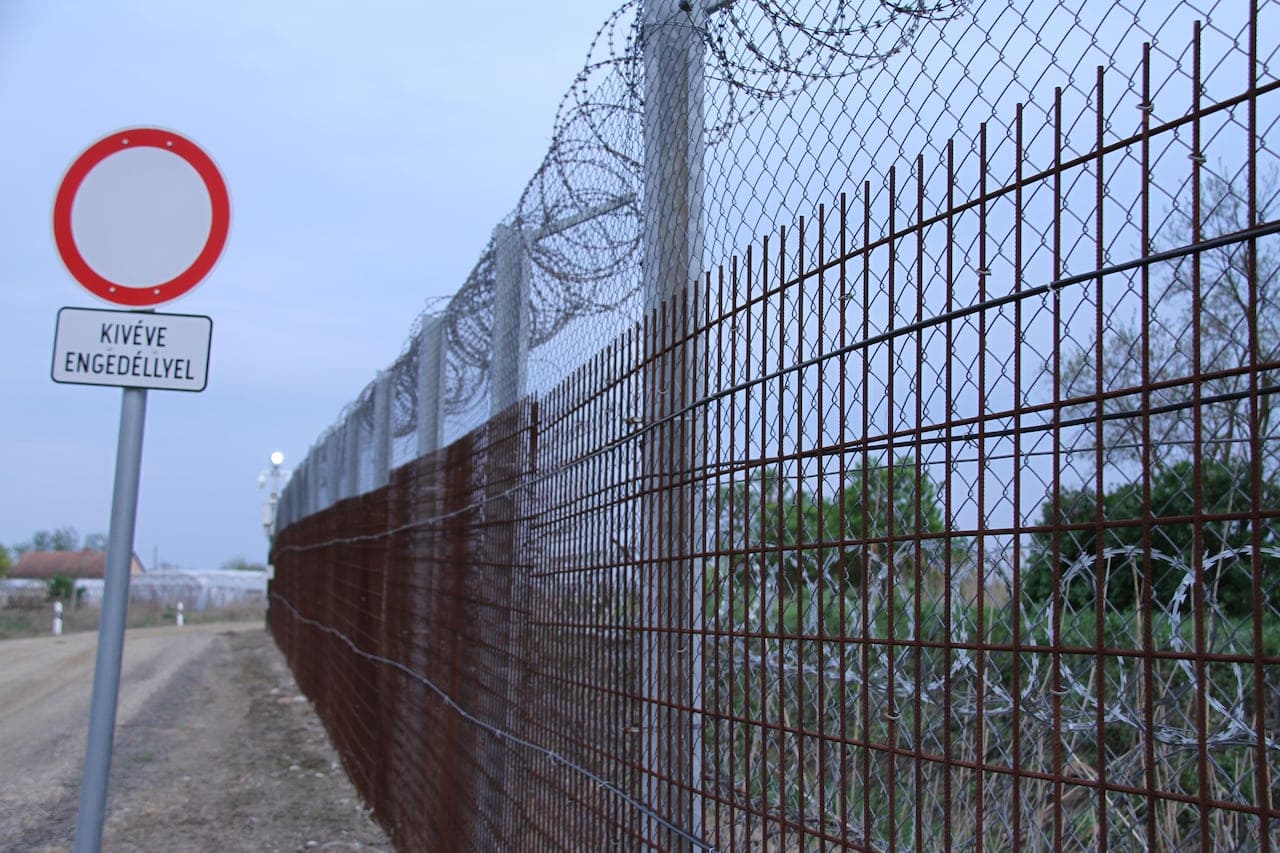}}
\fcolorbox{myOrange}{white}{\includegraphics[width=5cm,keepaspectratio]{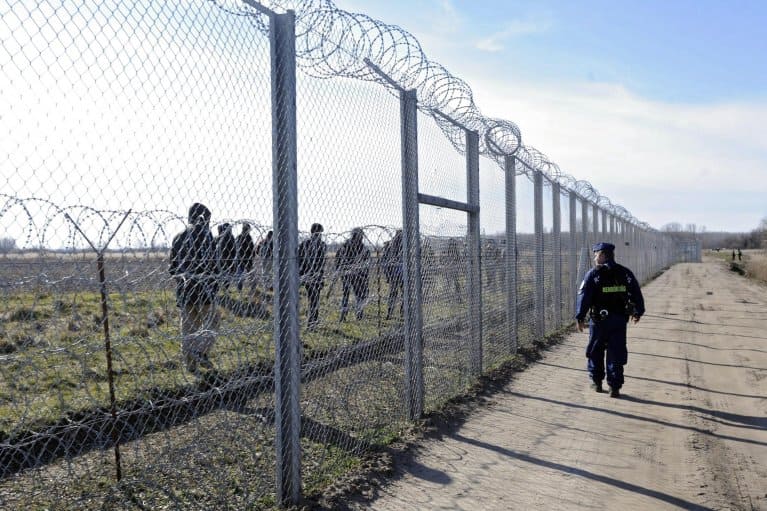}}}
\\ & \multicolumn{2}{c}{\hspace{-4cm}\large{\textbf{Prediction: \textcolor{ao(english)}{Pristine}}}} \\

\makecell{\fcolorbox{ao(english)}{lightgreen}{
\begin{varwidth}{\textwidth} \begin{center}\fcolorbox{myOrange}{white}{\includegraphics[width=5cm,keepaspectratio]{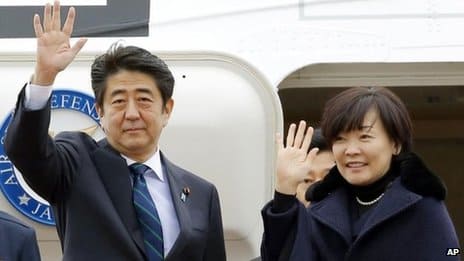}}\end{center} 
\fcolorbox{myblue}{white}{\begin{varwidth}{\textwidth}\normalsize{Last year Shinzo Abe said Africa\\would help drive global growth\\in the future
}\end{varwidth} }\end{varwidth}}} & 

\makecell{\fcolorbox{myblue}{white}{\begin{varwidth}{\textwidth} \normalsize{\hlc[light_yellow]{`Shinzo Abe'}, `Akie Abe',\\`Prime Minister of Japan',\\`Japan', `Prime minister',\\\hlc[light_yellow]{`Trinidad and Tobago'}, \\`Dominica Vibes News',\\\hlc[light_yellow]{`United National Congress'},\\`Week', `Businessperson', 'official'\\\hlc[light_yellow]{`Dominica Housing Recovery Project'}} \end{varwidth}}
\fcolorbox{myblue}{white}{\begin{varwidth}{\textwidth} \normalsize{\hlc[light_yellow]{1-Japanese Prime Minister}\\\hlc[light_yellow]{Shinzo Abe and his wife,}\\\hlc[light_yellow]{Akie Abe.}\\2-Japanese Prime Minister\\Shinzo Abe, center, and\\his wife Akie wave as they\\depart for Africa, at Haneda\\Airport in Tokyo Thursday.} \end{varwidth}}}
& 
\makecell{ \fcolorbox{myOrange}{light_yellow}{\includegraphics[width=5cm,keepaspectratio]{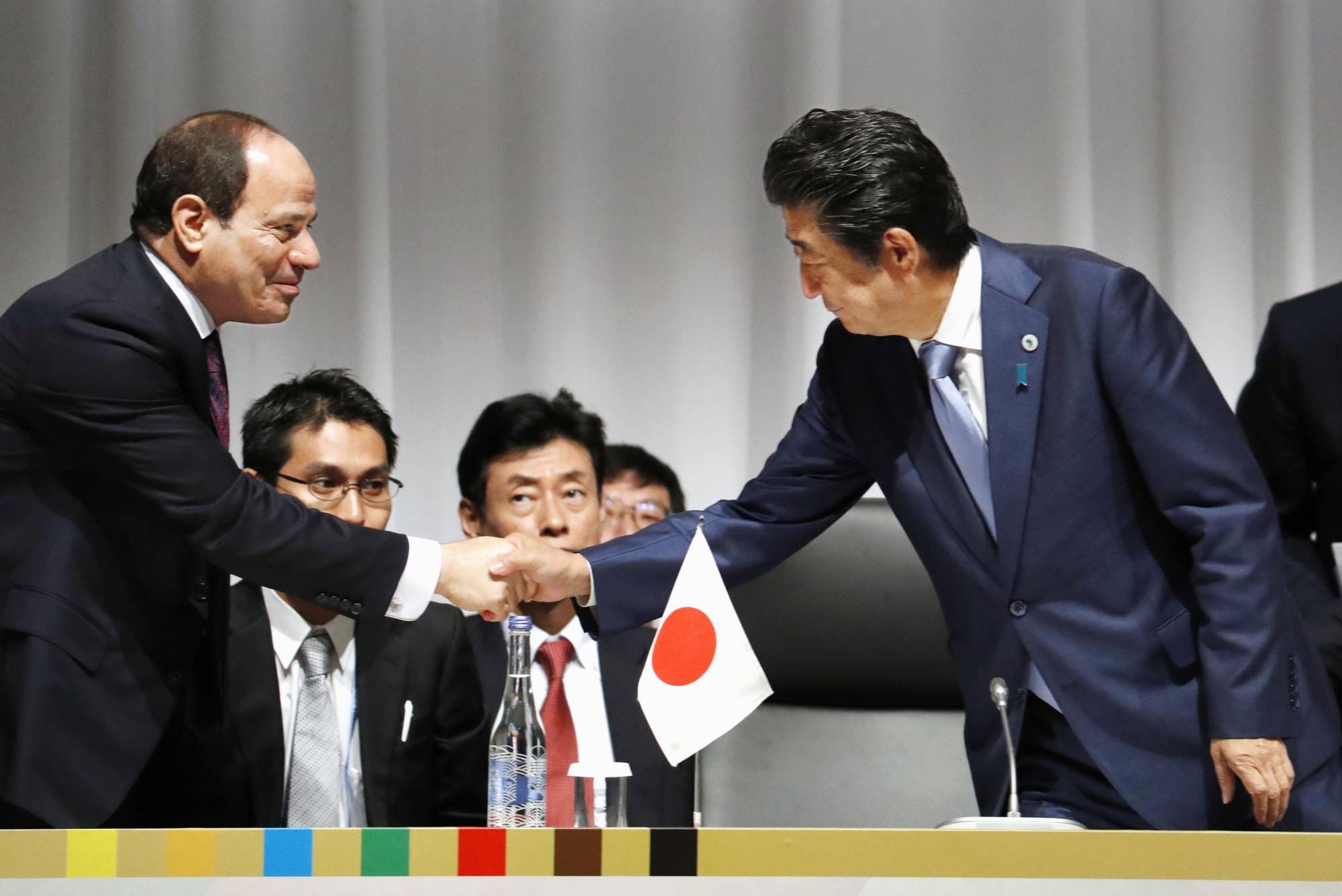}} \fcolorbox{myOrange}{white}{\includegraphics[width=5cm,keepaspectratio]{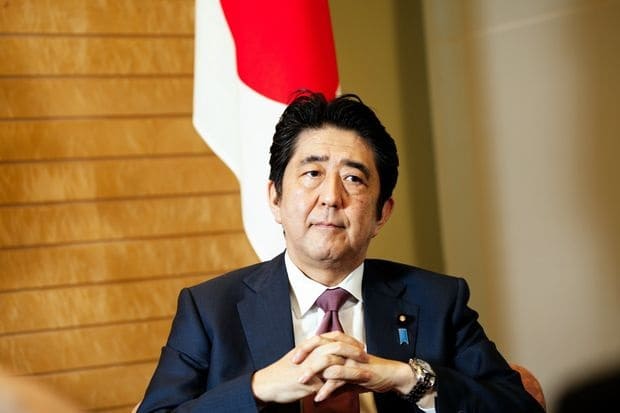}}
\fcolorbox{myOrange}{white}{\includegraphics[width=6cm,keepaspectratio]{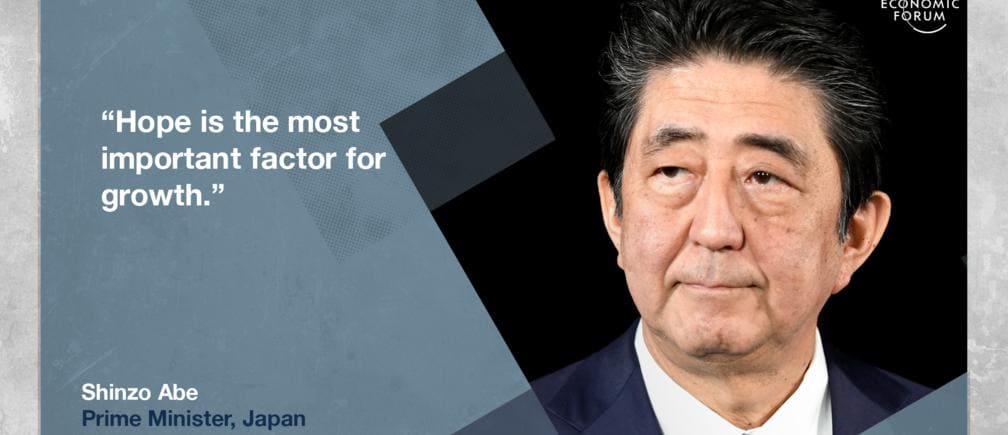}}}  
\\ & \multicolumn{2}{c}{\hspace{-4cm}\large{\textbf{Prediction: \textcolor{ao(english)}{Pristine}}}} \\

\makecell{\fcolorbox{ao(english)}{lightgreen}{
\begin{varwidth}{\textwidth} \begin{center}\fcolorbox{myOrange}{white}{\includegraphics[width=5cm,keepaspectratio]{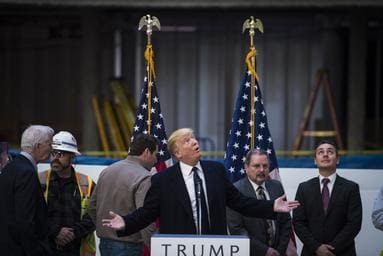}}\end{center} 
\fcolorbox{myblue}{white}{\begin{varwidth}{\textwidth}\normalsize{The GOP candidate at the soontobe\\Trump International Hotel a couple\\of blocks from the White House\\on Pennsylvania Avenue}\end{varwidth} }\end{varwidth}}} & 

\makecell{\fcolorbox{myblue}{white}{\begin{varwidth}{\textwidth} \normalsize{`Judge', \hlc[light_yellow]{`Legal case'}, `official'\\\hlc[light_yellow]{`Superior Court of the District}\\\hlc[light_yellow]{of Columbia'},\\`President-Elect', \hlc[light_yellow]{`Deposition'},\\`Court',`Plea', `Chef',\\\hlc[light_yellow]{`A Washington Law Firm'}} \end{varwidth}}
\fcolorbox{myblue}{white}{\begin{varwidth}{\textwidth} \normalsize{\hlc[light_yellow]{1-Republican presidential candidate}\\\hlc[light_yellow]{Donald Trump speaks during}\\\hlc[light_yellow]{a campaign press conference at}\\\hlc[light_yellow]{the at the Old Post Office Pavilion,}\\\hlc[light_yellow]{soon to be a Trump International Hotel}\\2-Judge rejects Trump plea to avoid\\deposition in José Andrés case} \end{varwidth}}}
& 
\makecell{ \fcolorbox{myOrange}{light_yellow}{\includegraphics[width=5cm,keepaspectratio]{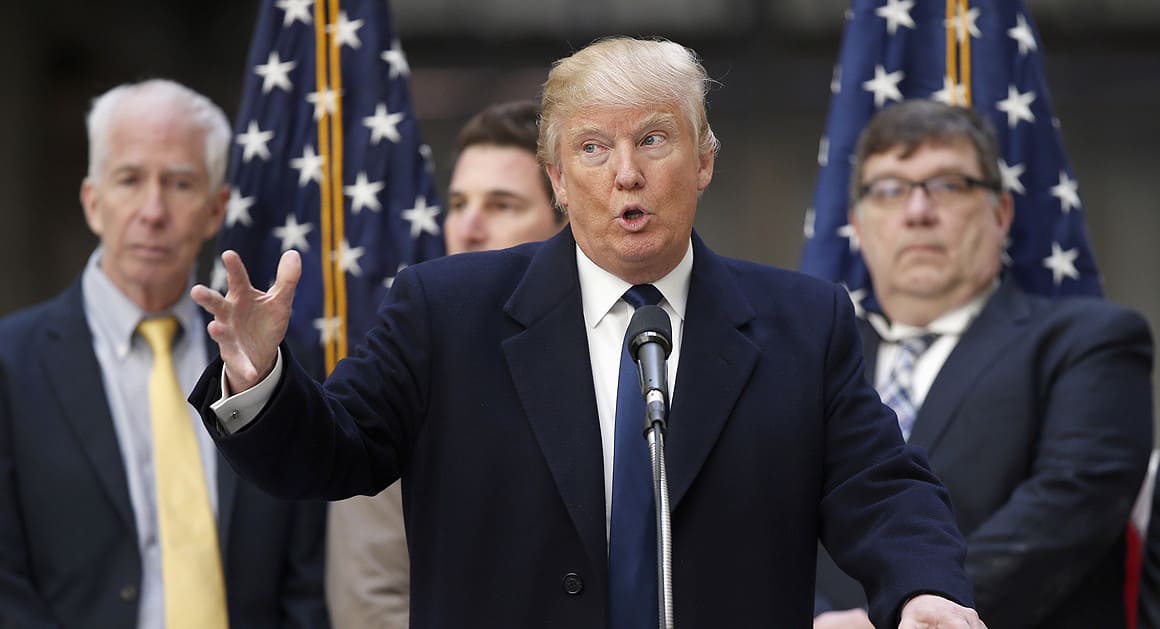}} \fcolorbox{myOrange}{white}{\includegraphics[width=5cm,keepaspectratio]{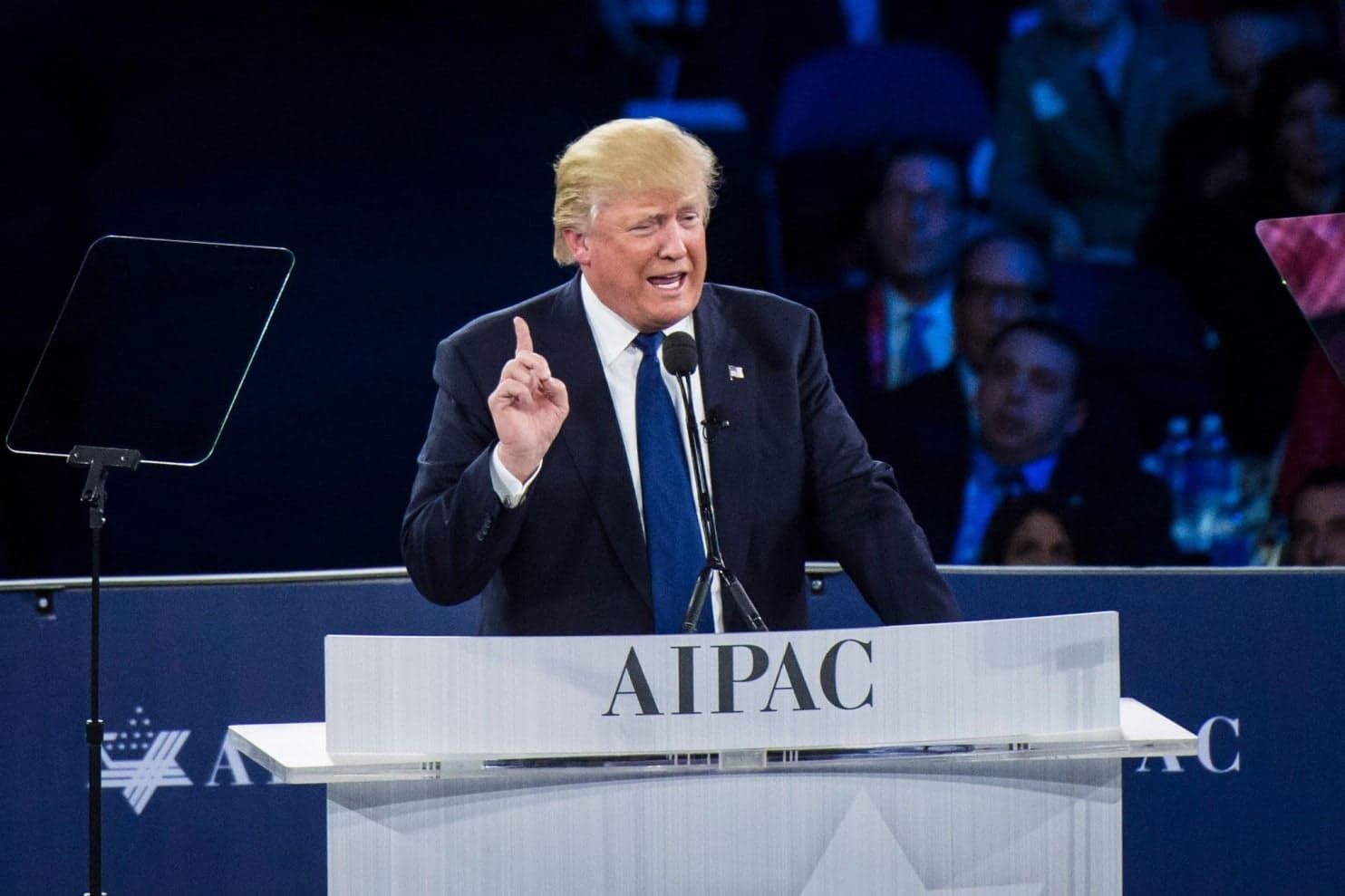}}
\fcolorbox{myOrange}{white}{\includegraphics[width=5cm,keepaspectratio]{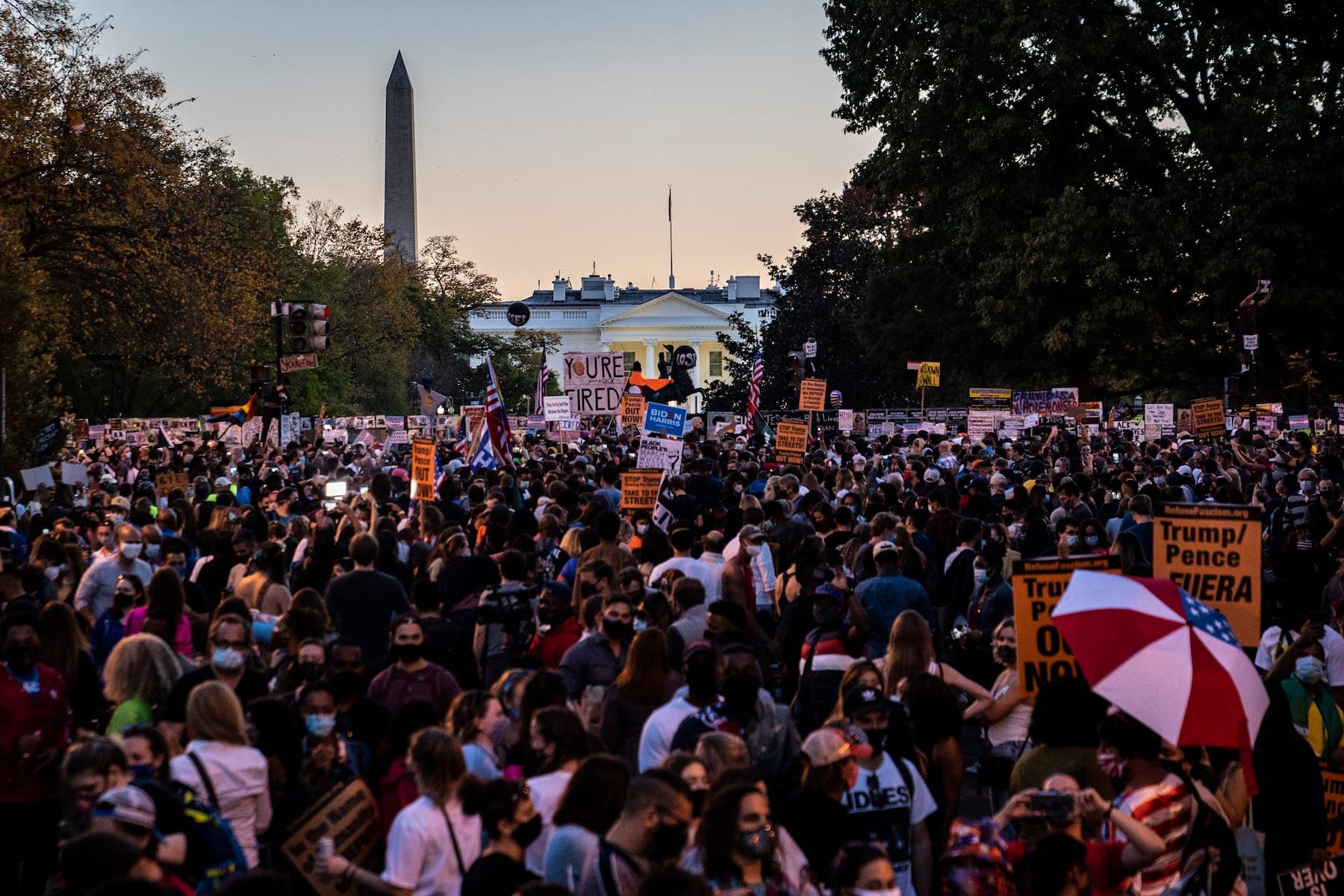}}}  
\\ & \multicolumn{2}{c}{\hspace{-4cm}\large{\textbf{Prediction: \textcolor{ao(english)}{Pristine}}}} \\

\makecell{\fcolorbox{darkred}{lightred}{\begin{varwidth}{\textwidth}   \begin{center} \fcolorbox{myOrange}{white}{\includegraphics[width=5cm,keepaspectratio]{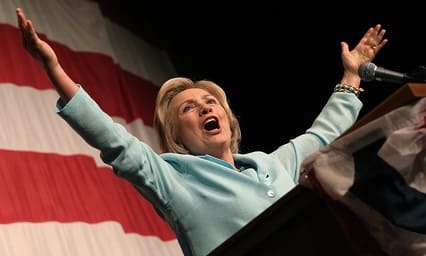}}\end{center}
\fcolorbox{myblue}{white}{\begin{varwidth}{\textwidth}\normalsize{Hillary Clinton speaks at a campaign\\event at Truckee Meadows Community\\College in Reno Nev Aug 25}\end{varwidth}}\end{varwidth}}} & 

\makecell{\fcolorbox{myblue}{white}{\begin{varwidth}{\textwidth} \normalsize{\hlc[light_yellow]{`United States'},`Commentator',\\\hlc[light_yellow]{`President of the United States'},\\\hlc[light_yellow]{`Clinton Foundation'},\\`Dinesh DSouza', `performance'} \end{varwidth}}   
\fcolorbox{myblue}{white}{\begin{varwidth}{\textwidth} \normalsize{\hlc[light_yellow]{1-Democratic presidential candidate}\\\hlc[light_yellow]{Hillary Clinton speaks at the}\\\hlc[light_yellow]{Iowa Democratic Wing Ding on}\\\hlc[light_yellow]{Friday in Clear Lake, Iowa.}\\2-At Wing Ding dinner, Clinton\\proves she still dominates Iowa} \end{varwidth} }}
& 
\makecell{ \fcolorbox{myOrange}{light_yellow}{\includegraphics[width=5cm,keepaspectratio]{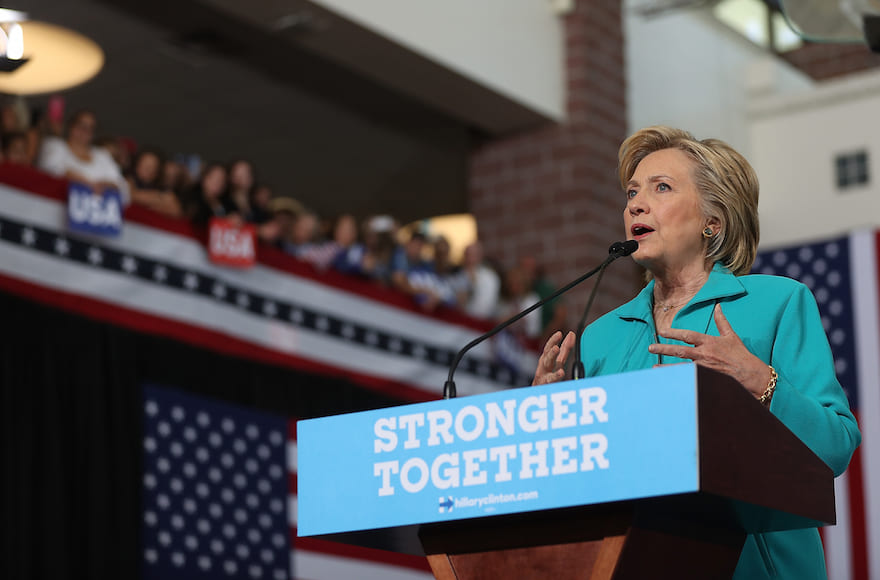}} \fcolorbox{myOrange}{white}{\includegraphics[width=5cm,keepaspectratio]{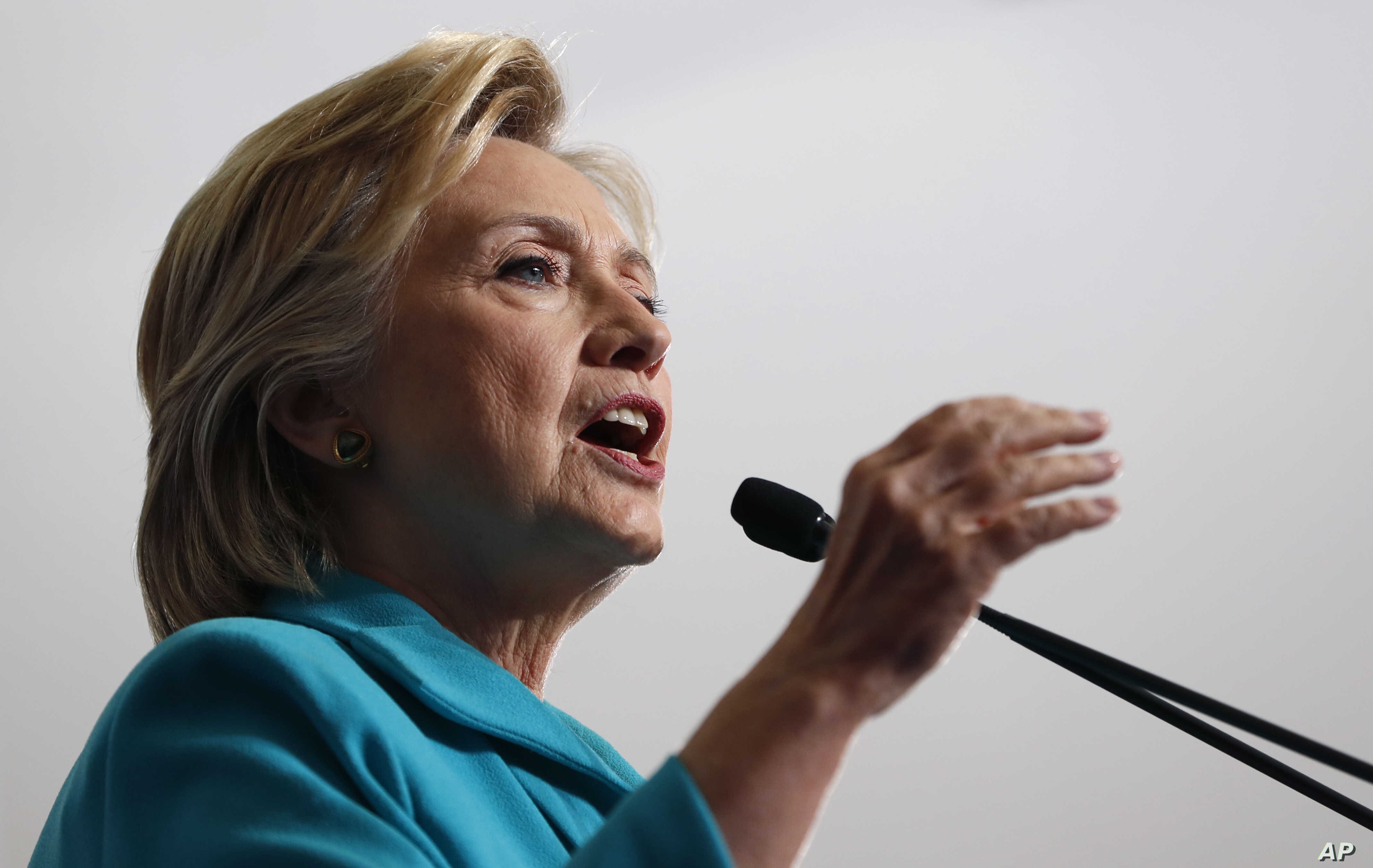}}
\fcolorbox{myOrange}{white}{\includegraphics[width=5cm,keepaspectratio]{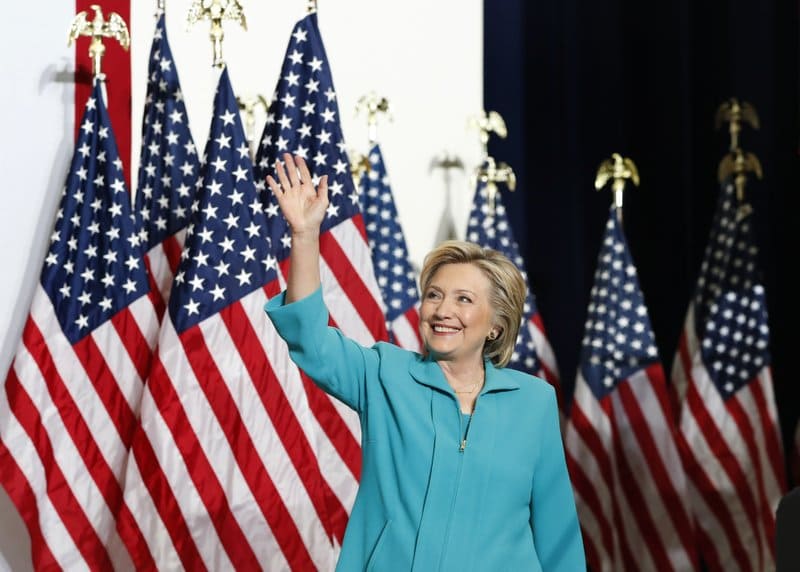}}} \\
&\multicolumn{2}{c}{\large{\hspace{-4cm}\textbf{Prediction: \textcolor{darkred}{Falsified}}}}\\

\makecell{\fcolorbox{darkred}{lightred}{\begin{varwidth}{\textwidth}   \begin{center} \fcolorbox{myOrange}{white}{\includegraphics[width=5cm,keepaspectratio]{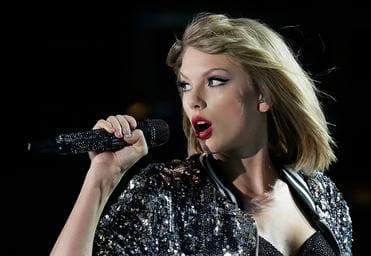}}\end{center}
\fcolorbox{myblue}{white}{\begin{varwidth}{\textwidth}\normalsize{Taylor Swift performs during a concert at\\the Lanxess Arena in Cologne Germany}\end{varwidth}}\end{varwidth}}} & 

\makecell{\fcolorbox{myblue}{white}{\begin{varwidth}{\textwidth} \normalsize{\hlc[light_yellow]{`Taylor Swift'}, `The 1989 World Tour',\\\hlc[light_yellow]{`Grammy Award for Album of the Year'},\\\hlc[light_yellow]{`Grammy Awards'}, `Album', \\`Welcome to New York', `Speak Now',\\`Pop music', `reputation',\\`Apple Music', `Kendrick Lamar',\\`Adele', \hlc[light_yellow]{'taylor swift 1989'}} \end{varwidth}}   
\fcolorbox{myblue}{white}{\begin{varwidth}{\textwidth} \normalsize{\hlc[light_yellow]{1-Taylor Swift performs during}\\\hlc[light_yellow]{her '1989' World Tour Nov. 28,}\\\hlc[light_yellow]{2015, in Sydney, Australia.}\\2-Taylor Swift earned nominations\\in the major categories.\\3-Taylor Swift performs\\during her `1989' tour.} \end{varwidth} }}
& 
\makecell{ \fcolorbox{myOrange}{light_yellow}{\includegraphics[width=5cm,keepaspectratio]{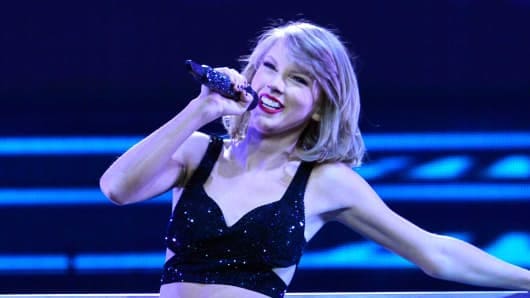}} \fcolorbox{myOrange}{white}{\includegraphics[width=5cm,keepaspectratio]{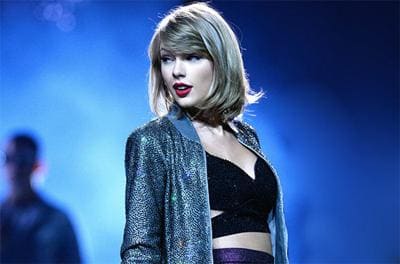}}
\fcolorbox{myOrange}{white}{\includegraphics[width=5cm,keepaspectratio]{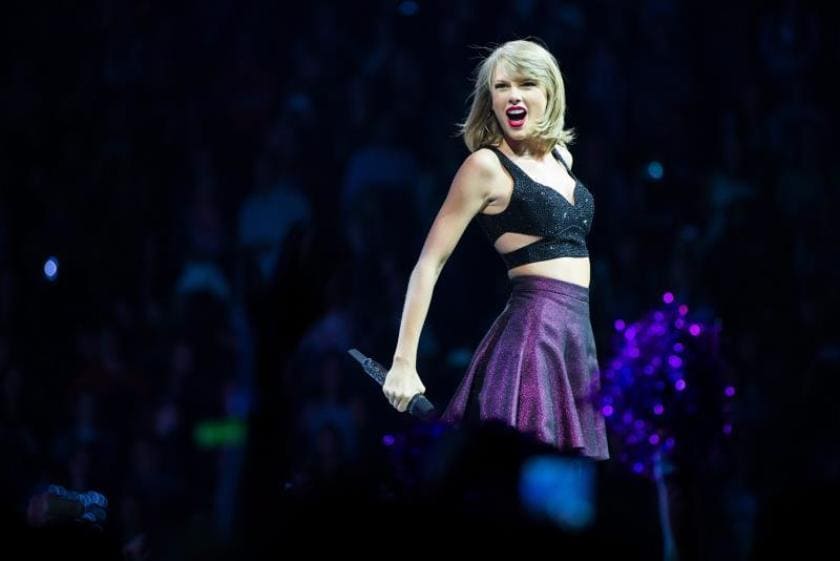}}} \\
&\multicolumn{2}{c}{\hspace{-4cm}\large{\textbf{Prediction: \textcolor{darkred}{Falsified}}}}\\ 

\makecell{\fcolorbox{darkred}{lightred}{\begin{varwidth}{\textwidth}   \begin{center} \fcolorbox{myOrange}{white}{\includegraphics[width=4cm,keepaspectratio]{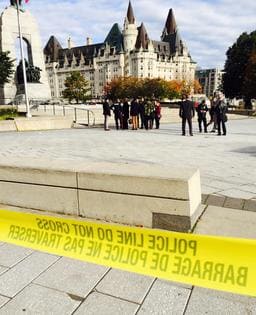}}\end{center}
\fcolorbox{myblue}{white}{\begin{varwidth}{\textwidth}\normalsize{The Blue House the executive office\\and residence of Korea s president}\end{varwidth}}\end{varwidth}}} & 

\makecell{\fcolorbox{myblue}{white}{\begin{varwidth}{\textwidth} \normalsize{`Parliament Hill',`Parliament of Canada'\\, \hlc[light_yellow]{`2014 shootings at Parliament Hill, Ottawa'},\\`Prime Minister of Canada', \\`Royal Canadian Mounted Police', \hlc[light_yellow]{`Ottawa'},\\`Terrorism', `Prime minister',\\`Michael Zehaf-Bibeau', `Stephen Harper',\\ \hlc[light_yellow]{`Kevin Vickers'}, \hlc[light_yellow]{`Ontario'},\\`Canada', `Ottawa'} \end{varwidth}}   
\fcolorbox{myblue}{white}{\begin{varwidth}{\textwidth} \normalsize{\hlc[light_yellow]{1-Police tape surrounds the Canadian}\\\hlc[light_yellow]{War Memorial in Ottawa after a soldier}\\\hlc[light_yellow]{guarding the monument was shot on}\\\hlc[light_yellow]{Wednesday.}\\2-Shooting Near Canada's Parliament.\\3-Shooting at War Memorial in Canada\\Photos} \end{varwidth} }}
& 
\makecell{ \fcolorbox{myOrange}{light_yellow}{\includegraphics[width=5cm,keepaspectratio]{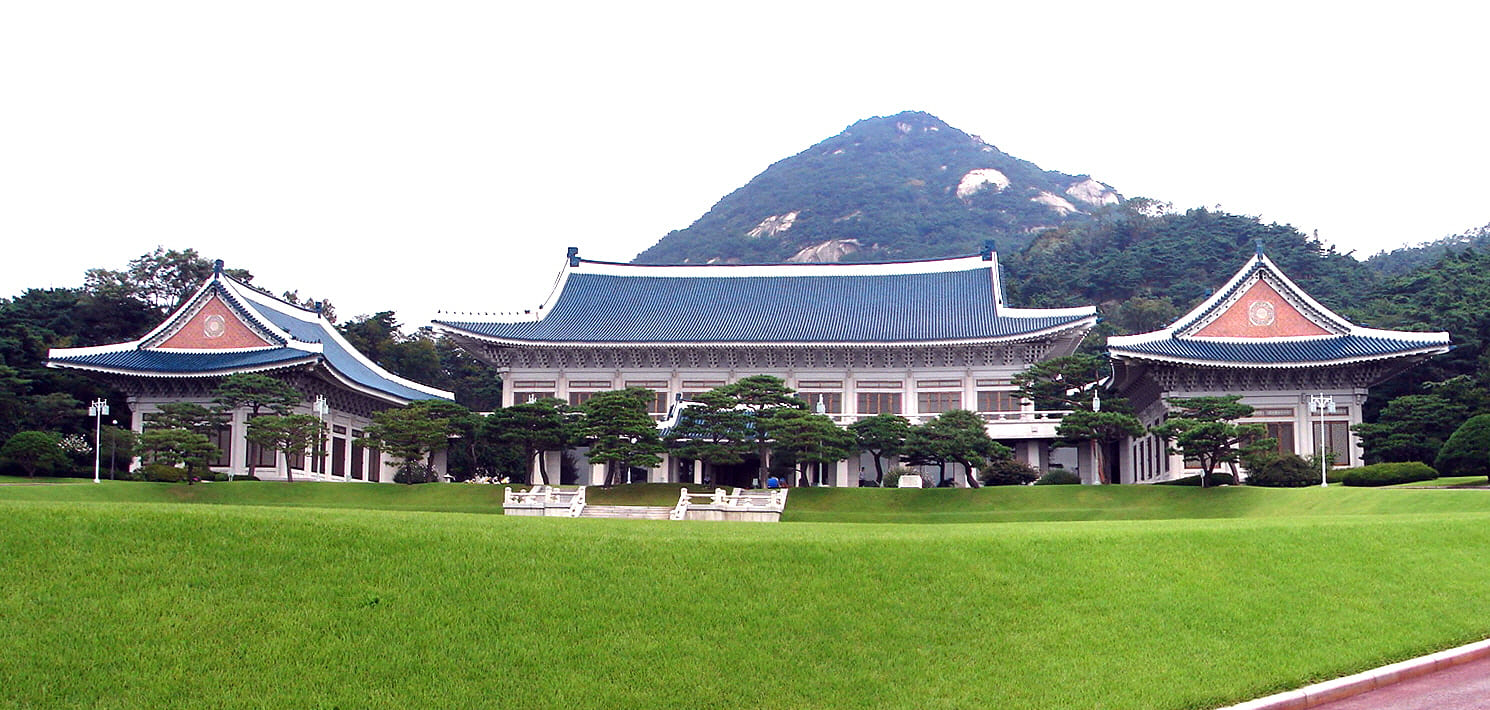}} \fcolorbox{myOrange}{white}{\includegraphics[width=5cm,keepaspectratio]{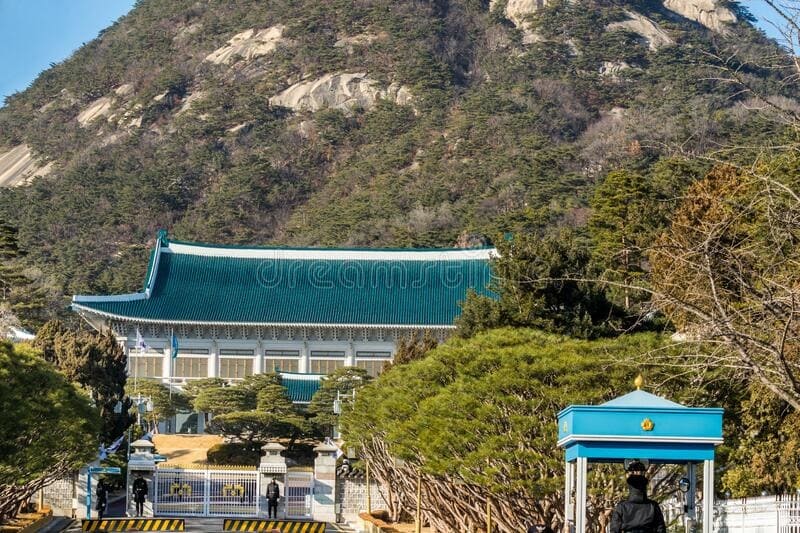}}
\fcolorbox{myOrange}{white}{\includegraphics[width=5cm,keepaspectratio]{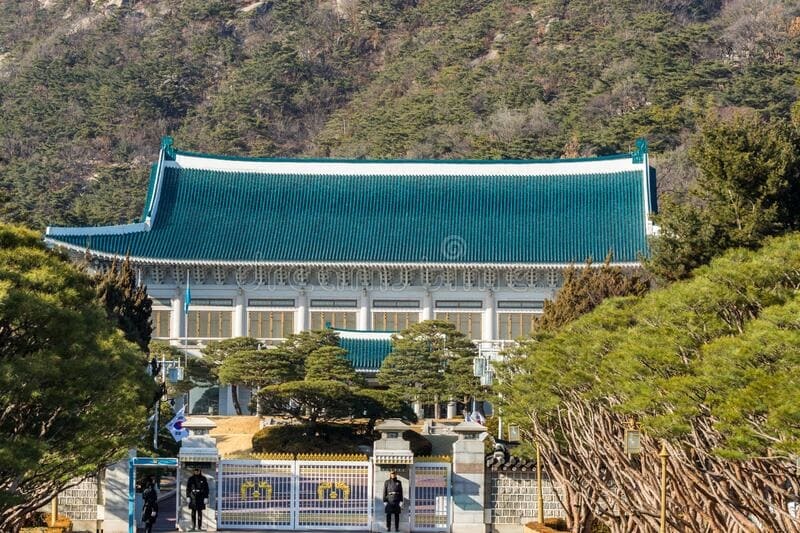}}}\\ 
&\multicolumn{2}{c}{\hspace{-4cm}\large{\textbf{Prediction: \textcolor{darkred}{Falsified}}}}\\ \bottomrule 

\end{tabular}}
\captionof{figure}{Other qualitative examples. The ground truth is indicated by the pairs' background colour; examples with \hlc[lightgreen]{green background} are pristine, \hlc[lightred]{red background} are falsified. The model's prediction is indicated below each example's set; \textbf{\textcolor{ao(english)}{green}} for predicting pristine and \textbf{\textcolor{darkred}{red}} for predicting falsified. \hlc[light_yellow]{Highlighted items} are the ones with the highest attention.}
\label{tbl:qual_appendix}
\end{table*}

\end{document}